\RequirePackage{fix-cm}
\documentclass[smallextended]{svjour3}       % onecolumn (second format)
\smartqed  % flush right qed marks, e.g. at end of proof
\usepackage[english]{babel}
\usepackage[utf8]{inputenc}
\usepackage{times} % assumes new font selection scheme installed
\usepackage{cite}
\usepackage{balance}

\usepackage{hyperref}
\hypersetup{
    colorlinks=true,
    linkcolor=black,
    urlcolor=blue
}

\usepackage[nolist]{acronym}

\usepackage{multirow}

\usepackage[binary-units]{siunitx}
\usepackage{graphicx} % for pdf, bitmapped graphics files
\usepackage{epsfig} % for postscript graphics files
\usepackage{pgfplots}
\usepackage[font=small]{caption}
\usepackage[font=footnotesize]{subcaption}
\usepackage{pgfplots}

\usepackage{float}

\usepackage[export]{adjustbox}

\usepackage{booktabs}

\usepackage[markup=bfit, deletedmarkup=sout, authormarkup=brackets]{changes}
\definechangesauthor[name={Pavel S.}, color=blue]{PS}

\usepackage{mathtools}
\usepackage{nicefrac}
\usepackage{amsmath}
\usepackage{amssymb}
\usepackage{bm} % to bold symbols

\DeclareSIUnit{\litre}{l}

\usepackage{siunitx}
\usepackage{algorithm}
\usepackage[noend]{algpseudocode}

\makeatletter
\def\BState{\State\hskip-\ALG@thistlm}
\makeatother

\usepackage{tikz}
\pgfdeclarelayer{foreground}
\pgfsetlayers{background, main, foreground}
\usetikzlibrary{quotes, angles, backgrounds, arrows, automata, shapes, positioning, calc, through, spy, decorations.pathreplacing, decorations.markings, arrows.meta, automata, petri}

\newcommand{\reffig}[1]{Fig.~\ref{#1}}

\tikzset{
    imglabel/.style={
      rectangle,
      inner sep=2pt,
      text=black,
      minimum height=1em,
      text centered,
      fill=white,
      fill opacity=1.0,
      text opacity=1,
      anchor=south west,
    },
  }
\tikzset{
	state/.style={
		rectangle,
		draw=black, very thick,
		minimum height=1.0em,
		text centered,
	},
  smallstate/.style={
    rectangle,
    draw=black, very thick,
    minimum height=0.2em,
    text centered,
  },
}
\tikzset{
  on each segment/.style={
    decorate,
    decoration={
      show path construction,
      moveto code={},
      lineto code={
        \path [#1]
        (\tikzinputsegmentfirst) -- (\tikzinputsegmentlast);
      },
      curveto code={
        \path [#1] (\tikzinputsegmentfirst)
        .. controls
        (\tikzinputsegmentsupporta) and (\tikzinputsegmentsupportb)
        ..
        (\tikzinputsegmentlast);
      },
      closepath code={Acp
        \path [#1]
        (\tikzinputsegmentfirst) -- (\tikzinputsegmentlast);
      },
    },
  },
  mid arrow/.style={postaction={decorate,decoration={
        markings,
        mark=at position .5 with {\arrow[#1]{stealth}}
      }}},
}
\graphicspath{{./figures/}}
\usepackage{enumitem}

\begin{document}

\title{%A Modular UAV Platform for Real-World Experimental Verification of Research of Multi-Robot Systems%
MRS Drone: A Modular Platform for Real-World Deployment of Aerial Multi-Robot Systems%
%
%\thanks{This work was partially funded by the CTU grant no. SGS20/174/OHK3/3T/13, by the Czech Science Foundation (GAČR) under research project no. 20-10280S, no. 20-29531S, no. 22-24425S and no. 23-06162M, by TAČR project no. FW01010317, by the OP VVV funded project CZ.02.1.01/0.0/0.0/16 019/0000765 ``Research Center for Informatics", by the NAKI II project no. DG18P02OVV069, by the European Union's Horizon 2020 research and innovation program AERIAL-CORE under grant agreement no. 871479, by the Defense Advanced Research Projects Agency (DARPA), and by the Technology Innovation Institute - Sole Proprietorship LLC, UAE. Furthermore, computational resources were supplied by the project "e-Infrastruktura CZ" (e-INFRA LM2018140) provided within the program Projects of Large Research, Development and Innovations Infrastructures.}%Grants or other notes
%placed here.
}
% \subtitle{Do you have a subtitle?\\ If so, write it here}

\titlerunning{MRS Drone: A Modular Platform for Real-World Deployment of Aerial Multi-Robot Systems}        % if too long for running head

\author{Daniel Hert \and Tomas Baca \and Pavel Petracek \and Vit Kratky \and Robert Penicka \and Vojtech Spurny \and Matej Petrlik \and Matous Vrba \and David Zaitlik \and Pavel Stoudek \and Viktor Walter \and Petr Stepan \and Jiri Horyna \and Vaclav Pritzl \and Martin Sramek \and Afzal Ahmad \and Giuseppe Silano \and Daniel Bonilla Licea \and Petr Stibinger \and Tiago Nascimento* \and Martin Saska
% First Author         \and        Second Author %etc.
}


\institute{Corresponding author: Tiago Nascimento \at
              Faculty of Electrical Engineering, \\
              Czech Technical University in Prague, \\
              Czech Republic \\
              Tel.: +420 22435 7634\\
              \email{pereiti1@fel.cvut.cz}           
}

\date{Received: date / Accepted: date}

\maketitle

\begin{abstract}

This paper presents a modular autonomous~\acf{UAV} platform called the~\acf{MRS} Drone that 
can be used in a large range of indoor and outdoor applications.
The~\ac{MRS} Drone features unique modularity~\acl{wrt} changes in actuators, frames, and sensory configuration.
As the name suggests, the platform is specially tailored for deployment within a~\ac{MRS} group.
The~\ac{MRS} Drone contributes to the state-of-the-art of~\ac{UAV} platforms by allowing smooth real-world deployment of multiple aerial robots, as well as by outperforming other platforms with its modularity.
For real-world multi-robot deployment in various applications, the platform is easy to both assemble and modify.
Moreover, it is accompanied by a realistic simulator to enable safe pre-flight testing and a smooth transition to complex real-world experiments.
In this manuscript, we present mechanical and electrical designs, software architecture, and technical specifications to build a fully autonomous multi~\ac{UAV} system.
Finally, we demonstrate the full capabilities and the unique modularity of the~\ac{MRS} Drone in various real-world applications that required a diverse range of platform configurations.

\end{abstract}

%%% END SECTION ============================================================

%%% START SECTION ==========================================================

\section{Introduction}
\label{sec:introduction}
%[TO BE WRITTEN] - 1 page
%\textbf{TODO: Tiago + Martin}}

%%% 1. What is the problem?
In recent years, many research groups have developed~\acf{UAV} platforms with a wide range of capabilities and applications in mind.
Still, only a percentage of worldwide~\ac{UAV} research is tested in real-world 
experiments~\cite{nascimento2019position}.
At the same time, research groups and startups starting out with~\ac{UAV} research and development can benefit from open-source platforms, as they significantly decrease their initial development costs.
%%% Why is it important?
Thus, there exists a need for an open-source platform allowing for easy real-world deployment 
and a high degree of modularity, such as that presented by the~\ac{MRS} Drone.
In particular, the modularity aspect allows the platform to be deployed in different 
applications and has the potential to push the frontiers of~\ac{UAV} research worldwide for 
the development of new~\ac{UAV} use cases for important industries.
Moreover, the~\ac{MRS} Drone was developed to be deployed within a~\acl{MRS} group, which can further increase its application potential.

\begin{figure*}[!htb]
    \centering
    \resizebox{1.0\textwidth}{!}{
    \input{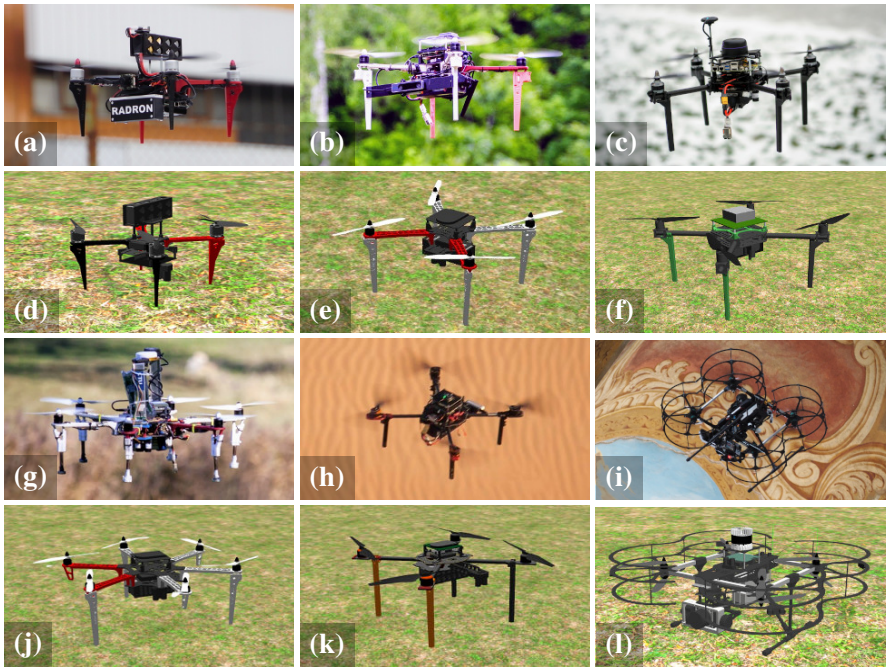}
    }
    \caption{\ac{MRS}~\ac{UAV} platforms used for a diverse range of real-world applications (a)-(c), (g)-(i) and their simulated variants (d)-(f), (j)-(l), respectively. 
    The numerous applications, ranging from aerial indoor inspection to multi-robot wall building, is enabled thanks to the high modularity of the~\ac{MRS} drone.}
    \label{fig:mrs_platforms}
    %\vspace{-1em}
\end{figure*}

%%% 5. Why hasn’t the problem been solved? What is the stumbling block?
Although many~\ac{UAV} platforms have been developed~\cite{Foehn22Agilicious}, there is still the need for a platform that is simultaneously modular, agile, and robust, while also having a sufficient flight time duration to perform complex tasks and the ability to process highly-demanding algorithms.
%%% 6. What does our paper contribute?
To this end, the~\ac{MRS} group in Prague\footnote{\url{http://mrs.felk.cvut.cz}} has, for the past seven years, developed the~\ac{MRS} Drone platforms shown in Fig.~\ref{fig:mrs_platforms}. 
%%%%%%%%%%%%%%%%%%%%%%%%
%%%%%%%%%%%%%%%%%%%%%%%%%%%%%%%%%%
%%% 8. What do the experiments say?
Our system has allowed more than $300$ bachelor's, master, and Ph.D. students from more than $100$ research groups worldwide to perform real-world experiments in indoor and outdoor conditions during the~\ac{MRS} summer schools held in 2019, 2020, 2022, and 2023\footnote{\url{http://mrs.felk.cvut.cz/summer-school-2023/}}. 
The modularity of the~\ac{MRS} Drone has been exploited in a large number of industrial applications, such as firefighting and drone hunting, as well as in robotic competitions, including~\ac{darpa}~\ac{subt} Challenge,~\ac{MBZIRC} 2017, and~\ac{MBZIRC} 2020.
With the~\ac{MRS} Drone platform, research groups and startups are able to build a multi-rotor~\acp{UAV} equipped with the~\ac{MRS} Drone~\ac{SW} system~\cite{baca2021mrs} on their own, or with the support of the modular DroneBuilder web page\footnote{\url{https://dronebuilder.fly4future.com}}.

The remainder of the paper is organized as follows.
An overview of the state-of-the-art~\ac{UAV} systems is presented in the rest of this section.
Section~\ref{sec:mechanicalDesignPrototyping} presents the mechanical design of the~\ac{MRS} Drone while Section~\ref{sec:electricalDesignAndConfiguration} presents the electrical design.
In Section~\ref{sec:MRSUAVSystem}, we overview the~\ac{MRS} Drone system architecture.
Finally, Section~\ref{sec:applications} presents the diverse applications of our platform, and Section~\ref{sec:conclusions} concludes the paper.

%%% END SECTION ============================================================

%%% START SECTION ==========================================================

\begin{figure}[h]
    \centering
    \includegraphics[width=\columnwidth]{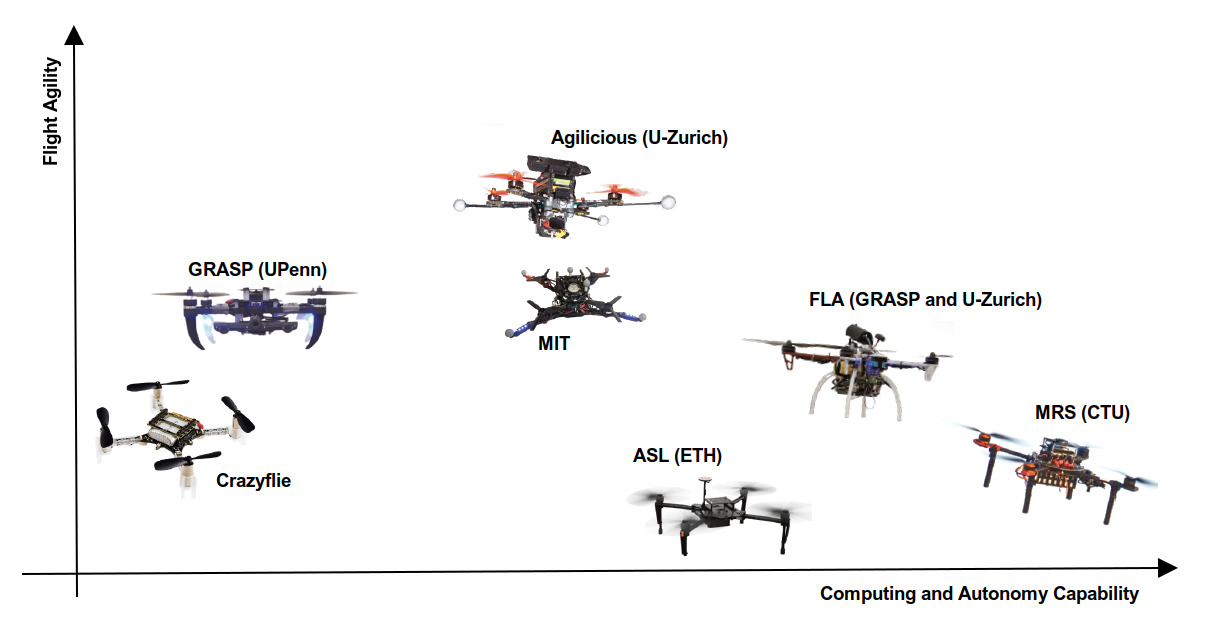}
    \vspace{-5mm}
    \caption{\ac{UAV} comparison of research platforms between the~\ac{MRS} (CTU) 
    platform~\cite{HertICUAS2022, baca2021mrs} and other existing platforms, as provided by an 
    independent study from the well-recognized robotic group of 
    U-Zurich~\cite{Foehn22Agilicious}.}
    \label{fig:comparison}
    \vspace{-1em}
\end{figure}

\subsection{Contributions beyond the State-of-the-art}
\label{sec:contributionsBeyondStateOfArt}

%%% 3. Why is the problem hard? What makes it challenging?
Developing a new~\ac{UAV} platform is not an easy task. 
State-of-the-art research aims to achieve a~\ac{UAV} that has both high flight agility and a high computational and autonomy capability (see Fig.~\ref{fig:comparison}). 
In addition, the large variety of applications demands different sets of sensors, which in turn changes the technical specifications and power demands of each drone. 
Moreover, some applications require the inclusion of different types of actuators, such as arm manipulators or even fire extinguishers~\cite{PatelRAM2022}.

Nowadays, novel concepts of autonomous helicopter platforms continue to develop~\cite{Stingu2009} as different designed platforms are required for new applications that appear every year. 
The design of new~\acp{UAV} continues to experience difficulties (e.g. long design periods, high manufacturing costs, and difficulty in performing hardware maintenance).
%Those are the main obstacles to more frequent verification of scientific achievements in real conditions for which they are intended. 
A recent work~\cite{Guo2021} presented a novel methodology to design lightweight and maintainable frames of micro aerial vehicles by using a configurable design. 
%proposed a design method to obtain a lightweight and maintainable~\ac{UAV} frame using a configurable design. 
Many other aspects were also investigated in recent years such as the cost reduction for UAV construction and maintenance \cite{Flynn2013}, multi-robot systems design with commercial benchmark platforms \cite{Wang2021}, and the experimental hexacopter platform design and construction \cite{SchachtRodriguez2018}.
%The work of Flynn~\cite{Flynn2013} focused on the aspect of reducing cost for building and maintaining~\ac{UAV} platforms. Wang et al.~\cite{Wang2021} focused on designing multi-robot systems using commercial benchmark platforms, while the work of Schacht-Rodriguez et al.~\cite{SchachtRodriguez2018} addressed the design, construction, and instrumentation of a~\ac{UAV} hexacopter experimental platform. 
%%% 4. How far has existing work come? What is the frontier?
Furthermore, several~\ac{UAV} research groups worldwide have proposed open-source platforms.
A summary can be found in a recently published paper from Foehn et al.~\cite{Foehn22Agilicious} that presents the most prominent~\ac{UAV} platforms from research groups in the field.
In particular, the most recent one from D. Scaramuzza's group from U-Zurich~\cite{Foehn22Agilicious} has the best balance between onboard computational capability and agility. 
There are also platforms from ASL (created by R. Siegwart's group at ETH)~\cite{Siegwart2018} and FLA (a joint platform from GRASP and U-Zurich)~\cite{Kartik2018} that have relatively high weights and low agility. 
On the other hand, the relatively small Crazyflie~\cite{Wojciech2017, Silano2019ROSVolume4, SilanoMED18} platform does not allow for sufficient onboard computation or sensing. 
Finally, the platforms from S. Karaman's group from MIT~\cite{Karaman2020} and GRASP (created by V. Kumar's group from UPenn)~\cite{Kumar2017} do not provide open-source software and hardware.
Many proposed platforms try to create general-use UAV hardware for research purposes. Nevertheless, none of the above-mentioned ones have been repeatedly tested in real-world environments in such a wide range of applications as the~\ac{MRS} Drone.
Thus, these UAV platforms do not facilitate the desired minimization of the reality gap between the experiments and real-world deployment.
Moreover, as shown in Fig.~\ref{fig:comparison}, the~\ac{MRS} Drone has the best computing and autonomy capability, which allows for high modularity in diverse applications. 

% Related Works

The years of field experience gained by the~\acf{MRS} group has provided the foundation for the state-of-the-art research that has resulted in the proposed~\ac{MRS} Drone system, which in turn has yielded dozens of publications by numerous distinct research groups.
Our platforms were used in various demanding robotic applications, such as remote sensing, aerial manipulation, aerial pick-and-place tasks, marine surveillance, and autonomous wall building, as described in Section~\ref{sec:applications}.
Such uses have demonstrated the high modularity and versatility of our platform.
%%% 7. What is the key idea? What is the magic trick? What is the new insight or technique that enables us to advance the frontier?
The~\ac{HW} and Software~\ac{SW} system parts of the~\ac{MRS} Drone have allowed researchers worldwide to perform their experiments in real-world conditions, which in turn has allowed us to improve our platform over time. 
The key features of the~\ac{MRS} Drone include its modular~\ac{HW} construction, 
open-source~\ac{SW} and plugin-based architecture, and high computational and autonomous 
capabilities. The~\ac{MRS} software system has been described in our past 
works~\cite{baca2021mrs, HertICUAS2022}.
Furthermore, the~\ac{MRS} system provides an actively maintained and well-documented implementation on GitHub\footnote{\url{https://github.com/ctu-mrs/mrs_uav_system}}, including a realistic~\ac{UAV} ROS-gazebo-based simulation, a large variety of sensor setups, diverse~\ac{UAV} control strategies, and a robust multi-modal localization system.

%%%% See if we move that up...
% This is the first time the hardware part of the co-designed MRS-drone is detailed.
% The contributions of this work are considered as follows.
% We propose a modular~\ac{UAV} platform that can be used in various applications with distinct actuator and sensor configurations while minimizing the effort required for changing the~\ac{UAV} setup and for its maintenance. 
% This is an important aspect mainly in the initial stages of experimentation with new approaches due to a high probability of collisions.
% We extend the initial explanation of our \ac{UAV} platform from~\cite{HertICUAS2022}.
% We propose a platform that facilitates the transition from simulation and simplified  laboratory  experiments, into  the deployment of aerial robots in real-world conditions with a minimal sim-to-real gap.
% Finally, we designed the system with the intent of supporting the initial steps of researchers and students from different scientific areas, in which the~\acp{UAV} is a necessary tool for experimental validation of proposed concepts, with the goal of enlarging the community of active users of fully autonomous~\acp{UAV}.

In contrast to our previous work~\cite{HertICUAS2022}, this is the first time the hardware aspect of the co-designed~\ac{MRS} Drone has been detailed with its experimental hardware results, and its applications in numerous different environments expanded upon. Therefore, the contributions of this work are considered as follows:
\begin{enumerate}
    \item We propose a modular UAV platform that can be used in various applications with distinct actuator and sensor configurations while minimizing the effort required for maintenance and substitution of broken parts. This feature is especially appealing due to the fact that real robot experiments have a high probability of collisions, mainly in the initial stages of experimentation with new approaches, and often need maintenance.
    \item We present an analysis of the results of experiments that aim to test the propulsion systems of our~\acp{UAV}~\ac{wrt} vibration during flight.
    \item Our proposed platform facilitates the transition from simulation and simplified laboratory experiments to the deployment of aerial robots in real-world conditions with a minimal sim-to-real gap.
    \item We expand upon the applications of our modular~\ac{UAV} platforms, including subterranean environments, package delivery, marine environments, human-\ac{UAV} interaction, and so on.
    \item Finally, with the goal of enlarging the user community of fully autonomous-\ac{UAV}, we designed the system with the intent to support the initial steps of researchers and students from different scientific areas where \ac{UAV} are a necessary tool for the experimental validation of proposed concepts.
\end{enumerate}

%%% END SECTION ============================================================

%%% START SECTION ==========================================================

\section{Mechanical Design and Prototyping}
\label{sec:mechanicalDesignPrototyping}

In this section, we present the mechanical design of our proposed platforms.
We discuss the frame and propulsion system selection, the design of 3D printed parts to support the mechanical enhancement of the~\ac{UAV}, the \ac{UAV} sensor selection, multi-robot sensor selection, the communication technologies used (e.g.,~\ac{UV} cameras), and the actuators used for specific applications.

%%% END SECTION ============================================================

%%% START SECTION ==========================================================

%%[OWNER]: Daniel Hert
\subsection{Frames and propulsion system consideration}
\label{sec:framesPropulsionSystemConsideration}

UAV frames are an essential part of the aircraft. The correct choice of frame results in the size of the aircraft and the maximum diameter of the propellers it must use, which in turn affects the resulting payload the UAV will be able to carry and the endurance it will have.
%The frame is the fundamental component of each~\ac{UAV}, as it determines its final size and the maximum diameter of its propellers, which affect the overall payload, endurance, and enabled set of sensors and actuators. 
The propeller size and motor choice define the maximum thrust and, therefore, the maximum payload of the~\ac{UAV}. The propeller produces thrust in one direction by accelerating air in the opposite direction. In general, to produce a defined amount of thrust, it is more efficient to accelerate a larger mass of air by a smaller amount, than to accelerate a small amount of air by a greater amount. This means that to produce the same amount of thrust, a larger propeller spinning at a slower speed is more efficient than a smaller propeller spinning at a faster speed~\cite{HamandiIJRR2021}. In Table~\ref{tab:motor_tests}, we compare the performance of two propellers (one 8-inch and one 9.4-inch) with the same motor and~\ac{ESC}. It is evident that the larger propeller is more efficient and produces higher maximum thrust.

The design of UAVs capable of longer flights and able to carry high payloads requires larger propellers. In contrast, the smaller the UAVs are, the easier to operate they are. The small size of UAVs enables the to fly in obstacle-crowded environments. In the end, there is always a compromise between the UAV size and the propulsion efficiency. 
%smaller platforms are easier to transport and operate, and their smaller size is more suited for environments with obstacles. When choosing the frame size, a compromise must be made between the final size of the platform and its propulsion efficiency.
The standard~\ac{MRS} platforms are based on three basic frame sets and are suitable for most of our research experiments and tests. Some tasks require unusual designs and capabilities that cannot be accommodated by the standard frames. Therefore, custom task-specific frames are also presented.

\begin{table}[tb]
  \centering
  \sisetup{per-mode=symbol}
   %\captionsetup{width=0.8\textwidth}
  \caption{Motor and propeller tests were done on a static thrust-measuring stand. Both propellers were tested with the same Readytosky 2312 920KV and a MultiStar BLHeli32 51A~\ac{ESC}.
       \vspace{-1mm}
       }
  \resizebox{\columnwidth}{!}{%
  \begin{tabular}{l l l l l l l}
    \toprule
    % source - https://ctu-mrs.github.io/docs/hardware/motor_tests.html
    \textbf{8045 propeller} \vspace{0.5em} \\
    Throttle (\%)   & Thrust (N)    & RPM   & Voltage (V)   & Current (A)   & Power (W) & Efficiency (g/W)  \\\midrule
    50              & 3.23          & 5939  & 16.70         & 2.73          & 45.59     & 7.08              \\
    60              & 4.29          & 6902  & 16.66         & 4.17          & 69.47     & 6.18              \\
    70              & 5.42          & 7648  & 16.59         & 6.06          & 100.54    & 5.39              \\
    80              & 6.63          & 8499  & 16.52         & 8.22          & 135.79    & 4.88              \\
    90              & 7.81          & 9076  & 16.45         & 10.50         & 172.73    & 4.52              \\
    100             & 8.91          & 9722  & 16.37         & 13.27         & 217.23    & 4.10              \\\midrule
    \textbf{9450 propeller} \vspace{0.5em} \\
    Throttle (\%)   & Thrust (N)    & RPM   & Voltage (V)   & Current (A)   & Power (W) & Efficiency (g/W)  \\\midrule
    50              & 4.01          & 5688  & 16.66         & 3.09          & 51.48     & 7.79              \\
    60              & 5.29          & 6508  & 16.61         & 4.80          & 79.73     & 6.64              \\
    70              & 6.56          & 7213  & 16.54         & 6.72          & 111.15    & 5.90              \\
    80              & 7.87          & 7788  & 16.48         & 9.05          & 149.14    & 5.28              \\
    90              & 9.11          & 8608  & 16.40         & 11.48         & 188.27    & 4.84              \\
    100             & 10.24         & 9068  & 16.32         & 14.26         & 232.72    & 4.40              \\
    \bottomrule
  \end{tabular}
  }
  \label{tab:motor_tests}
  \vspace{-1.5em}
\end{table}

%%% END SECTION ============================================================

%%% START SECTION ==========================================================

%%[OWNER]: Daniel Hert
\subsection{MRS Drone used frames}
\label{sec:basicFrames}

The frames that are used to support the majority of the recent works have similar construction patterns that use four arms positioned between a central structure. In general, the arms are made from plastic or carbon fiber, while the central structure is made from glass~\ac{REL} or carbon fiber. Carbon fiber composite is preferred as a frame material, as it offers great stiffness while remaining lightweight. Plastic parts are heavier and not as stiff as carbon fiber, but they are also much cheaper. The top or bottom structural boards can be replaced with a custom-made printed circuit board containing the electronics necessary for the functionality of the~\ac{UAV}. Integration of electronics directly to the structural board of the frame reduces the amount of additional electronic modules, thereby reducing the mass and complexity of the~\ac{UAV}.

\begin{figure}[tb]
    \centering
    \includegraphics[width=0.60\columnwidth]{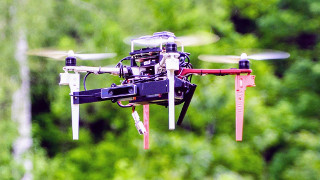}
    \caption{DJI F450 frame used for swarming research.}
    \label{fig:f450drone}
\end{figure}

The first and smallest basic frame is the DJI F450, shown in Fig.~\ref{fig:f450drone}, equipped with 2212 KV920 motors and plastic 9.4-inch propellers. Plastic propellers are not as stiff and efficient as carbon fiber composite propellers, but they are more damage-resistant. This makes them suitable for a smaller and cheaper~\ac{UAV}. The platform is powered by a 4S 6750\,mAh lithium polymer battery. This combination offers about 0.5\,kg of usable payload and flight times between 10-15 minutes, depending on the payload. This platform is the smallest and cheapest out of the three standard frames and is therefore ideal for swarming research, where a large number of~\acp{UAV} are required. It is also employed in experiments that do not require larger payloads. The 4S 6750\,mAh battery has a capacity of 99.9\,Wh, which is convenient for air transport, as most airlines limit the size of a battery that can be carried without any special provisions to 100\,Wh.

\begin{figure}[tb]
    \centering
    \includegraphics[width=0.60\columnwidth]{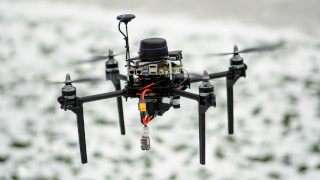}
    \caption{Holybro X500 frame with upgraded motors.}
    \label{fig:x500drone}
\end{figure}

The mid-sized Holybro X500 frame, showcased in Fig.~\ref{fig:x500drone}, is the preferred option. Its 3510 KV700 motors and 13-inch carbon fiber composite propellers offer exceptional high thrust and propulsion efficiency. The drone can be powered by either one or two 4S 6750 mAh lithium polymer batteries. To increase the payload capacity and flight duration, our X500 has an enhanced propulsion system compared to Holybro's default kit (2216 KV920 motors with 10-inch propellers). The X500 can carry a usable payload of up to 1.5 kg. If a more extended flight time is necessary, an additional battery can be connected in parallel, enabling over 20 minutes of flight time even with a full payload. Additionally, the individual batteries are under 100 Wh, making it easy to transport by plane. Due to its compact design, high payload capacity, and long flight duration, this platform was used in the~\ac{darpa}~\ac{subt} Challenge. Moreover, it achieved the best performance in the~\ac{darpa}~\ac{subt} Virtual Challenge, where all teams occupying the first six places in the final competition used the~\ac{MRS} X500. In summary, this platform's compact design, high payload capacity, and propulsion efficiency make it an excellent choice for long flights, indoor research, and carrying heavier sensors.
%The mid-size selected frame is the Holybro X500, shown in Fig.~\ref{fig:x500drone}. It is equipped with 3510 KV700 motors and carbon fiber composite 13-inch propellers, which provide a great combination of high thrust and propulsion efficiency. The~\ac{UAV} is powered by either one or two 4S 6750\,mAh lithium polymer batteries. The propulsion system of our X500 is upgraded over the default kit provided by Holybro (2216 KV920 motors with 10-inch propellers) in order to increase the payload capacity and flight time. Our X500 can carry 1.5\,kg of usable payload. If longer flight times are required, a second battery can be connected in parallel. This enhances the flight time to over 20 minutes even with a full payload, while still keeping the individual batteries under 100\,Wh, allowing for easy transport by plane. This platform was used in the~\ac{darpa}~\ac{subt} Challenge, as it combines compact size, large payload, and long flight time. In addition to the real~\ac{darpa}~\ac{subt} Challenge, this platform also achieved the best performance in the~\ac{darpa}~\ac{subt} Virtual Challenge, where the~\ac{MRS} X500 was used by all teams occupying the first six places in the final competition. In general, this platform combines compact design with high payload and propulsion efficiency. Therefore, it is suitable for long flights, indoor research, and carrying heavier sensors.

The Tarot T650, displayed in Fig.~\ref{fig:x500drone}, is the largest frame chosen for the project. It features 4114 KV320 motors, carbon fiber 15-inch propellers, and a 6S 8000 mAh lithium polymer battery. The T650 is the heaviest standard~\ac{MRS} platform with a payload capacity of 2.5 kg.
%The largest selected frame is the Tarot T650, shown in Fig.~\ref{fig:x500drone}. It is equipped with 4114 KV320 motors, carbon fiber 15-inch propellers, and a 6S 8000\,mAh lithium polymer battery. This is the heaviest standard~\ac{MRS} platform with a payload capacity of 2.5\,kg. 
It offers a higher payload and a notably bigger payload volume than the X500, which can be required for some applications. It was used in the~\ac{MBZIRC} 2020 competition, where large payloads and clearances were required for carrying heavy bricks~\cite{baca2020autonomous}, a ball-catching net~\cite{vrba_ras2022}, and bags of water for extinguishing fires~\cite{walter2020fr}. 
%This platform is also used in most of the projects that require~\ac{UAV} manipulation, as manipulators are usually relatively heavy, and they require more clearance.

\begin{figure}[tb]
    \centering
    \includegraphics[width=0.60\columnwidth]{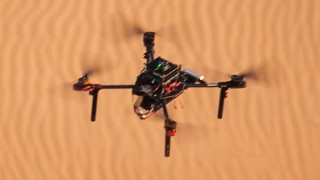}
    \caption{Tarot T650 frame.}
    \label{fig:t650drone}
\end{figure}

%%% END SECTION ============================================================

%%% START SECTION ==========================================================

%%[OWNER]: Pavel Stoudek
\subsection{Custom platforms for specific applications}
\label{sec:taskSpecificCustomPlatforms}

In some cases, the use of a basic frame is not ideal. For instance, drones with higher payloads require bigger frames, sometimes being preferable even a non-square frame shape. A specific setup is also useful for physical interaction with the environment, or due to some special end effector, such as a gripper, magnets, or hooks.

\begin{figure}[tb]
    \centering
    \includegraphics[width=0.60\columnwidth]{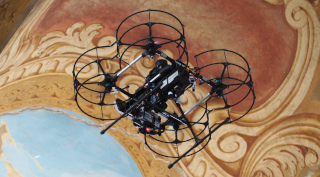}
    \caption{\ac{UAV} platform developed for the Dronument project.}
    \label{fig:nakidrone}
\end{figure}

To provide an example, the Dronument project\footnote{\url{http://mrs.felk.cvut.cz/dronument}} utilizes a high-payload coaxial octocopter for autonomous inspection of historical buildings~\cite{kratky2021documentation, petracek2020dronument}. This platform is meant to carry a regular-size camera with a dual-axis gimbal stabilization, OS0-128 high-resolution~\ac{lidar}, ultrasonic sensors, and a custom carbon fiber safety cage to protect the platform and its surroundings. Moreover, the platform size must be minimized to allow for flight in confined environments. To maximize the thrust, coaxial propulsion was used. The drone dimensions stay reasonable, while the maximum thrust is increased and, therefore, the maximum payload. However, the coaxial propulsion setup is less efficient, with~20\% more power~\cite{Bondyra2016PerformanceOC} required to generate the same thrust as a standard propulsion dual motor setup. The platform is shown in Fig.~\ref{fig:nakidrone}.

A platform based on similar requirements and principles was used for extinguishing fires on the above-ground floors~\cite{spurny2020autonomous} of buildings, utilizing a unique pneumatic launcher that was able to discharge a fire-suppressing capsule propelled with compressed CO$_2$ gas. The capsule (containing water and fire suppressant substances) was able to break through a window of a burning building and extinguish a fire inside. The entire frame was built around the design requirement imposed by the long shape and heavy weight of the launcher. Similar to~\cite{kratky2021documentation, petracek2020dronument}, coaxial propulsion was also used for this task in order to minimize the size of the~\ac{UAV}. This platform is shown in Fig.~\ref{fig:dofec_2}.

A prototype drone hunter~\cite{stasinchuk2020multiuav} for protected no-fly zones was based on a T18 frame by Tarot (see Fig.~\ref{fig:eagle}). This octocopter platform with 18-inch propellers had up to 10\,kg of payload, which was required for Ouster~\ac{lidar}, a Reach~\ac{RTK} module, and a special folding net to capture and carry other drones. After the proof of concept with the commercial frame, the team designed a fully custom platform made out of carbon fiber and CNC-machined aluminum. 

\begin{figure}[tb]
    \centering
    \includegraphics[width=0.55\columnwidth]{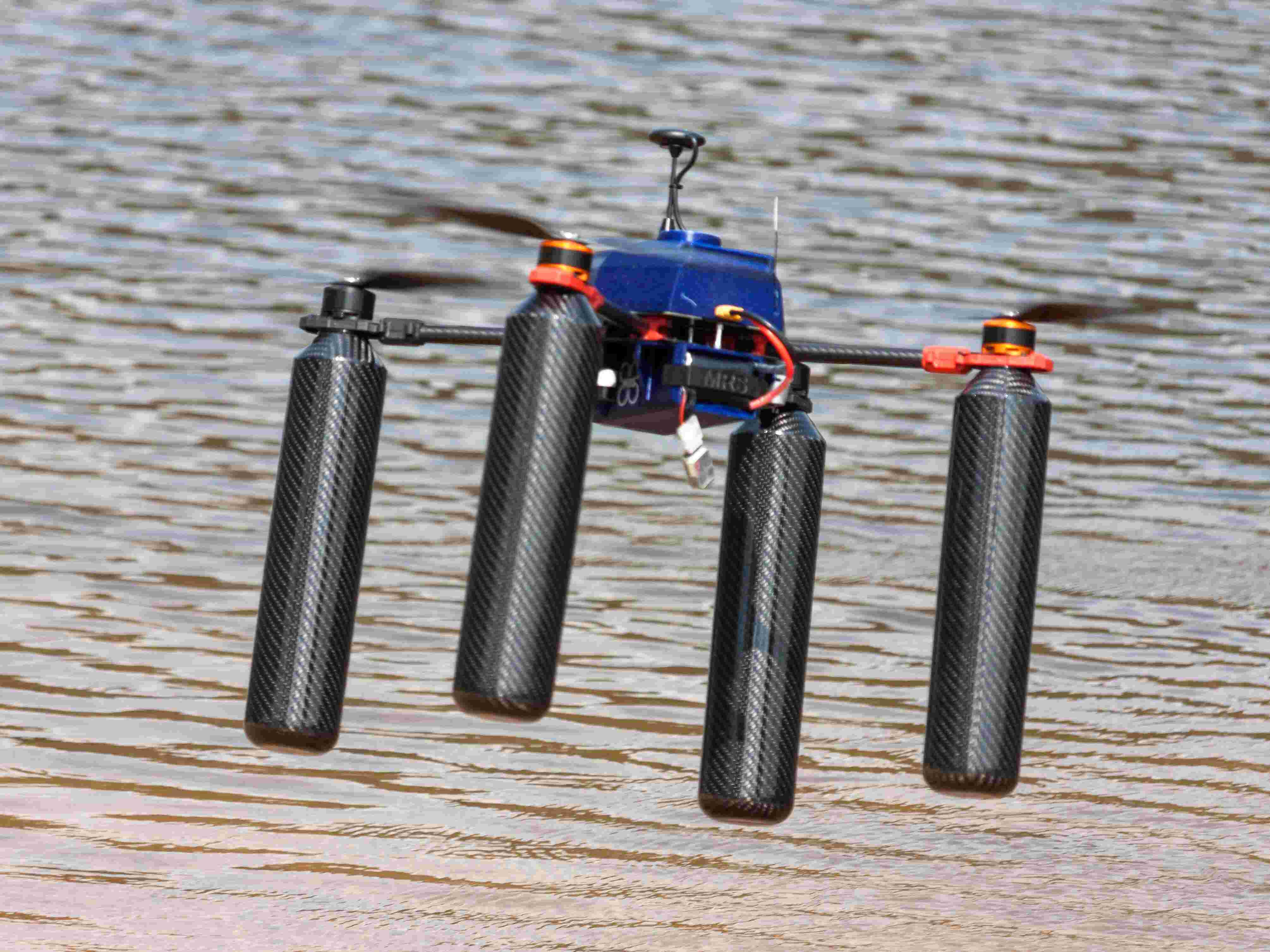}
    \caption{Waterproof T650 platform with carbon fiber floaters.}
    \label{fig:marinedrone}
\end{figure}

A specially modified T650 platform is used for experiments involving flight over water surfaces. This~\ac{UAV} is equipped with a waterproof shell and custom carbon fiber floaters. All electronics are located inside the shell and are thus protected from water. The floaters are required for emergency landings on water and for safe retrieval of the~\ac{UAV}. This platform is used to follow and land on a~\ac{USV}, as well as to detect objects on the water's surface. The motivation of this research is to address water pollution using drones, and a waterproof blue robotics gripper will be mounted on the platform in the future. The~\ac{UAV} can be seen in Fig.~\ref{fig:marinedrone}.

%%% END SECTION ============================================================

%%% START SECTION ==========================================================

%%[OWNER]: Pavel Stoudek
\subsection{Prototyping via 3D printing}
\label{sec:3DprintedPrototyping}

%[TO BE WRITTEN] Only few words
%Legs that can break easily protect the important parts. You can print a spare one even in the field.
%Possibility of changing camera orientation even during the experiments by printing a new holder.
%You can put some links to SW for 3D designing. Hints for beginners.
%\textbf{TODO: Pavel St}}
The technology of additive manufacturing quickly found its place in the design process. Every single one of our drone platforms has custom parts made by 3D printing. The parts are designed from the start to comply with the constraints of additive manufacturing. Filaments from PLA and PETG are mostly used, although they can be swapped for more resilient materials like ABS, ASA, and PC. The prototyping phase is quick, even by fast prototyping standards, using cheaper filaments and low resolution, while the final print of a polished part takes more time. The parts can be easily replicated and mounted on the platform, even during experimental campaigns outside of the laboratory. Moreover, designing unique frame-sensor combinations and modifying sensory attachments to adapt to the ever-changing requirements is a fast and simple task, even during ongoing or onsite research.

Some parts made by additive manufacturing include for example,~\ac{UV}-light holders, RGB camera mounts, \ac{lidar} mounts, onboard computer covers, and battery holders. During the design, stress simulation, and shape optimization are used to achieve the best strength-to-weight ratio for the part. This is discussed more in Section~\ref{sec:vaseMode}.

A crucial component of the~\ac{UAV} is its set of custom legs. They are optimized to be strong enough to support the platform, while simultaneously being as light as possible. In addition, they are intended to be used as modular holders for primarily sensory equipment. For illustration, a Basler camera and LED lights were mounted on 3D printed legs to provide lighting for better image capture~\cite{saska2017documentation, kratky2021exploration}. In case a different size is needed, the leg's 3D model can be easily modified and extended, even during experimental campaigns. Another important feature of the leg design as motivated by experience from experiments is the safety of weak spots. In case of an emergency landing, a leg should break at a specific point to absorb the impact energy and, thus, protect the core of the platform. The broken leg can then be replaced in a matter of minutes. While testing novel research concepts that are, in principle, unreliable, these kinds of modifications proved to be crucial in practice.

\begin{figure}[tb]
    \centering
    \input{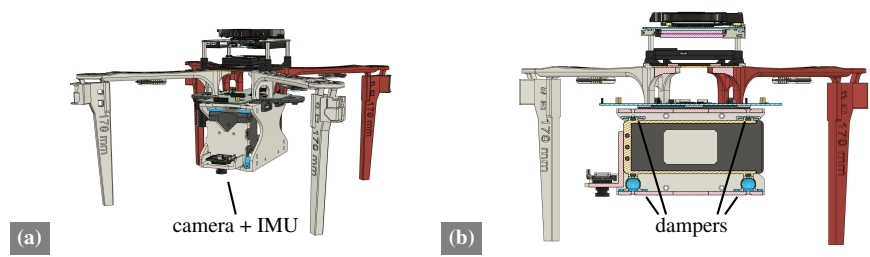}
    \vspace{-5mm}
    \caption{F450 \ac{UAV} platform with 3D printed vibration dampers between the battery holder and the rest of the \ac{UAV} for visual-inertial odometry applications: (a) 3D view, (b) cross-section with highlighted dampers.}
    \label{fig:vio_damping}
    \vspace{-1em}
\end{figure}

\begin{figure}[h]
    \centering
    % left - bottom - right - top
    \includegraphics[trim={1.4cm 0cm 0.2cm 0.25cm},clip, width=0.7\columnwidth]{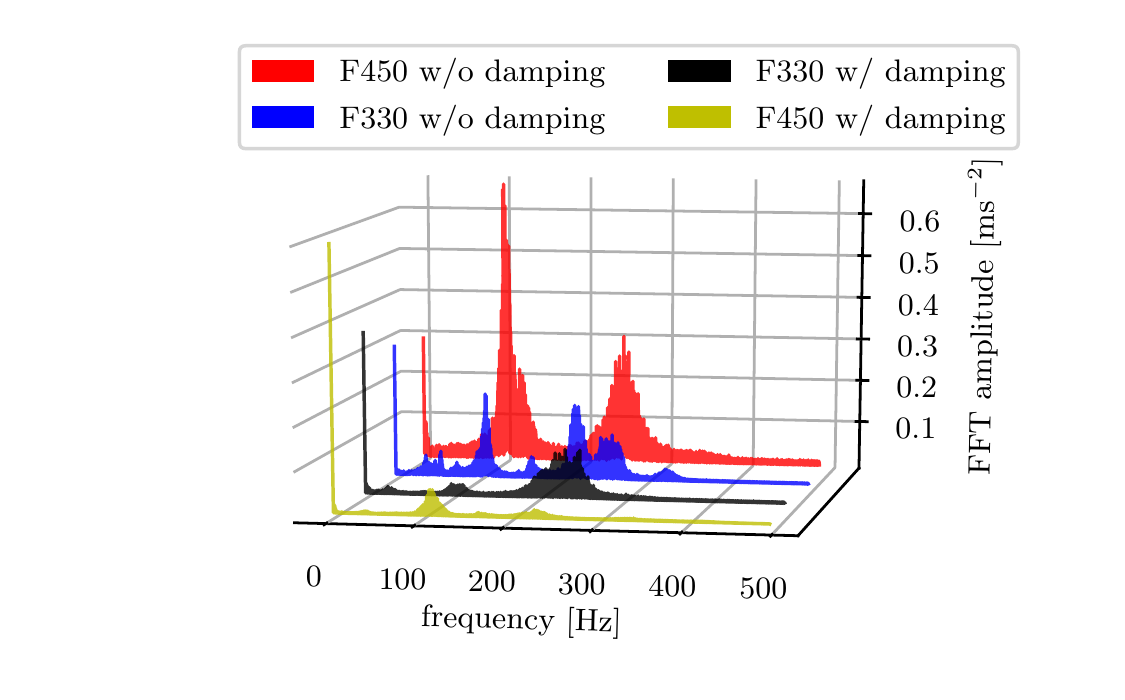}
    \vspace{-5mm}
    \caption{Spectra of vibrations measured in the $x$-axis (pointing to the front of the~\ac{UAV}) of the accelerometer mounted on the F450 and F330 platforms with and without damping. The damped F330 platform was equipped with rubber-based dampers between the top-mounted battery and the rest of the~\ac{UAV}, while the F450 platform was equipped with 3D-printed dampers. In both cases of damping, the IMU was mounted on the battery case, which served as added mass for improving the damping performance.}
    \label{fig:vio_vibrations}
    \vspace{-1em}
\end{figure}

3D-printed dampers and bushings from flexible filaments can be utilized to avert the unwanted vibrations that are ever-present with a multi-rotor~\ac{UAV}.
Figure~\ref{fig:vio_damping} shows the F450 platform equipped with a monocular camera and an IMU for~\ac{VIO}.
The sensors are mounted on a 3D-printed battery case, with the battery serving as an added mass for improving the damping performance.
The battery case is placed on a set of dampers printed out of the TPU 30D filament.
Such an approach significantly decreases the vibrations present in the IMU data and enables reliable usage of~\ac{VIO}, whose performance is greatly affected by the propeller-induced vibrations~\cite{Bednar2022ICUAS, Pritzl2022ICUAS}.
Moreover, a 3D-printed battery holder with 3D-printed bushings can have a variable position based on the appliances attached and a varying center of gravity of the drone.
Figure~\ref{fig:vio_vibrations} shows a comparison of vibration spectra during flight with and without damping, measured using the ICM-42688 IMU.
 The IMU was mounted on the damped part of the \ac{UAV} and measured accelerometer data at the rate of \SI{1}{kHz}.
In the case of the F450 platform, the IMU was mounted on the 3D-printed battery case (see Figure~\ref{fig:vio_damping}), and 3D-printed dampers from the TPU 30D filament were used.
In the case of the F330 platform, the IMU was attached to the top-mounted battery holder (see Figure~\ref{fig:mrs_platforms}a), and rubber-based dampers between the battery holder and the \ac{UAV} frame were utilized.
Peaks corresponding to the propeller rotation frequency and its second harmonic frequency are clearly visible, and their amplitudes are significantly reduced by damping.

%%% END SECTION ============================================================
%%% START SECTION ==========================================================

%%[OWNER]: Martin Sramek
\subsection{Using vase mode for strong lightweight parts}
\label{sec:vaseMode}

\begin{figure}[tb]
%\vspace{0.3cm}
\centerline{\includegraphics[width = 0.50\textwidth]{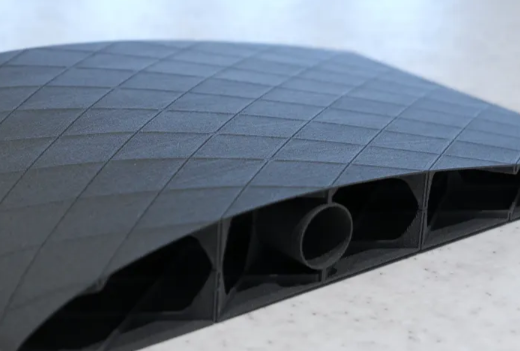}}
\caption{Cross section of a plane wing printed in vase mode with inner struts formed by incisions.}
\label{fig:vaseWing}
\end{figure}

Further optimization regarding weight must be done; one way to do so is to use lighter and stronger materials, such as aluminum or carbon fiber. Unfortunately, these materials are either expensive or hard to manufacture, which makes them unsuitable for the iterative design process. When using 3D printing, we can utilize the so-called vase mode, where only the single outer perimeter of the model is printed. If parts are designed correctly, a high strength-to-weight ratio can be achieved. Also, the absence of filament retractions makes the prints look cleaner and, in combination with the absence of the travel behavior, moves faster. The general idea is that thin incisions are extruded to the basic shape, which then forms the vase perimeters. The incisions should be thin enough that the resulting perimeters melt into each other and form a wall, as seen in Fig.~\ref{fig:vaseWing}. This method was originally pioneered by Tom Staton on his YouTube channel\footnote{\url{https://youtu.be/QJjhMan6T_E}}.

%%% END SECTION ============================================================
%%% START SECTION ==========================================================

%%[OWNER]: Daniel Hert, Daniel Bonilla, Viktor Walter

\subsection{Sensory equipment}
\label{sec:sensoryEquipment}

The mass of a sensor is critical in the selection process, as it dictates the payload capacity of the~\ac{UAV}.~\acp{lidar} show a significant change in capabilities as their mass increases. One of the lightest~\ac{lidar}, the Garmin~\ac{lidar}-Lite, is just a 1D sensor, which is used to measure the~\ac{UAV}'s altitude on most~\ac{MRS} platforms. RP\ac{lidar} A3 is heavier, but it provides much more data, being a 2D~\ac{lidar} that produces a planar \SI{360}{\degree} scan of the surrounding environment. The heaviest~\acp{lidar}, which is regularly used on the~\ac{MRS} platforms, is the 3D multi-beam~\acp{lidar} manufactured by Ouster or Velodyne. These sensors produce a high-resolution \SI{360}{\degree} scan in multiple planes. The heavier sensors provide much more useful data, but they are also much more expensive, and the~\ac{UAV} platform has to accommodate them with higher payload capacity.

RGB and RGB-D cameras are very attractive, as they are usually very lightweight and produce a lot of useful data. Depth cameras, like the Intel RealSense\footnote{\url{https://www.intelrealsense.com/depth-camera-d435/}}, are often used for navigation or to detect obstacles in the path of the~\ac{UAV}. These cameras use onboard processing to calculate the depth image from the stereo camera pair, easing the load on the main mission computer. Apart from the stereo pair, Intel RealSense also integrates a regular RGB camera. However, high bandwidth cameras, which use the USB-3 bus, provide strong interference with the signals used by GNSS. Special care has to be taken to shield the GNSS antennas properly.

Cheaper cameras often come with rolling shutters, which cause image distortions due to vibrations and fast movements of the~\ac{UAV}. These effects are detrimental to most computer vision algorithms. To counteract them, a camera with a global shutter can be used. The resolution and frame rate of the camera must also be considered. Processing images with higher resolution can yield better results, but it also requires more computational resources and causes larger processing delays, which can be critical to the performance of the entire system.

Other sensors include sonars, which can be used to measure the altitude of the \acp{UAV} above water, where other sensors may fail. Many other sensors for specific phenomena can be used for autonomous flight in special conditions and applications; for example, infrared and thermal cameras can be used to detect objects which are colder or hotter than the rest of the background, RGB cameras with a custom UV-pass filter can detect neighboring drones equipped with~\ac{UV} LEDs in~\ac{UAV} teams, TimePix radiation sensors can be used for radioactive objects tracking~\cite{baca2018timepix}, and gas detectors can sample air quality and deduce toxicity. 

Sensors that are used onboard the~\ac{MRS}~\ac{UAV} platforms are as follows:

\begin{itemize}
    \item \textit{Rangefinder}: Garmin~\ac{lidar} Lite V3, MB1340 MaxBotix ultrasound, 
    \item Planar~\ac{lidar}: RP\ac{lidar} A3,
    \item \textit{3D~\ac{lidar}}: Ouster OS1 and OS0 series, Velodyne VLP16,
    \item \textit{Cameras}: Basler Dart daA1600, Bluefox MLC200w (grayscale or RGB),
    \item \textit{RGB-D cameras}: Intel RealSense D435i and D455,
    \item \textit{GNSS}: NEO-M8N and~\ac{RTK} Emlid Reach M2,
    \item \textit{Thermal camera}: FLIR Lepton, FLIR Boson
    \item \textit{Pixhawk sensors}: gyroscopes, barometers, accelerometers (also available as separate sensors with better performance),
    \item \textit{High-rate IMU}: ICM-42688,
    \item \ac{UVDAR} system~\cite{walter2019uvdar}.
\end{itemize}

Every sensor has unique properties and disadvantages and is suited to different conditions and environments. The~\ac{MRS} system can fuse measurements from multiple different sensors to achieve reliable localization and navigation, even in environments where a singular sensor would be insufficient. 

\begin{figure}[tb]
    \centering
    \includegraphics[trim={0mm 40mm 0mm 110mm},clip,width=\columnwidth]{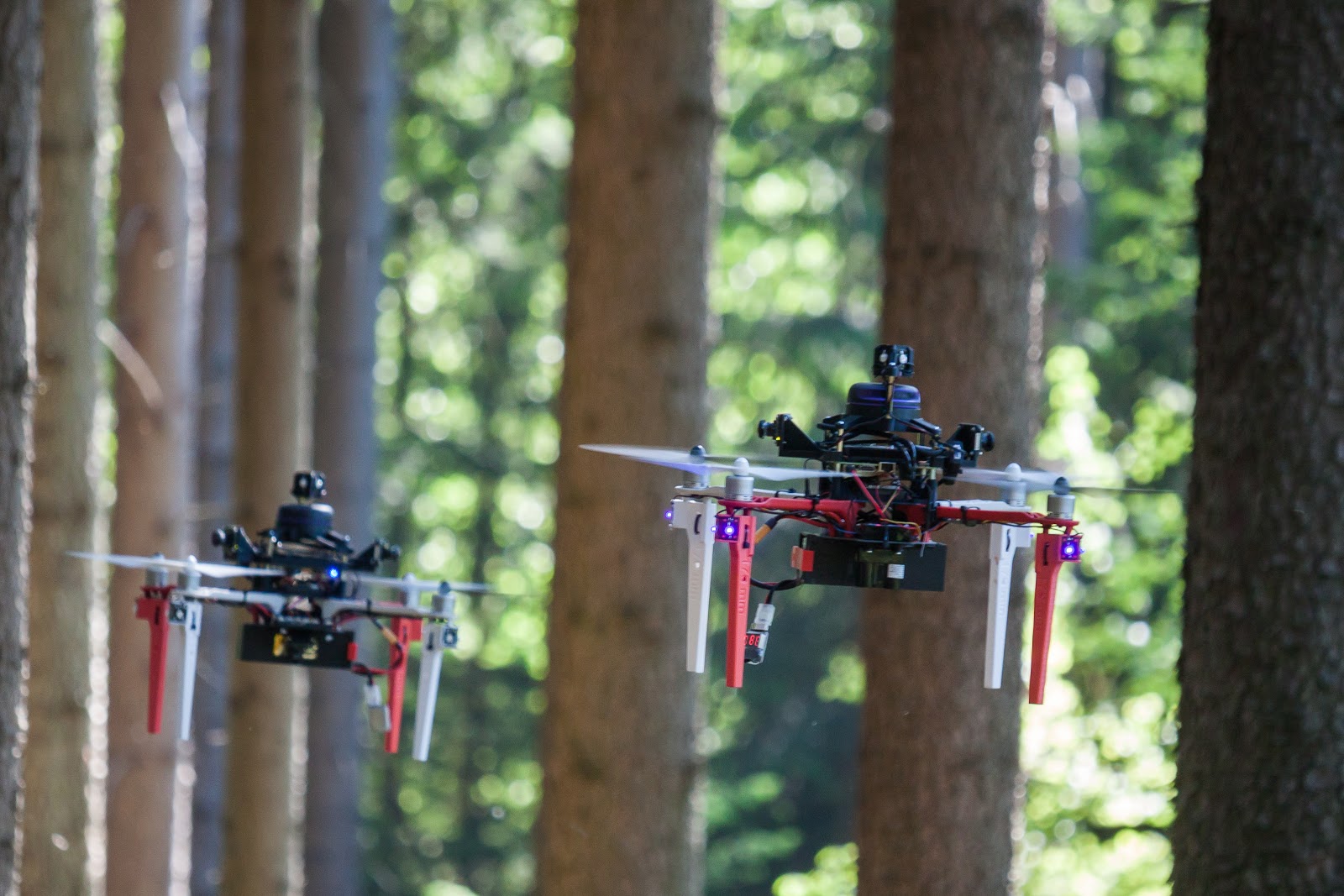}
    \vspace{-5mm}
    \caption{Pair of DJI F450-based~\ac{UAV} platforms equipped with the~\ac{UVDAR} system. Note, the ultraviolet LEDs and dual cameras on the sides of the~\ac{UAV} body.}
    \label{fig:uvdar_hw}
    \vspace{-1em}
\end{figure}

\begin{figure}[tb]
    \centering
    \includegraphics[trim={0mm 40mm 60mm 40mm},clip,width=\columnwidth]{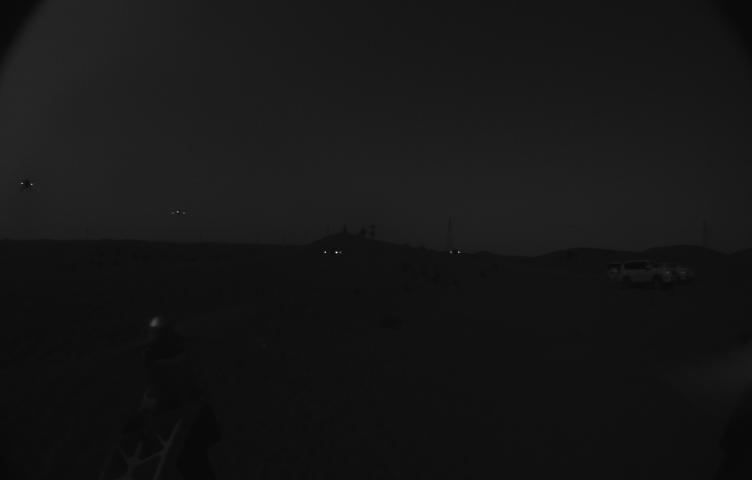}
    \vspace{-5mm}
    \caption{View from onboard ultraviolet-sensitive cameras used in the~\ac{UVDAR} system. Note, the clarity of the onboard LED markers of neighboring~\acp{UAV}, despite this image being captured at midday in a desert setting.}
    \label{fig:uvdar_hw2}
    \vspace{-1em}
\end{figure}

In order to successfully deploy multiple UAVs, the MRS platforms must have the ability to fly and cooperate with one another, even in cases where direct radio communication is unreliable or external localization systems are unavailable. In order to achieve this cooperation, the swarm or formation agents must be able to locate each other using their onboard sensory equipment. While computer vision is a cutting-edge technique for mutual relative localization of robots~\cite{swarm_survey, vrba2019onboard, vrba2020markerless}, its current implementations often suffer from degraded performance in general outdoor and indoor conditions, especially when lighting conditions are a factor, as well as high computational complexity when the UAV's payload is limited. To address this challenge, the MRS UAV platforms have integrated a smart sensor known as the UVDAR system~\cite{walter2019uvdar}, which is showcased in Fig.~\ref{fig:uvdar_hw} and Fig.~\ref{fig:uvdar_hw2}.
%When deploying multiple~\acp{UAV}, the~\ac{MRS} platforms must be able to fly and cooperate, even when direct radio communication is unreliable or external localization systems are unavailable. Such cooperation requires the agents of a swarm or formation to be able to localize each other based on their onboard sensory equipment. Computer vision is a state-of-the-art paradigm to mutual relative localization of robots~\cite{swarm_survey, vrba2019onboard, vrba2020markerless}. However, most of its current implementations suffer from degraded performance in general outdoor and indoor conditions, particularly where lighting conditions are concerned, in addition to significant computational complexity when the payload of~\ac{UAV} is limited. One solution used by the~\ac{MRS}~\ac{UAV} platforms is an optional smart sensor, called the~\ac{UVDAR} system~\cite{walter2019uvdar}, shown in Fig.~\ref{fig:uvdar_hw} and Fig.~\ref{fig:uvdar_hw2}.

This open-source\footnote{\url{https://github.com/ctu-mrs/uvdar_core}} system can be used to perform robust mutual localization. This is done by using ultraviolet LEDs that emit particular optical identification signals, which are detected and decoded using cameras equipped with an optical filter that significantly attenuates most of the image background~\cite{walter2018mutual, walter2018fast}, and thus simplifies the detection of the optical signals on the camera frame. This method has been tested with great success in multiple real-world deployments of swarms and formations of~\acp{UAV} \cite{petracek2020bioinspired, dmytruk2020safe, novak2021fast, afzal2021Icra, walter2019uvdar, midgard, Horyna2022ICUAS}. 

%%[TIAGO] I uncommented the paragraph bellow
Recently, we have improved the identification capacity of the~\ac{UVDAR} by designing special sets of binary signals for the blinking patterns of the ultraviolet LEDs~\cite{BonillaTRO22, BonillaIEEEProceedings}. One of the key properties of these sets of binary sequences is that two arbitrary binary sequences of the same length are considered different if, and only if, their circular Hamming distance is not zero. This property allows the~\ac{UVDAR} system to distinguish different ultraviolet LED blinking patterns without the need for implementing any type of synchronization and thus reduces the latency of the identification process.

Given the length of the binary sequences used for the blinking patterns, our design algorithm proposed in~\cite{BonillaTRO22, BonillaIEEEProceedings} creates a set with a maximum number of circularly different binary sequences, satisfying various properties that simplify the detection of the LEDs blinking patterns. Among these properties, we have the maximum duration in which a LED can remain turned off. Limiting this time helps the~\ac{UVDAR} system to better track the blinking LEDs on the camera frame when there is relative motion among the\acp{UAV}. The larger the binary sequences, the more~\acp{UAV} can be identified. However, this length is limited by the stability of the frame rate of the~\ac{UVDAR} camera and by the number of different~\acp{UAV}. See~\cite{BonillaTRO22, BonillaIEEEProceedings} for further details. 

If each~\ac{UAV} is assigned only one different blinking sequence, then the~\ac{UVDAR} system can be used only for positional localization of~\acp{UAV}. However, if more blinking sequences are assigned to each~\ac{UAV}, then the~\ac{UVDAR} system can also be used to estimate their relative heading.
The output of the~\ac{UVDAR} localization is a mean relative pose with error distribution, addressing the noise and ambiguity inherent in real-world vision systems.

A modified version of the~\ac{UVDAR} system has been developed to enable optical communication between the \acp{UAV}. This modified system is called~\ac{UVDAR}-COM~\cite{Horyna2022ICUAS} and allows simultaneous communication and localization between the \acp{UAV}. Currently, research is being performed to extend its communication range and increase its bit rate, as well as to enable the utilization of the known measurement error covariances for more reliable sensor-based cooperative behavior.

%%% END SECTION ============================================================

%%% START SECTION ==========================================================

%%[OWNER]: Pavel Stoudek
\subsection{Additional actuators}
\label{sec:additionalActuators}

The MRS UAV platforms are versatile and can be employed in various indoor and outdoor applications. Some of these applications require specialized actuators that are unique to the particular challenge. For example, in the 2017 MBIZRC competition, electromagnetic grippers were designed to grasp metallic tokens~\cite{spurny2019cooperative, loianno2018localization}, while in MBIZRC 2020, polyurethane foam blocks were used for wall construction~\cite{baca2020autonomous, stibinger2020mobile}. In the fire extinguishing challenge of the same competition, a water cannon and a device that deploys a fire-suppression blanket were utilized, and in another challenge, a system for intercepting a flying target was required~\cite{stasinchuk2020multiuav}.
%The~\ac{MRS}~\ac{UAV} platforms are used in many applications, both indoor and outdoor. Such applications often require the use of a specific (and sometimes unique) actuator. Actuators used by the~\ac{MRS}~\ac{UAV} platforms include the electromagnetic grippers designed for grasping metallic tokens in the 2017~\ac{MBZIRC} competition~\cite{spurny2019cooperative, loianno2018localization}, and polyurethane foam blocks used for wall construction in the~\ac{MBZIRC} 2020~\cite{baca2020autonomous, stibinger2020mobile}. In~\ac{MBZIRC} 2020, a water cannon and a device deploying a fire-suppression blanket in the fire extinguishing challenge were used. Another challenge in the competition required the use of a system for intercepting a flying target~\cite{stasinchuk2020multiuav}.

\begin{figure*}[tb]
    \centering
    \resizebox{1.0\textwidth}{!}{
    \pgfdeclarelayer{foreground}
\pgfsetlayers{background,main,foreground}

\makeatletter
\newcommand{\gettikzxy}[3]{%
  \tikz@scan@one@point\pgfutil@firstofone#1\relax
  \edef#2{\the\pgf@x}%
  \edef#3{\the\pgf@y}%
}
\makeatother

\tikzset{radiation/.style={{decorate,decoration={expanding waves,angle=90,segment length=4pt}}}}

\begin{tikzpicture}[->,>=stealth', node distance=3.0cm,scale=1.0, every node/.style={scale=1.0}]

  %%{ nodes

  \node[state, shift = {(0.0, 0.0)}] (battery) {
      \begin{tabular}{c}
        \footnotesize Battery
      \end{tabular}
    };

  \node[state, right of = battery, shift = {(0.5, 0.0)}] (measurement) {
      \begin{tabular}{c}
        \footnotesize U and I\\
        \footnotesize measurement
      \end{tabular}
    };
    
  \node[state, right of = measurement, shift = {(0.5, 0.0)}] (main_psu) {
      \begin{tabular}{c}
        \footnotesize Main computer\\
        \footnotesize power supply
      \end{tabular}
    };
        
  \node[state, right of = main_psu, shift = {(0.5, 0.0)}] (FT4232H) {
      \begin{tabular}{c}
        \footnotesize FT4232H
      \end{tabular}
    };
            
  \node[state, right of = FT4232H, shift = {(0.5, 0.0)}] (5v) {
      \begin{tabular}{c}
        \footnotesize 3x 5V\\
        \footnotesize power supply
      \end{tabular}
    };
            
  \node[state, right of = 5v, shift = {(0.5, 0.0)}] (plant) {
      \begin{tabular}{c}
        \footnotesize UAV plant\\
        \footnotesize ESCs, motors
      \end{tabular}
    };
 
  \node[state, above of = main_psu, shift = {(0.0, -0.5)}] (main_computer) {
      \begin{tabular}{c}
        \footnotesize Main computer
      \end{tabular}
    };

  \node[state, right of = main_computer, shift = {(0.5, 0.0)}] (mrs_modules) {
      \begin{tabular}{c}
        \footnotesize 3x MRS modules 
      \end{tabular}
    };
    
  \node[state, right of = mrs_modules, shift = {(0.5, 0.0)}] (pixhawk) {
      \begin{tabular}{c}
        \footnotesize Pixhawk 
      \end{tabular}
    };

  %%}

  %%{ paths
  
 \path[->, line width=0.5mm] ($(battery.east) + (0.0, 0)$) edge [] node[above, midway, shift = {(0.1, 0.05)}] {
   \begin{tabular}{c}
 \end{tabular}} ($(measurement.west) + (0.0, 0.00)$);

 \draw[-, line width=0.5mm] ($(measurement.south)+(0.0, 0.0)$) -- ($(measurement.south) + (0.0, -0.5)$) node[right, midway, shift = {(0.5, 0.0)}] {
    \begin{tabular}{c}
        \footnotesize Battery power 
 \end{tabular}} -- ($(plant.south) + (0.45, -0.5)$) edge [->] ($(plant.south)+ (0.45, -0.0)$);
 \path[->, line width=0.5mm] ($(plant.south) + (0.15, -0.5)$) edge ($(plant.south) + (0.15, 0.0)$);
 \path[->, line width=0.5mm] ($(plant.south) + (-0.15, -0.5)$) edge ($(plant.south) + (-0.15, 0.0)$);
 \path[->, line width=0.5mm] ($(plant.south) + (-0.45, -0.5)$) edge ($(plant.south) + (-0.45, 0.0)$);
    
  \path[->, line width=0.5mm] ($(main_psu.south) + (0.0, -0.5)$) edge ($(main_psu.south) + (0.0, 0.0)$);
  \path[->, line width=0.5mm] ($(5v.south) + (0.0, -0.5)$) edge ($(5v.south) + (0.0, 0.0)$);
  \path[->, line width=0.5mm] ($(main_psu.north) + (-0.4, 0.0)$) edge ($(main_computer.south) + (-0.4, 0.0)$);
  
 \draw[<-] ($(main_computer.south) + (0.4, 0.0)$) -- node[right, near start, shift = {(-0.2, -0.1)}] {
    \begin{tabular}{c}
        \footnotesize USB 
 \end{tabular}} ($(main_computer.south |- FT4232H.north) + (0.4, 0.6)$) -- ($(FT4232H.west |- FT4232H.north) + (-0.6, 0.6)$)  -- ($(FT4232H.west) + (-0.6, .0)$) edge [->] ($(FT4232H.west) + (0.0, 0.0)$);
 
   \path[<->] ($(FT4232H.north) + (-0.3, 0.0)$) edge node[left, near end, shift = {(0.2, 0.05)}] {
    \begin{tabular}{c}
        \footnotesize UART 
 \end{tabular}} ($(mrs_modules.south) + (-0.3, 0.0)$);
   \path[<->] ($(FT4232H.north) + (-0.1, 0.0)$) edge ($(mrs_modules.south) + (-0.1, 0.0)$);
   \path[<->] ($(FT4232H.north) + (0.1, 0.0)$) edge ($(mrs_modules.south) + (0.1, 0.0)$);
   
    \gettikzxy{(pixhawk.south)}{\psx}{\psy}
    \gettikzxy{(FT4232H.north)}{\fnx}{\fny}
   \draw[<-] ($(FT4232H.north) + (0.3, 0.0)$) -- ($(\fnx,\psy) + (0.3, -0.45)$) -- ($(pixhawk.south) + (-0.45, -0.45)$) edge [->] ($(pixhawk.south) + (-0.45, 0.0)$);
   
    \path[->, line width=0.5mm] ($(5v.north) + (0.15, 0.0)$) edge ($(pixhawk.south) + (0.15, 0.0)$);
    \path[->, line width=0.5mm] ($(5v.north) + (-0.15, 0.0)$) edge ($(pixhawk.south) + (-0.15, 0.0)$);
    \draw[-, line width=0.5mm] ($(5v.north) + (-0.45, 0.0)$) -- ($(5v.north |- mrs_modules.south) + (-0.45, -1.2)$) -- ($(mrs_modules.south) + (0.8, -1.2)$) edge [->] ($(mrs_modules.south) + (0.8, 0.0)$);
    % \draw[-, line width=0.5mm] ($(5v.north) + (-0.45, 0.0)$) -- ($(mrs_modules.south ) + (-0.55, -1.5)$) edge [->] ($(mrs_modules.south) + (0.8, 0.0)$);
    
    \draw[-] ($(pixhawk.south) + (0.45, 0.0)$)  -- node[right, near start, shift = {(-0.2, -0.1)}] {
    \begin{tabular}{c}
        \footnotesize PWM/Dshot 
 \end{tabular}}($(pixhawk.south |- plant.east) + (0.45, 1.0)$) -- ($(pixhawk.south |- plant.east) + (1.45, 1.0)$) -- ($(pixhawk.south |- plant.west) + (1.45, -0.3)$) edge [->] ($(plant.west) + (0.0, -0.3)$);
 
  \path[->] ($(pixhawk.south |- plant.west) + (1.45, -0.1)$) edge [->] ($(plant.west) + (0.0, -0.1)$);
  \path[->] ($(pixhawk.south |- plant.west) + (1.45, 0.1)$) edge [->] ($(plant.west) + (0.0, 0.1)$);
  \path[->] ($(pixhawk.south |- plant.west) + (1.45, 0.3)$) edge [->] ($(plant.west) + (0.0, 0.3)$);
    
  %%}

  %%{ backgrounds

  \begin{pgfonlayer}{background}
    \path (measurement.west |- measurement.north)+(-0.45,0.8) node (a) {};
    \path (measurement.south -| 5v.east)+(0.75,-0.80) node (b) {};
    \path[fill=gray!3,rounded corners, draw=black!70, densely dotted]
      (a) rectangle (b);
  \end{pgfonlayer}
  \node [rectangle, above of=measurement, shift={(0.0,0.4)}, node distance=1.7em] (board) {\footnotesize Integrated circuit board};

  %%}

\end{tikzpicture}
    }
    \caption{A block diagram of the integrated distribution board. Thick lines in the diagram represent power connections, while thin lines show data connections.}
    \label{fig:integrated_board}
    \vspace{-1em}
\end{figure*}
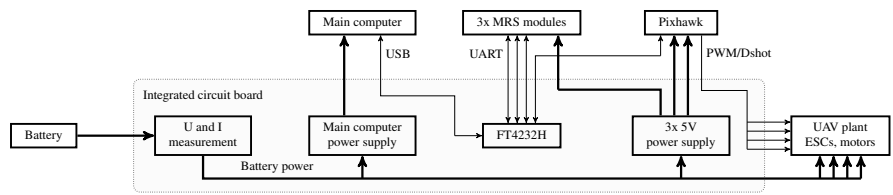

Furthermore, MRS platforms have been equipped with other types of actuators such as 
manipulators attached to the UAVs for pick-and-place tasks~\cite{vrba_ras2022}, gimbals for 
camera stabilization in the Dronument project~\cite{kratky2020autonomous, 
kratky2021documentation}, and even a capsule launcher for fire extinguishing in the DOFEC 
project~\cite{spurny2020autonomous}. Additionally, an 
Eagle.One\footnote{\url{https://eagle.one/en}} drone~\cite{vrba2020markerless} was equipped 
with a net launcher to capture invading drones in aerial no-fly zones (see 
Section~\ref{sec:applications} for more details).
%The~\ac{MRS} platforms have also used actuators, such as manipulators attached to the~\acp{UAV}~\cite{vrba_ras2022}, gimbals for camera stabilization (Dronument project)~\cite{kratky2020autonomous, kratky2021documentation}, and a capsule launcher for fire extinguishing (DOFEC project)~\cite{spurny2020autonomous}. In addition, even a net launcher, mounted on Eagle.One\footnote{\url{https://eagle.one/en}} drone~\cite{vrba2020markerless}, was used to capture invading drones in aerial no-fly zones (for details, see Section~\ref{sec:applications}).

%%% END SECTION ============================================================

%%% START SECTION ==========================================================

%%[OWNER]: Martin Sramek
\subsection{Thrust measurement with different configurations}
\label{sec:thrustMeasurement}

\begin{figure}[tb]
\centering
    \input{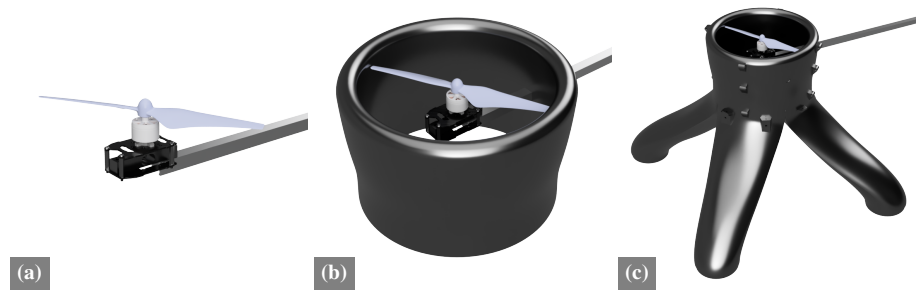}
    \vspace{-5mm}
    \caption{Thrust measurement configurations: (a) lone propeller, (b) ducted fan, and (c) ducted fan with ducting.}
    \label{fig:thMesConfig}
    \vspace{-1em}
\end{figure}

 Regarding future projects that will encompass unusual drone frames, it is necessary to measure the thrust of a propeller in different configurations. The main goal is to determine to what extent additional hardware near or below the propeller affects its thrust performance. Thus, we conducted some experiments with three types of configurations. They are (1) a lone propeller, (2) a ducted fan, and (3) a ducted fan with ducting (see Fig.~\ref{fig:thMesConfig}). In theory, the ducted fan configuration should improve turbulent flow leaking near the edge of the propeller. In contrast, the addition of ducting below the propeller should lower maximal thrust, thereby lowering the efficiency of the setup.

\begin{figure}[h]
%\vspace{0.3cm}
\centerline{\includegraphics[width = 0.47\textwidth]{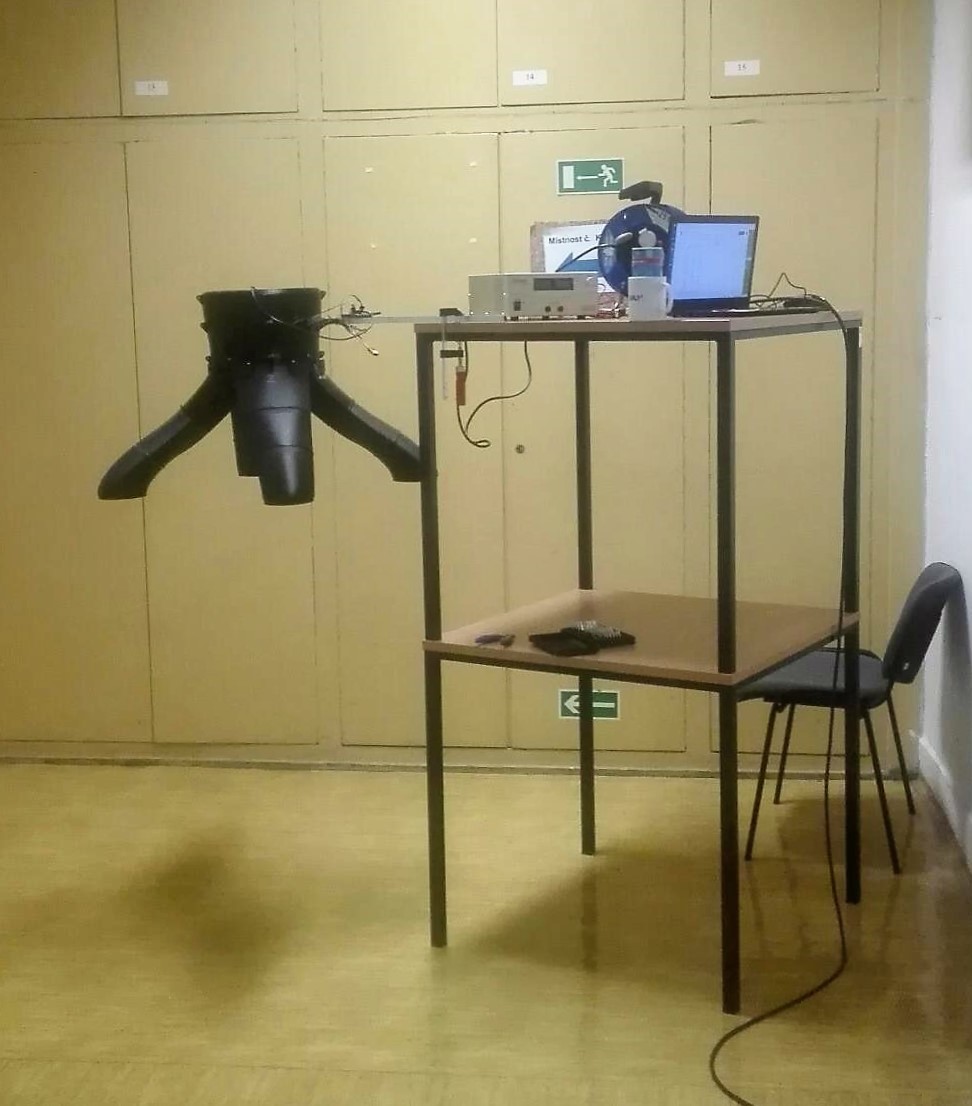}}
\caption{Experimental test rig for measuring the thrust of the ducted fan with additional ducting.}
\label{fig:thMesSetup}
\end{figure}

\begin{figure}[tb]
%\vspace{0.3cm}
\centerline{
\includegraphics[width = 0.5\textwidth]{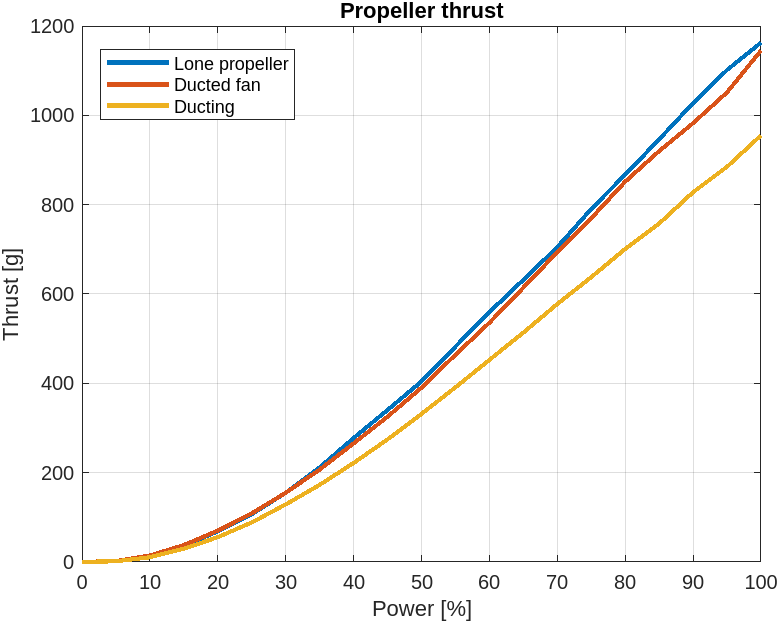}
\includegraphics[width = 0.5\textwidth]{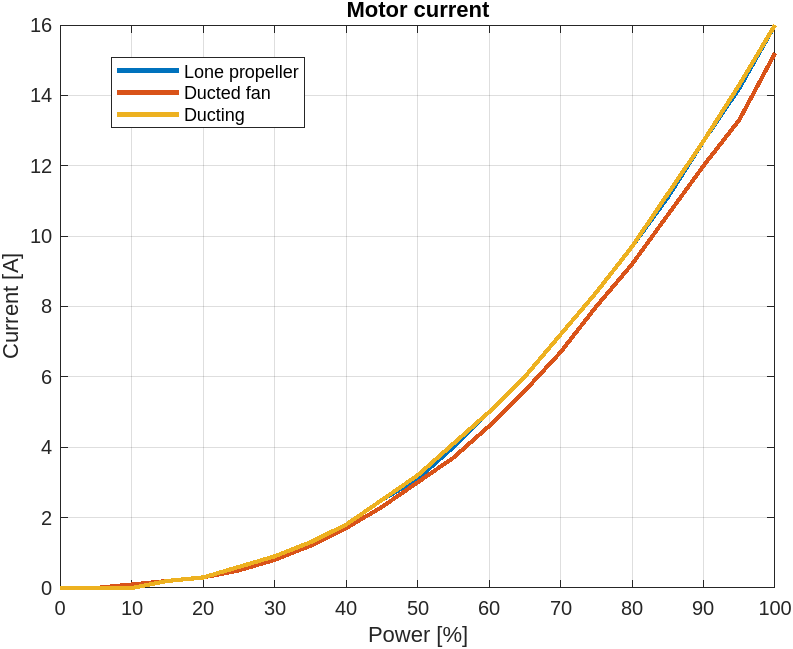}}
\caption{Thrust and motor current curves with various add-ons.}
\label{fig:thMesGraph}
\end{figure}

 The whole experimental setup consists of our F450 UAV motor configuration (motor, propeller, and motor holder) attached to a load cell through an aluminum extrusion. Said extrusion was affixed to an experimental test rig, forming a single inertial system with all add-ons. The test rig was set high enough to counteract the ground effect, as seen in Fig.~\ref{fig:thMesSetup}. All add-ons were printed on a standard FFF 3D printer and allowed for easy modification. In addition, the speed of the motor was incrementally increased. Thrust and the motor current were also measured, and are plotted in the graphs presented in Fig.~\ref{fig:thMesGraph}

 From the first graph, we can observe several outcomes. In the ducted fan configuration, the thrust is not affected very much. With the added ducting, we lose approximately 20\% of thrust, which is comparable with the information we received from external experts. In the current graph (see Fig.~\ref{fig:thMesGraph}), there is a visible 5\% lower current requirement using the ducted fan configuration, which means this setup is more energy efficient.

 In future testing, we will explore a coaxial propeller configuration with the same setup. More concretely, we will explore the distance between propellers, the angle of attack of each propeller, the number of blades, and, finally, the use of compressor propellers instead of traditional ones.

%%% END SECTION ============================================================

%%% START SECTION ==========================================================

%%[OWNER]: Daniel Hert
\section{Electrical Design and Configuration}
\label{sec:electricalDesignAndConfiguration}

The architecture and modularity of the~\ac{MRS}~\ac{UAV} platforms mean that additional electronic modules are required in order to provide additional features, such as low-level interfaces for communication with sensors and actuators or power supplies of different voltages, among others. The~\ac{MRS} group solved this necessity by designing and manufacturing a series of printed circuit distribution boards for the F450, T650, and X500 platforms, which are integrated into the platforms as structural pieces, replacing the top or the bottom board of the frames. This way, the custom board fulfills multiple roles, is a structural member, distributes power, and also integrates additional electronics. This reduces the mass of the~\ac{UAV} and reduces additional clutter and wiring.

The custom~\ac{MRS} printed circuit board contains two redundant 5\,V@3\,A buck converters to power the Pixhawk\footnote{The Pixhawk autopilot is an open-hardware and open-software architecture, which is advantageous for research in the field of aerial robotics~\cite{meier2015px4}.}, measures the current drawn from the battery and the battery voltage, routes power through high-current capable traces, and distributes throttle signals for all of the motors. The board also comes with standardized expansion slots to connect other low-level boards called MRS modules, which can provide the~\ac{UAV} with additional capabilities. The communication is facilitated by FT4232H: a USB-to-quad serial converter that connects up to four separate UARTs through a single USB 2.0 cable to the main computer. When connected, the device appears as four separate serial ports on the main computer, which makes the development of accompanying software very straightforward. One of the UARTs is used to provide a link between Pixhawk and the main computer using the Mavlink protocol, while the three remaining UARTs connect to the slots for~\ac{MRS} modules. One additional 5\,V power supply powers the~\ac{MRS} modules. A functional diagram of the integrated distribution board is shown in Fig.~\ref{fig:integrated_board}.

\begin{figure}[tb]
    \centering
    \input{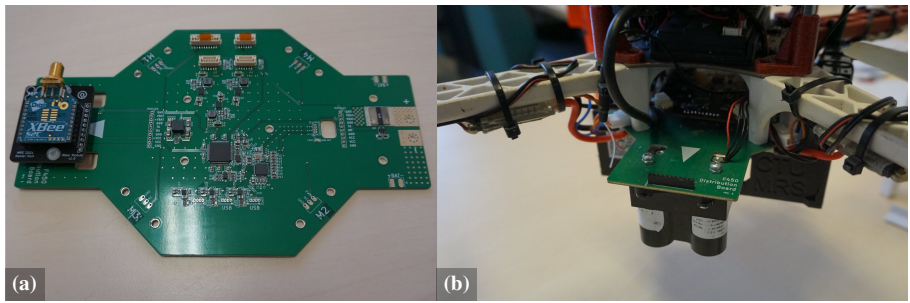}
    \vspace{-5mm}
    \caption{Distribution board for the F450 shown with an Xbee~\ac{MRS} module (a), and integrated into the F450 platform with an~\ac{UVDAR} controller~\ac{MRS} module (b).}
    \label{fig:distboard}
    \vspace{-1em}
\end{figure}

\begin{figure}[tb]
    \centering
    \includegraphics[width=0.6\columnwidth]{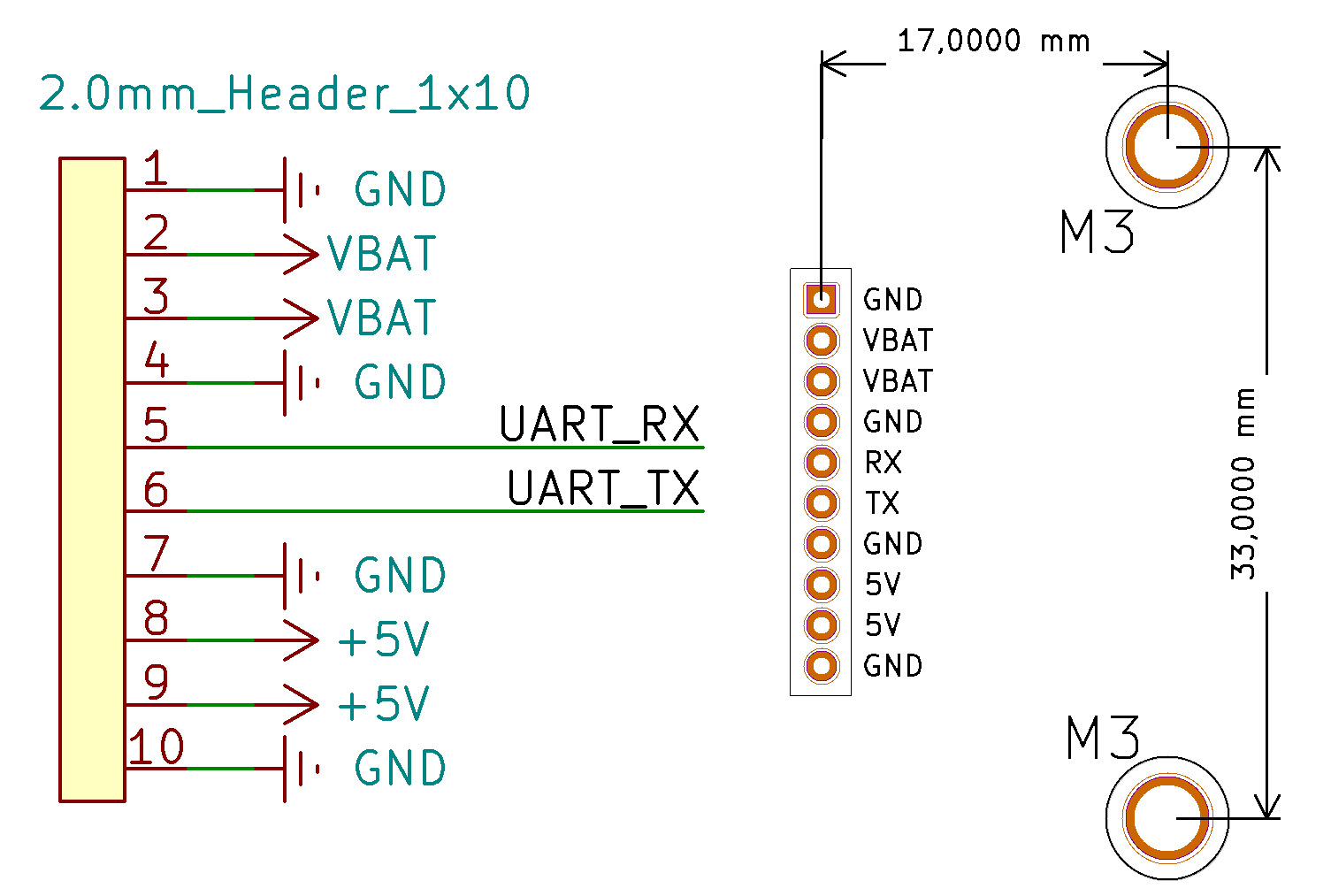}
    \vspace{-5mm}
    \caption{Standardized electrical and mechanical interface for~\ac{MRS} modules.}
    \label{fig:mrsmodule}
    \vspace{-1em}
\end{figure}

\ac{MRS} modules are small daughterboards that can be connected to the main distribution board through a standardized interface. The electrical connection is through a standard 2.0\, mm 10-pin header, which provides UART communication to the main computer through the FT4232H, stabilized 5\,V rail to power electronics of the MRS module, and direct battery connection for applications that require higher voltage. The~\ac{MRS} module can draw up to 2\,A on both the 5\,V rail and direct battery rail. The mechanical connection is through two M3 mounting posts, which are installed into the main distribution board. The module is then connected through the header and secured with two M3 bolts to the mounting posts. The standardized electrical and mechanical interface for the MRS modules is shown in Fig.~\ref{fig:mrsmodule}.

Up to three~\ac{MRS} modules can be added to the main board to increase the system's capabilities. For example, one module controls the LEDs of the~\ac{UVDAR} system, allowing up to eight high-power LEDs to be individually controlled and their blinking frequencies to be changed based on commands from the high-level computer. Another module has a slot for XBee radios, providing the~\ac{UAV} with a means of low-level communication. An MRS module with an integrated 12\,V buck regulator and power switches can be used to control LED strips attached to the~\ac{UAV}, providing additional lighting for cameras. Figure~\ref{fig:distboard} shows a distribution board with an XBee~\ac{MRS} module.

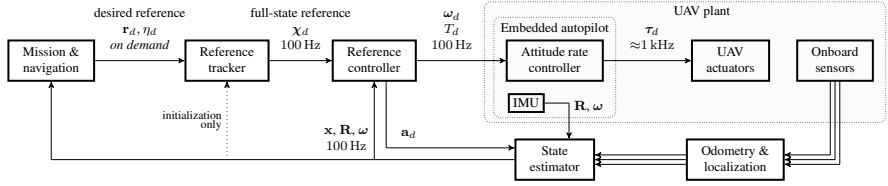
\begin{figure*}[tb]
    \centering
    \resizebox{1.0\textwidth}{!}{
    \pgfdeclarelayer{foreground}
\pgfsetlayers{background,main,foreground}

\tikzset{radiation/.style={{decorate,decoration={expanding waves,angle=90,segment length=4pt}}}}

\begin{tikzpicture}[->,>=stealth', node distance=3.0cm,scale=1.0, every node/.style={scale=1.0}]

  %%{ nodes

  \node[state, shift = {(0.0, 0.0)}] (navigation) {
      \begin{tabular}{c}
        \footnotesize Mission \&\\
        \footnotesize navigation
      \end{tabular}
    };

  % \node[state, left of = navigation, shift = {(0.5, 0.0)}] (nimbro) {
  %     \begin{tabular}{c}
  %       \footnotesize Nimbro \\
  %       \footnotesize Network
  %     \end{tabular}
  %   };

  \node[state, right of = navigation, shift = {(0.7, 0)}] (tracker) {
      \begin{tabular}{c}
        \footnotesize Reference \\
        \footnotesize tracker
      \end{tabular}
    };

  \node[state, right of = tracker, shift = {(0.1, 0)}] (controller) {
      \begin{tabular}{c}
        \footnotesize Reference \\
        \footnotesize controller
      \end{tabular}
    };

  \node[state, right of = controller, shift = {(0.8, -0)}] (attitude) {
      \begin{tabular}{c}
        \footnotesize Attitude rate\\
        \footnotesize controller
      \end{tabular}
    };

  \node[smallstate, below of = attitude, shift = {(-0.6, 2.1)}] (imu) {
      \footnotesize IMU
    };

  \node[state, right of = attitude, shift = {(0.7, -0)}] (actuators) {
      \begin{tabular}{c}
        \footnotesize UAV \\
        \footnotesize actuators
      \end{tabular}
    };

  \node[state, right of = actuators, shift = {(-0.8, -0)}] (sensors) {
      \begin{tabular}{c}
        \footnotesize Onboard \\
        \footnotesize sensors
      \end{tabular}
    };

  \node[state, below of = attitude, shift = {(0, 0.9)}] (estimator) {
      \begin{tabular}{c}
        \footnotesize State \\
        \footnotesize estimator
      \end{tabular}
    };

  \node[state, right of = estimator, shift = {(0.8, 0.0)}] (localization) {
      \begin{tabular}{c}
        \footnotesize Odometry \& \\
        \footnotesize localization
      \end{tabular}
    };

  %%}

  %%{ paths

  \path[->] ($(navigation.east) + (0.0, 0)$) edge [] node[above, midway, shift = {(0.0, 0.05)}] {
      \begin{tabular}{c}
        \footnotesize desired reference\\
        \footnotesize $\mathbf{r}_d, \eta_d$\\
        \footnotesize \textit{on demand}
    \end{tabular}} ($(tracker.west) + (0.0, 0.00)$);

  % \path[->] ($(nimbro.east) + (0.0, 0)$) edge [] node[above, midway, shift = {(0.0, 0.05)}] {
  %     \begin{tabular}{c}
  %   \end{tabular}} ($(navigation.west) + (0.0, 0.00)$);

  \path[->] ($(tracker.east) + (0.0, 0)$) edge [] node[above, midway, shift = {(0.0, 0.05)}] {
      \begin{tabular}{c}
        \footnotesize full-state reference\\
        \footnotesize $\bm{\chi}_d$\\
        \footnotesize \SI{100}{\hertz}
    \end{tabular}} ($(controller.west) + (0.0, 0.00)$);

  \path[->] ($(tracker.south |- estimator.west) + (0.0, 0.0)$) edge [dotted] node[left, midway, shift = {(0.2, 0.00)}] {
      \begin{tabular}{r}
        \scriptsize initialization\\[-0.5em]
        \scriptsize only
    \end{tabular}} ($(tracker.south) + (0.0, 0.00)$);

  \path[->] ($(controller.east) + (0.0, 0)$) edge [] node[above, midway, shift = {(-0.2, 0.05)}] {
      \begin{tabular}{c}
        \footnotesize $\bm{\omega}_d$\\
        \footnotesize $T_d$ \\
        \footnotesize \SI{100}{\hertz}
    \end{tabular}} ($(attitude.west) + (0.0, 0.00)$);

  \draw[-] ($(controller.south)+(0.25,0)$) -- ($(controller.south |- estimator.west) + (0.25, 0.25)$) edge [->] node[above, near start, shift = {(-0.2, 0.05)}] {
      \begin{tabular}{c}
        \footnotesize $\mathbf{a}_d$
    \end{tabular}} ($(estimator.west) + (0, 0.25)$);

  \path[->] ($(attitude.east) + (0.0, 0)$) edge [] node[above, midway, shift = {(0.1, 0.05)}] {
      \begin{tabular}{c}
        \footnotesize $\bm{\tau}_d$ \\
        \footnotesize $\approx$\SI{1}{\kilo\hertz}
    \end{tabular}} ($(actuators.west) + (0.0, 0.00)$);

  \path[-] ($(estimator.west)+(0, 0)$) edge [] node[above, near start, shift = {(-1.1, 0.0)}] {
      \begin{tabular}{c}
        \footnotesize $\mathbf{x}$, $\mathbf{R}$, $\bm{\omega}$\\
        \footnotesize \SI{100}{\hertz}
    \end{tabular}} ($(navigation.south |- estimator.west)$) -- ($(navigation.south |- estimator.west)$) edge [->,] ($(navigation.south)+(0, 0)$);

  % \path[-] ($(estimator.west)+(0, 0)$) edge [] node[above, near start, shift = {(-1.0, 0.0)}] {
  %     \begin{tabular}{c}
  %   \end{tabular}} ($(nimbro.south |- estimator.west)$) -- ($(nimbro.south |- estimator.west)$) edge [->,] ($(nimbro.south)+(0, 0)$);

  \path[->] ($(controller.south |- estimator.west)+(0, 0)$) edge [] ($(controller.south)$);

  \draw[-] ($(imu.east) + (0.0, 0.0)$) -- ($(estimator.north |- imu.east) + (0.3, 0)$) edge [->] node[right, midway, shift = {(-0.2, 0.3)}] {
      \begin{tabular}{c}
        \footnotesize $\mathbf{R}$, $\bm{\omega}$
    \end{tabular}} ($(estimator.north) + (0.3, 0.0)$);

  \draw[-] ($(sensors.south)+(0, 0)$) -- ($(sensors.south |- localization.east)$) edge [->] ($(localization.east)$);
  \draw[-] ($(sensors.south)+(0.1, 0)$) -- ($(sensors.south |- localization.east) + (0.1, -0.1)$) edge [->] node[midway, shift = {(0.0, -0.20)}] {
      \begin{tabular}{c}
        % \footnotesize $\vdots$
    \end{tabular}} ($(localization.east) + (0.0, -0.1)$);
  \draw[-] ($(sensors.south)+(-0.1, 0)$) -- ($(sensors.south |- localization.east) + (-0.1, 0.1)$) edge [->]  ($(localization.east) + (0.0, 0.1)$);

  \draw[->] ($(localization.west)+(0, 0)$) -- ($(estimator.east)$);
  \draw[->] ($(localization.west)+(0, 0.1)$) -- ($(estimator.east) + (0, 0.1)$);
  \draw[->] ($(localization.west)+(0, -0.1)$) -- node[midway, shift = {(0.0, -0.2)}] {
      \begin{tabular}{c}
        % \footnotesize $\vdots$
    \end{tabular}} ($(estimator.east) + (0, -0.1)$);

  %%}

    % \draw[-, radiation, decoration={angle=45}] ($(nimbro.west) + (0.0, -0.0)$) -- +(0:-0.5);

  %%{ backgrounds

  \begin{pgfonlayer}{background}
    \path (attitude.west |- attitude.north)+(-0.45,0.8) node (a) {};
    \path (imu.south -| sensors.east)+(+0.25,-0.20) node (b) {};
    \path[fill=gray!3,rounded corners, draw=black!70, densely dotted]
      (a) rectangle (b);
  \end{pgfonlayer}
  \node [rectangle, above of=actuators, shift={(-0.6,0.55)}, node distance=1.7em] (autopilot) {\footnotesize UAV plant};

  \begin{pgfonlayer}{background}
    \path (attitude.west |- attitude.north)+(-0.25,0.47) node (a) {};
    \path (imu.south -| attitude.east)+(+0.25,-0.10) node (b) {};
    \path[fill=gray!3,rounded corners, draw=black!70, densely dotted]
      (a) rectangle (b);
  \end{pgfonlayer}
  \node [rectangle, above of=attitude, shift={(0,0.2)}, node distance=1.7em] (autopilot) {\footnotesize Embedded autopilot};

  %%}

\end{tikzpicture}
    }
    \caption{Diagram of the system architecture: \emph{Mission \& navigation} software supplies the position and heading reference ($\mathbf{r}_d$, $\eta_d$) to a reference tracker.
\emph{Reference tracker} creates a smooth and feasible reference $\bm{\chi}$ for the feedback controller.
The feedback \emph{Reference controller} produces the desired thrust and angular velocities ($T_d$, $\bm{\omega}_d$) for the Pixhawk embedded flight controller.
The \emph{State estimator} fuses data from \emph{Onboard sensors} and \emph{Odometry \& localization} methods to create an estimate of the~\ac{UAV} translation and rotation ($\mathbf{x}$, $\mathbf{R}$)~\cite{baca2021mrs}.
    }
    \label{fig:system_architecture}
    \vspace{-1.2em}
\end{figure*}

%%% END SECTION ============================================================

%%% START SECTION ==========================================================

%\section{UAV DYNAMIC MODEL}
%[TO BE WRITTEN] ?? do we need? Maybe to have at least one equation there :-)

%%% END SECTION ============================================================

%%% START SECTION ==========================================================

%%[OWNER]: Tomas
\section{MRS UAV System}
\label{sec:MRSUAVSystem}

%{\color{red} Tomas: I don't have good internet, so please, grammar check the text.}
%{\color{red} Tomas: I added more detail to this otherwise good paragraph.}

The presented subsection briefly describes the MRS UAV System, an open-source software package for controlling and deploying multi-rotor aerial vehicles~\cite{baca2021mrs}. The system is designed for use in both realistic simulations and real-world scenarios and is capable of executing complex missions in GNSS and GNSS-denied environments. The system includes two feedback control designs, one for precise and aggressive maneuvers and the other for a stable and smooth flight. Control reference generation is provided by the real-time virtual Model Predictive Control (MPC) tracker~\cite{baca2018model}. The system is modular, allowing users to develop and test their own feedback controllers, reference generators, and estimators. The MRS UAV control and estimation pipelines use rotation matrices and a novel heading-based convention to represent the one free rotational degree of freedom in 3D of a standard multi-rotor aircraft, instead of the Euler/Tait-Bryan angle representation which can cause confusion due to ambiguities and singularities. The system also includes realistic simulation of UAVs, sensors, and localization systems. The MRS UAV system has been used in various real-world system deployments that have helped shape the system into its current form, as shown in Fig.~\ref{fig:system_architecture}.

The MRS UAV System also provides features for managing and controlling multi-\acp{UAV} systems.
The \ac{MPC} tracker~\cite{baca2018model} integrates an outdoor mutual collision resolution system.
Mutual collision avoidance utilizes wireless communication for exchanging predicted future trajectories of each \ac{UAV}, which are consequently used to temporarily alter the \acp{UAV}' motion.
Therefore, even head-on collisions of an arbitrary number of \acp{UAV} are automatically avoided without human intervention.
However, the users are \textbf{discouraged} from relying on the avoidance system as a \emph{basic feature} of the system when designing their multi-\ac{UAV} control algorithms.
Users should implement their collision resolution primarily in their control design and only use the provided avoidance as a failsafe.

The wireless communication within the \ac{MRS}~\ac{UAV} system is managed by the NimbRo network package\footnote{\url{https://github.com/AIS-Bonn/nimbro_network}, Accessed on 2023-03-24}, which provides a multi-master-like communication scheme between multiple independent ROS-controller robots.
Using the NimbRo network package, the ROS topics and services of a single \ac{UAV} can be exposed on other \acp{UAV} for supporting distributed control algorithms.
The topics and services can also be exposed to a ground control station, which can control a group of multiple \acp{UAV} in a centralized manner.

%Short info from Tomas's paper - Tomas - Scheme, some equations??
% \textbf{TODO: Tiago (Tomas is on vacation, so please someone check it afterward)}}  
% TODO: Include scheme with MRS<->PX4 attitude controller}

%%% END SECTION ============================================================

%%% START SECTION ==========================================================

%%[OWNER]: Martin, Daniel, Vojtech Sp.
\section{Applications for the Aerial Platforms}
\label{sec:applications}

One important feature of the~\ac{MRS} system is that both~\ac{MRS} \ac{SW} and~\ac{HW} are smoothly coupled, independent of the real~\ac{UAV} platform used. This enables the~\ac{MRS} \ac{SW} system to be easily integrated with the robotic platforms independent of the hardware irregularities and modularities that may exist. This easy integration is often needed in general research. 

In this section, we present most of the~\ac{UAV} applications and the required platforms performed by our group. Realistic~\ac{SITL} simulations, robotic competitions, research on aerial autonomy in indoor and outdoor environments, and prototyping for industrial applications are among the case studies presented. Finally, we also present a summary of the principal attributes of our~\ac{UAV} platforms used for these applications in Table~\ref{tab:platforms}.

%TODO: Add table listing specifications (liftoff weight, configuration, purpose (research, industrial), custom-made/assembled from publicly-available stuff, good for}

\begin{table}[h]
  \centering
  \sisetup{per-mode=symbol}
   %\captionsetup{width=0.8\textwidth}
  \caption{Aerial platforms utilizing the~\ac{MRS} software stack in research, academic, and industry projects. \textit{Dimensions} is represented by the length of the main diagonal without propellers. \textit{Parts} denotes the \textbf{P}ublicly available and \textbf{C}ustom-made parts required for construction. \textit{Purpose} denotes the \textbf{R}esearch and \textbf{I}ndustrial platforms.%
   %Each platform is capable of carrying arbitrary sensors with appropriate weight and dimensions.
       \vspace{-1mm}
       }
  \resizebox{\columnwidth}{!}{%
  \begin{tabular}{l l l l l l l}
    \toprule
    % Omitted F550 due to lack of horizontal space
    Platform              & X500  & F450 & T650 & Dronument & Eagle.One & DOFEC\\\midrule
    Flight time (min)     & 25    & 15   & 20   & 7    & 10        & 10\\
    Weight (kg)           & 2.5   & 1.7  & 3.5  & 5.5  & 10        & 7\\
    Dimension (mm)        & 500   & 450  & 650  & 570  & 1250        & 657\\
    Propeller size (in)   & 13 & 9.4  & 15   & 12   & 18        & 15\\
    Battery capacity (Wh) & 199.8  & 99.9 & 177.6    & 355.2  & 355.2        & 355.2 \\
    Rotors count          & 4     & 4    & 4    & 8    & 8         & 8\\
    Parts                 & P     & P    & P    & C    & C         & C\\
    Purpose               & R/I   & R    & R    & R    & I         & I\\
    \bottomrule
  \end{tabular}
  }
  \label{tab:platforms}
  \vspace{-1.5em}
\end{table}

%%% END SECTION ============================================================

%%% START SECTION ==========================================================

\subsection{Key features of the MRS Drone}
\label{sec:keyfeatures}

A main advantage of our proposed platforms is the modularity feature. The modularity of our~\acp{UAV} can be seen not only in their mechanical design but on the electronic boards and in the~\ac{SW} system. Such a modular design allows our platforms to be applicable in a large range of applications and scenarios. Another advantage is the fix-yourself characteristic of our platforms; as an open-source platform, parts can be easily replaced and some can even be printed. This fix-yourself feature is intended to help young researchers to assemble the~\ac{UAV} themselves, thereby boosting research in aerial robotics. A third advantage includes the high computational and autonomous capabilities of our~\acp{UAV}, as could be seen in the comparison shown by Fig.~\ref{fig:comparison}. In contrast, the main disadvantage of our platforms is the low agility during flight. Nevertheless, our current research is aiming to suppress such disadvantages by creating more agile flight control and state estimation algorithms.

%Multi-robot system
Regarding multi-robot systems, our platforms can be easily used in a swarm of~\acp{UAV}. Furthermore, we use different technologies, such as a multi-spectral communication suite, traditional Wi-Fi, and low-power lightweight units based on RFM69HCW. Even bandwidth-intensive data up to \SI{1}{\mega\byte\per\second}, such as maps~\cite{musil2022spheremap} or images, are transmitted over a \SI{2.3}{\giga\hertz} Mobilicom MCU-30 Lite unit. These communication technologies enable our~\ac{MRS}~\ac{UAV} platforms to share data between robots and base stations. The cooperation algorithm itself is based on the application and is given to the user to create.

%%% END SECTION ============================================================

%%% START SECTION ==========================================================

%%[OWNER]: Tomas, Vojtech Sp., Petr Stib.
\subsection{Gazebo realistic simulations}
\label{sec:gazeboRealisticSimulations}

%[TO BE WRITTEN]Flight res
% \textbf{TODO:  Tomas, Vojta, Petr Stib???}}  
To facilitate the real-world deployment of novel aerial platforms, we have created a realistic simulation environment built on the open-source simulator Gazebo\footnote{\url{https://gazebosim.org/home}}.
The simulation stack is available to the community at the~\ac{MRS} GitHub\footnote{\url{https://github.com/ctu-mrs/simulation}}.
The package contains the realistic representation of our most used~\ac{UAV} hardware platforms, such as the DJI FlameWheel series (F330, F450, and F550), Tarot 650 sport, or the Holybro X500.
It provides a convenient interface for equipping the platforms with various payload configurations, sensory equipment, and actuators.

\begin{figure} [htb]
    \newcommand{\imheight}{10.50em}
    \newcommand{\xcap}{0.95em}
    \newcommand{\ycap}{0.8em}
    \newcommand{\fillopa}{0.3}
    \centering
    \begin{tikzpicture}
      \node[anchor=south west,inner sep=0] (b) at (0,0) {\adjincludegraphics[width=0.49\textwidth,trim={{0.0\width} {0.0\height} {0.0\width} {0.0\height}},clip]{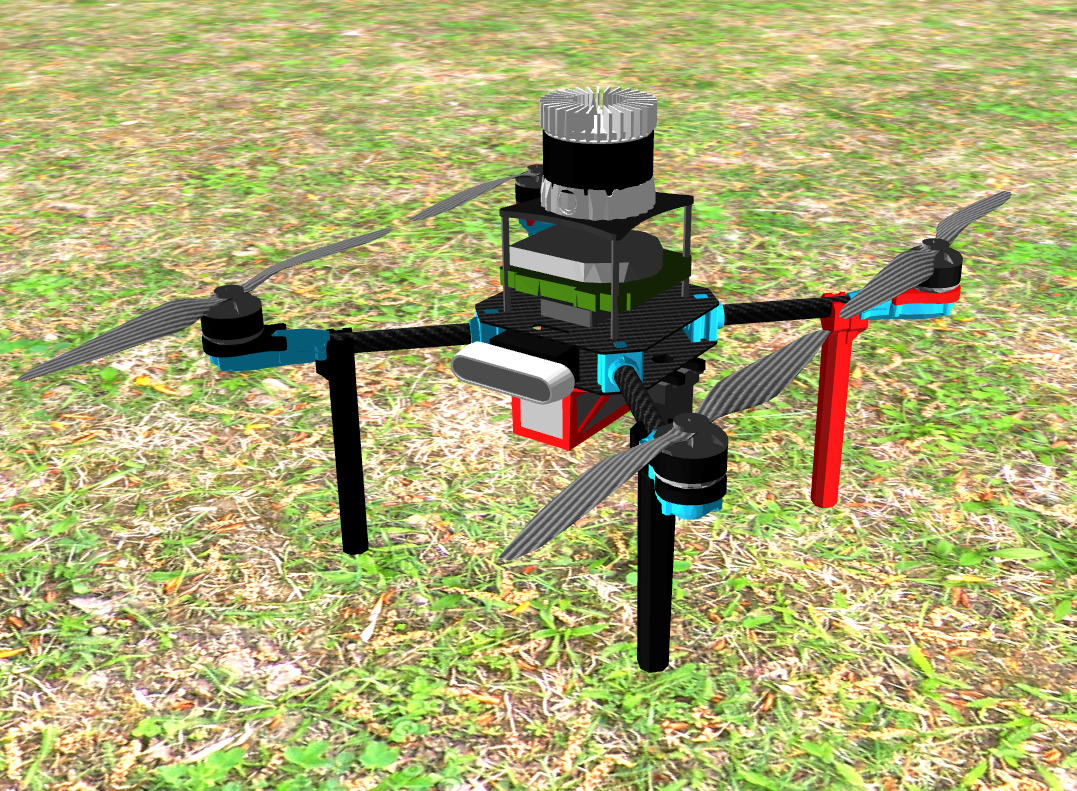}};%
    \begin{scope}[x={(b.south east)},y={(b.north west)}]
  \node[fill=black, fill opacity=\fillopa, text=white, text opacity=1.0] at (\xcap, \ycap) {\textbf{(a)}};
      \end{scope}
    \end{tikzpicture}
    \begin{tikzpicture}
      \node[anchor=south west,inner sep=0] (b) at (0,0) {\adjincludegraphics[width=0.49\textwidth,trim={{0.00\width} {0.0\height} {0.00\width} {0.00\height}},clip]{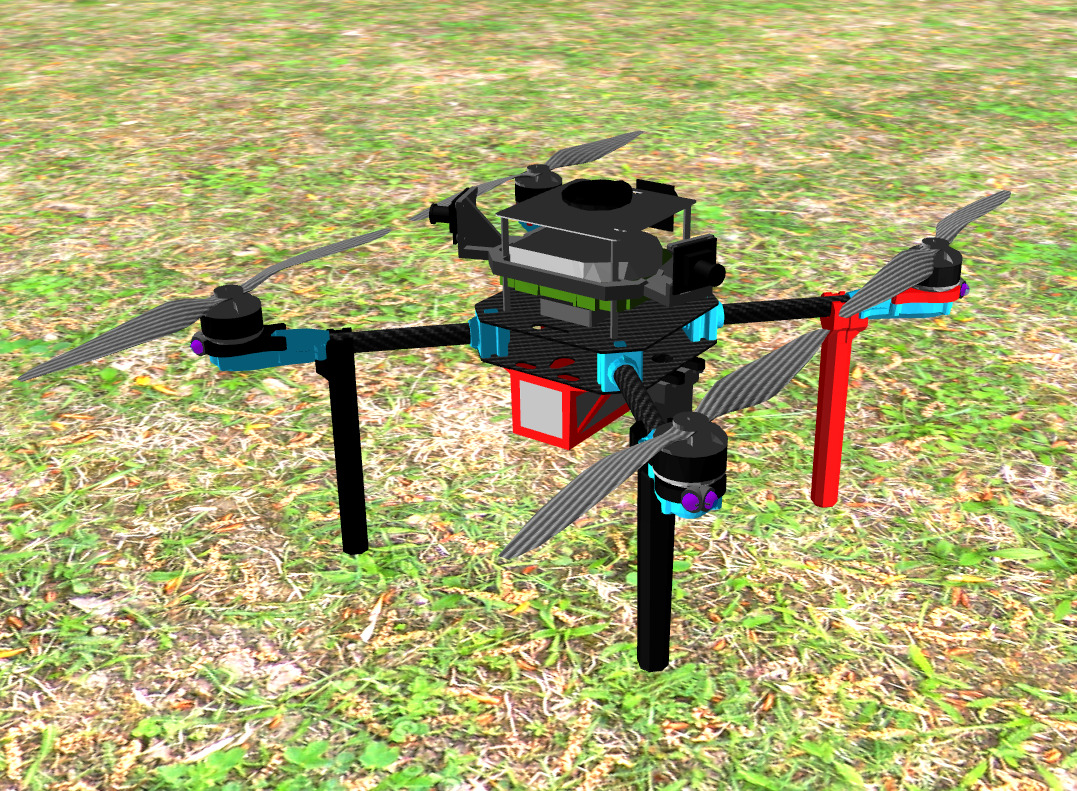}};%
    \begin{scope}[x={(b.south east)},y={(b.north west)}]
  \node[fill=black, fill opacity=\fillopa, text=white, text opacity=1.0] at (\xcap, \ycap) {\textbf{(b)}};
      \end{scope}
    \end{tikzpicture}
    \caption{The modularity of the MRS simulation stack: the X500 platform is equipped with the Ouster \ac{lidar} and the Realsense D435 (a), and a swarm configuration equipped with the UVDAR system including UV LEDs and three Bluefox cameras (b).}
    \label{fig:simulation_modularity}
\end{figure}

\begin{figure} [htb]
  \newcommand{\imheight}{10.50em}
  \newcommand{\xcap}{0.95em}
  \newcommand{\ycap}{0.8em}
  \newcommand{\fillopa}{0.3}
  \centering
  \begin{tikzpicture}
    \node[anchor=south west,inner sep=0] (b) at (0,0) {\adjincludegraphics[width=0.99\textwidth,trim={{0.0\width} {0.0\height} {0.0\width} {0.0\height}},clip]{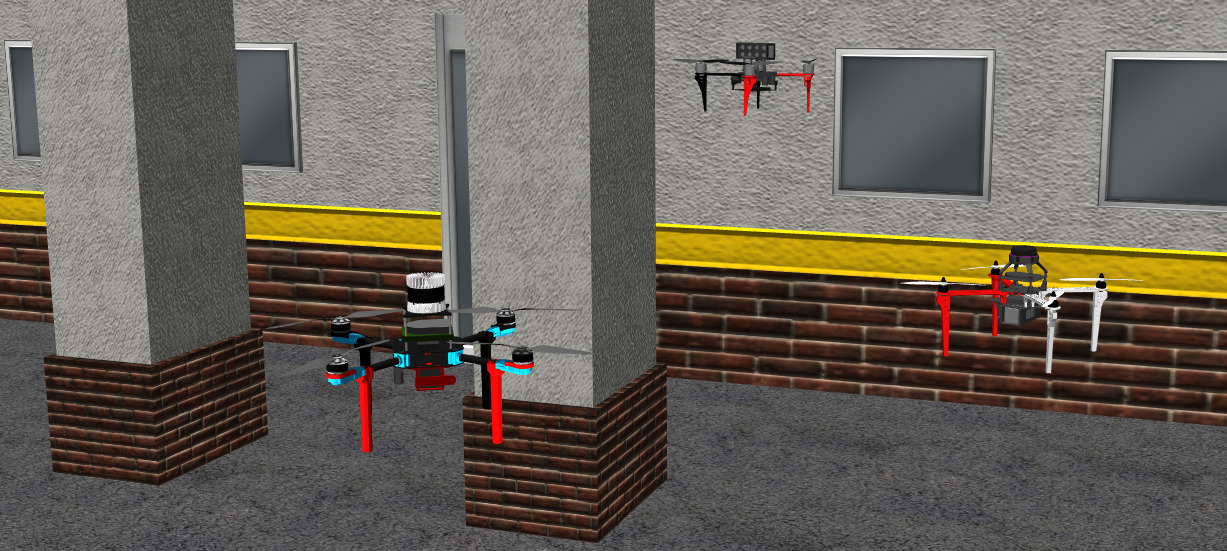}};%
  \end{tikzpicture}
  \caption{A snapshot from the Gazebo simulator showing a heterogeneous group of \acp{UAV} in a simulated urban environment.}
  \label{fig:simulation_heterogeneous}
\end{figure}

We employ the Jinja\footnote{\url{https://jinja.palletsprojects.com}} templating engine to assemble the simulated \acp{UAV} according to the user's requirements.
A base template is provided for each platform (i.e. propellers, motors, arms, and legs) and for individual components that can be attached to the base.
The user specifies the platform and the components to be attached, and the robot description for Gazebo is generated dynamically.
As a result, the environment is capable of multi-\ac{UAV} simulations, including heterogeneous~\ac{UAV} groups.
\reffig{fig:simulation_modularity} shows the X500 platform in two different flight configurations, and \reffig{fig:simulation_heterogeneous} shows a heterogeneous group of \acp{UAV} in a simulated urban environment.

The goal is to simulate the entire hardware setup with a high degree of realism, thereby minimizing the difference between simulation and the real world.
This way, the transition from the drawing board to real applications can be made seamlessly and with fewer safety risks.

% %%{ FIGURE: Simulation modularity
% %%}

  % %%{ FIGURE: Simulation heterogeneous

%%% END SECTION ============================================================

%%% START SECTION ==========================================================

%%[OWNER]: Matej and Vaclav
\subsection{Indoor real robot experiments}
\label{sec:indoorRealRobotExperiments}

Our proposed platforms are intended for both indoor and outdoor applications. In this subsection, we focus on indoor applications, such as historic monument documentation, the~\ac{sar}-motivated~\ac{darpa}~\ac{subt} Challenge, and industrial inspection applications. 
%In this section, we demonstrate in multiple scenarios that with a rapid sensory and actuation modification, the~\ac{UAV} is able to localize itself in an indoor environment and achieve the desired mission objectives.
Due to quick modifications of the platform realized by changing sensors and actuators, we showcase the ability to conduct complex mission objectives in multiple indoor environments.
The indoor sensory payload is able to sense the surroundings of the~\ac{UAV} to create an environment model, which is then used for navigation.

%%% END SECTION ============================================================

%%% START SECTION ==========================================================

%%[OWNER]: Matej and Vit
%\subsubsection{DARPA SubT --- S\&R competition and exploration of subterranean environments}
\subsubsection{\ac{darpa} Subterranean Challenge --- S\&R competition in underground environments}
\label{sec:darpaSubT}
 
% %%{ new stuff (rewritten in other words and extended)
The~\ac{subt} was organized by~\ac{darpa}\cite{orekhov2022darpa} in the years 2017-2021 to push the limits of robot deployment in underground~\ac{sar} operations.
In the competition, the robotic team was sent to search for mannequins representing survivors, their belongings, and other specific objects into a previously unvisited underground space. This space consisted of a combination of three domains: mining tunnels, natural caves, and man-made urban structures.
The harshness of each environment type posed demanding challenges on the robot autonomy, which challenged the robustness of each module of the developed solutions.
While the complicated topology of the unknown environment, which consisted of both narrow corridors and large caves/rooms, tested the navigation capabilities, the sensory perception was challenged by the severe degradation of sensing conditions.
The insufficient illumination combined with airborne particles of various origins (e.g., dust, fog, smoke) required a complementary multi-modal perception payload for reliable self-localization, the building of an accurate model of the environment, and the detection of search objects.
The requirement of a significant sensory payload competes with the need for minimization of the~\ac{UAV} in order to traverse such constrained corridors, as well as the need for a long flight time, in order to get to the furthest parts of the searched space.
Throughout the five-year-long competition, the platform developed evolved significantly~\cite{petrlik2020robust, kratky2021exploration, petracek2021caves, Roucek2022fr}, with each iteration improving upon the last.

% frame, propulsion 
The platform developed for the final event of~\ac{darpa}~\ac{subt} was based on the compact and lightweight Holybro X500 quadrotor frame.
The size of the carbon fiber frame could be adjusted to accommodate both 13-inch and 14-inch carbon propellers, which were paired with MN3510 KV700 motors from T-motor driven by easily configurable Turnigy Bl-Heli 32 51A~\acp{ESC}.
This propulsion system provides high efficiency and large payload capacity for mission-critical sensors.

% MRS power distribution board
The top board of the X500 frame is replaced by the custom~\ac{MRS} distribution board described in Section~\ref{sec:electricalDesignAndConfiguration}.
The board provides a communication interface between the main Intel NUC i7-10710U computer, Pixhawk flight controller, and two~\ac{MRS} modules.
One of the modules controls the~\SI{12}{\volt} LED strips that help achieve lower exposure times of RGB cameras.
The second module is an XBee radio receiver that allows the organizers to issue an emergency land command at the end of the mission or to prevent an accident.
Regarding power distribution, the distribution board provides 4S Li-Pol battery voltage (\SIrange{14.0}{16.8}{\volt}), which directly powers the~\acp{ESC} and the main computer.
A total of three \SI{5}{\volt}/\SI{3}{\ampere} buck converters are used with two providing a redundant power source for the Pixhawk flight controller, and the remaining for powering the~\ac{MRS} modules.
The 3D~\ac{lidar} is powered by a \SI{24}{\volt}/\SI{2}{\ampere} boost converter. % flight time, batteries
The long flight time of \SI{20}{\minute} with all sensors and total mass of \SI{3.3}{\kilo\gram} is achieved by connecting two 4S \SI{6750}{\milli\ampere\hour} Li-Pol batteries in parallel.
The advantage of using two smaller batteries lies in easier transportation since batteries under \SI{100}{\watt\hour} can be taken in carry-on baggage.

% perception payload
The perception payload consists of a horizontally mounted OS0-128 Ouster \ac{lidar} as the main sensor for establishing a model of the traversed environment and preventing collisions. 
A wide vertical field of \SI{90}{\degree} is covered by 128 scanning lines of the sensor, and the blind spots above and below the~\ac{UAV} are covered by two Realsense D435 RGB-D cameras.
The full coverage of the~\ac{UAV}'s surroundings enables collision-less motion in cluttered environments with high verticalities, such as caves or staircases.
The bottom-facing camera can also be used for safe landing spot detection.
The two RGB-D cameras, together with two RGB Basler Dart daA1600 cameras on the front legs, provide image streams for the object detection neural network running on the integrated GPU of the Intel NUC.

% communication
For multi-robot cooperation, the~\ac{UAV} was equipped with a multi-spectral communication suite.
In addition to the traditional Wi-Fi, the~\ac{UAV} can share data over two independent communication channels.
A low-power lightweight unit based on a RFM69HCW transceiver provides a low bandwidth of \SI{100}{\byte\per\second} at \SI{868}{\mega\hertz} or \SI{915}{\mega\hertz} for sharing critical mission data.
Bandwidth-intensive data up to \SI{1}{\mega\byte\per\second}, such as maps~\cite{musil2022spheremap} or images, are transmitted over \SI{2.3}{\giga\hertz} Mobilicom MCU-30 Lite unit.
Both technologies implement a meshing solution for \textit{ad-hoc} connections of arbitrary combinations of deployed robots.

\begin{figure}[tb]
    \centering
    \begin{tikzpicture}
      \node[anchor=south west,inner sep=0] (b) at (0,0) {\adjincludegraphics[width=\columnwidth,trim={{0.00\width} {0.0\height} {0.00\width} {0.00\height}},clip]{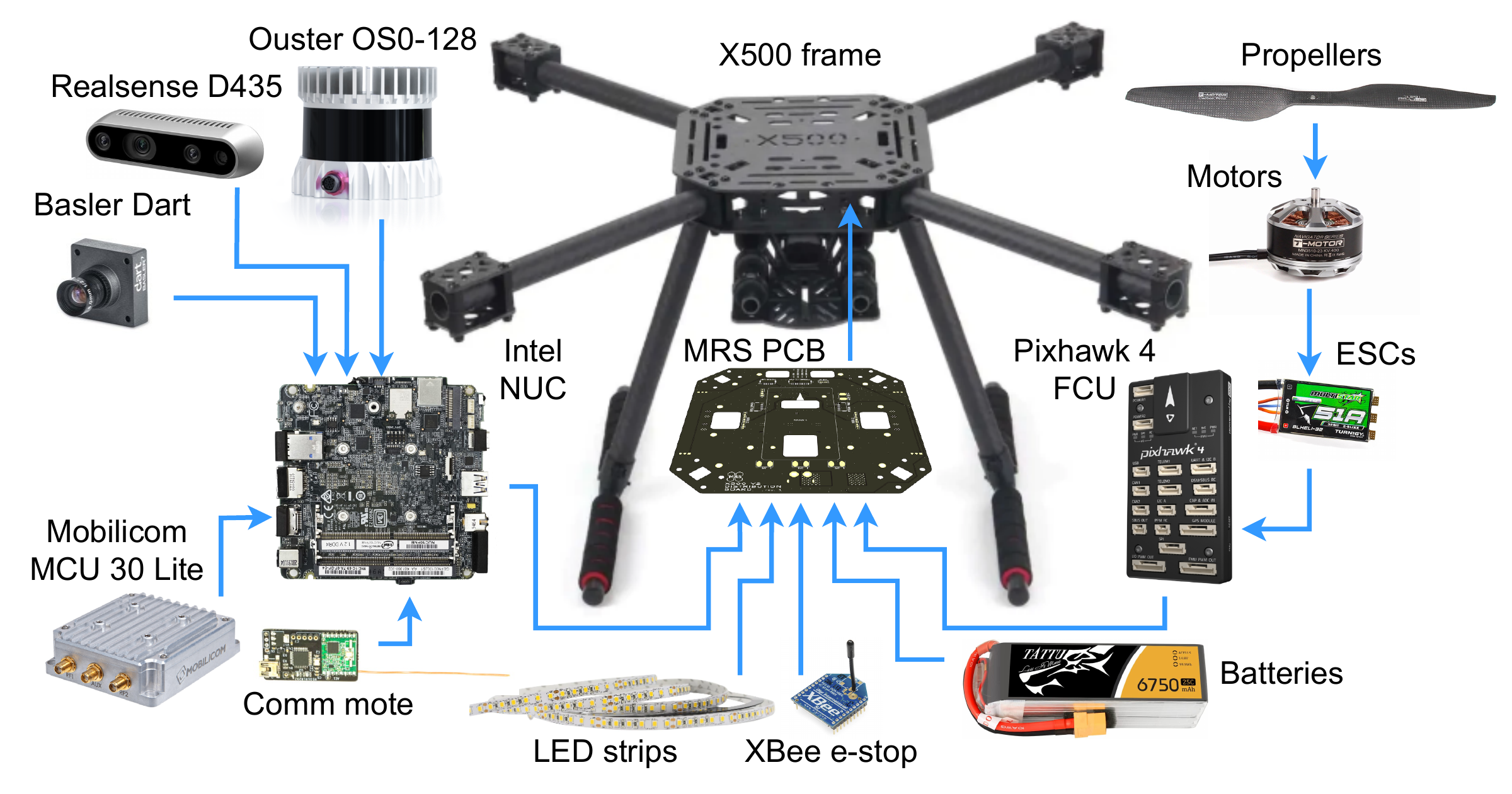}};%
    \end{tikzpicture}
    \caption{The most important components of the X500~\ac{UAV} platformed designed for the~\ac{darpa}~\ac{subt} Challenge.
    Blue arrows show the connections between individual modules.}
    \label{fig:darpa_hw}
    % \vspace{-1.5em}
\end{figure}

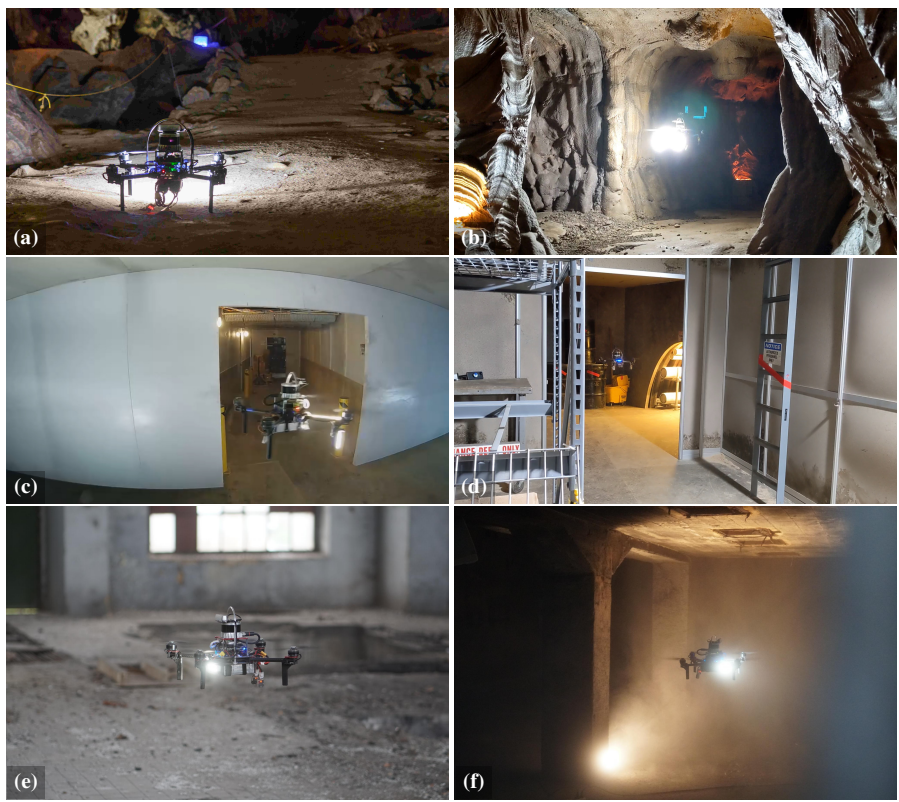
\begin{figure} [tb]
    \newcommand{\imheight}{10.50em}
    \newcommand{\xcap}{0.95em}
    \newcommand{\ycap}{0.8em}
    \newcommand{\fillopa}{0.3}
    \centering
    \begin{tikzpicture}
      \node[anchor=south west,inner sep=0] (b) at (0,0) {\adjincludegraphics[width=0.49\textwidth,trim={{0.00\width} {0.0\height} {0.00\width} {0.157\height}},clip]{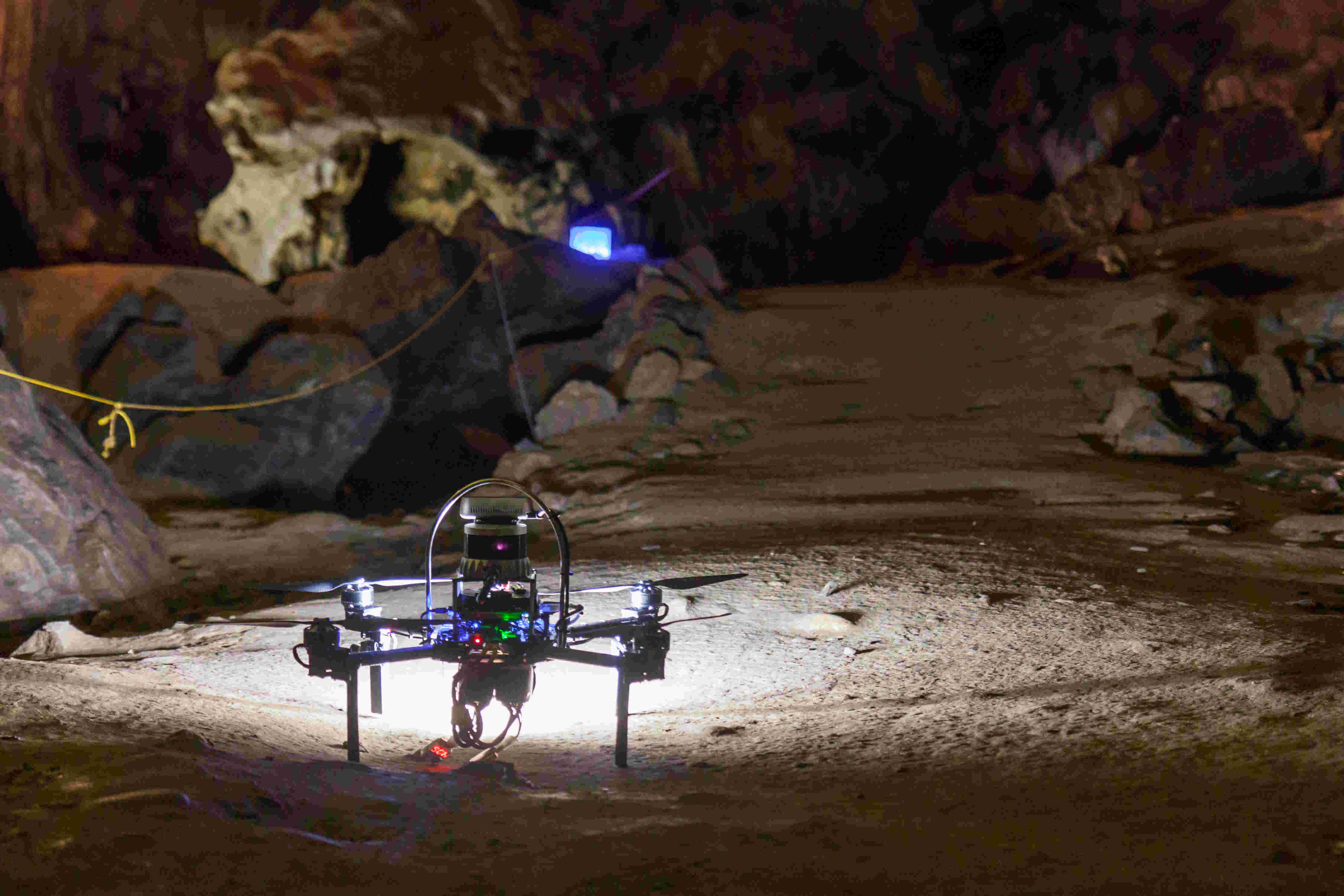}};%
    \begin{scope}[x={(b.south east)},y={(b.north west)}]
  \node[fill=black, fill opacity=\fillopa, text=white, text opacity=1.0] at (\xcap, \ycap) {\textbf{(a)}};
      \end{scope}
    \end{tikzpicture}
    \begin{tikzpicture}
      \node[anchor=south west,inner sep=0] (b) at (0,0) {\adjincludegraphics[width=0.49\textwidth,trim={{0.00\width} {0.0\height} {0.00\width} {0.00\height}},clip]{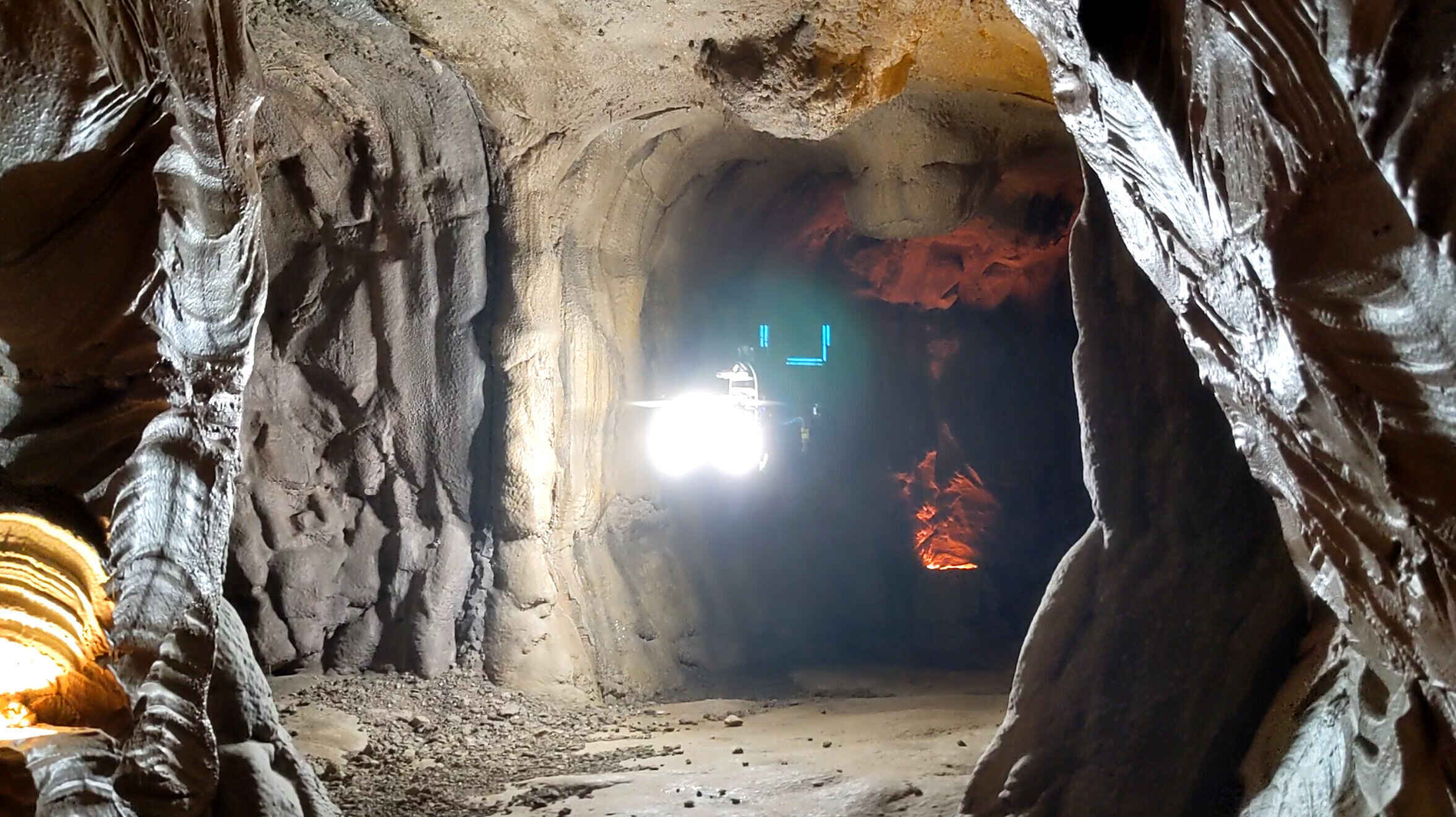}};%
    \begin{scope}[x={(b.south east)},y={(b.north west)}]
  \node[fill=black, fill opacity=\fillopa, text=white, text opacity=1.0] at (\xcap, \ycap) {\textbf{(b)}};
      \end{scope}
    \end{tikzpicture}
    \begin{tikzpicture}
      \node[anchor=south west,inner sep=0] (b) at (0,0) {\adjincludegraphics[width=0.49\textwidth,trim={{0.00\width} {0.0\height} {0.00\width} {0.00\height}},clip]{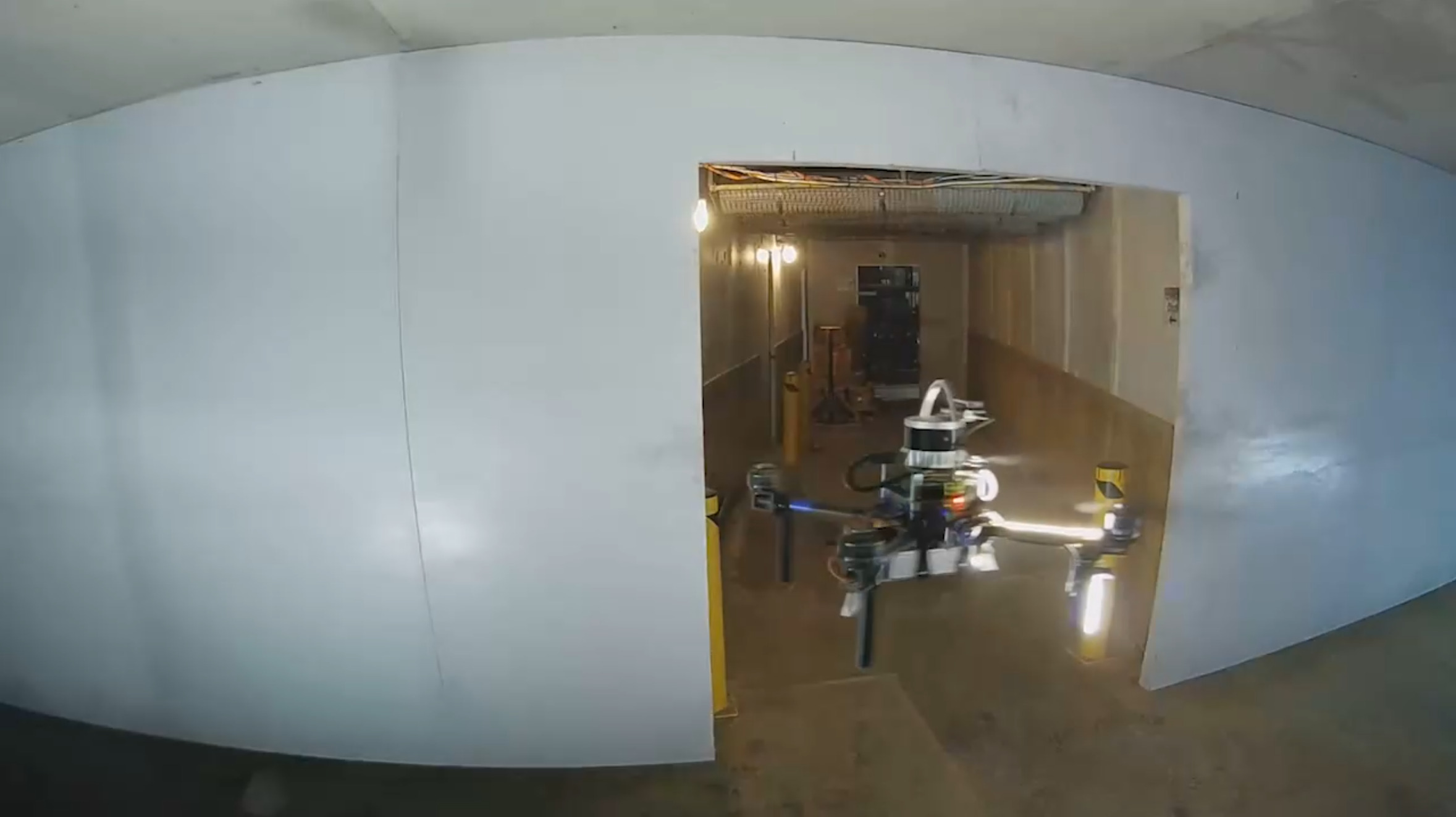}};%
    \begin{scope}[x={(b.south east)},y={(b.north west)}]
  \node[fill=black, fill opacity=\fillopa, text=white, text opacity=1.0] at (\xcap, \ycap) {\textbf{(c)}};
      \end{scope}
    \end{tikzpicture}
    \begin{tikzpicture}
      \node[anchor=south west,inner sep=0] (b) at (0,0) {\adjincludegraphics[width=0.49\textwidth,trim={{0.00\width} {0.0\height} {0.00\width} {0.00\height}},clip]{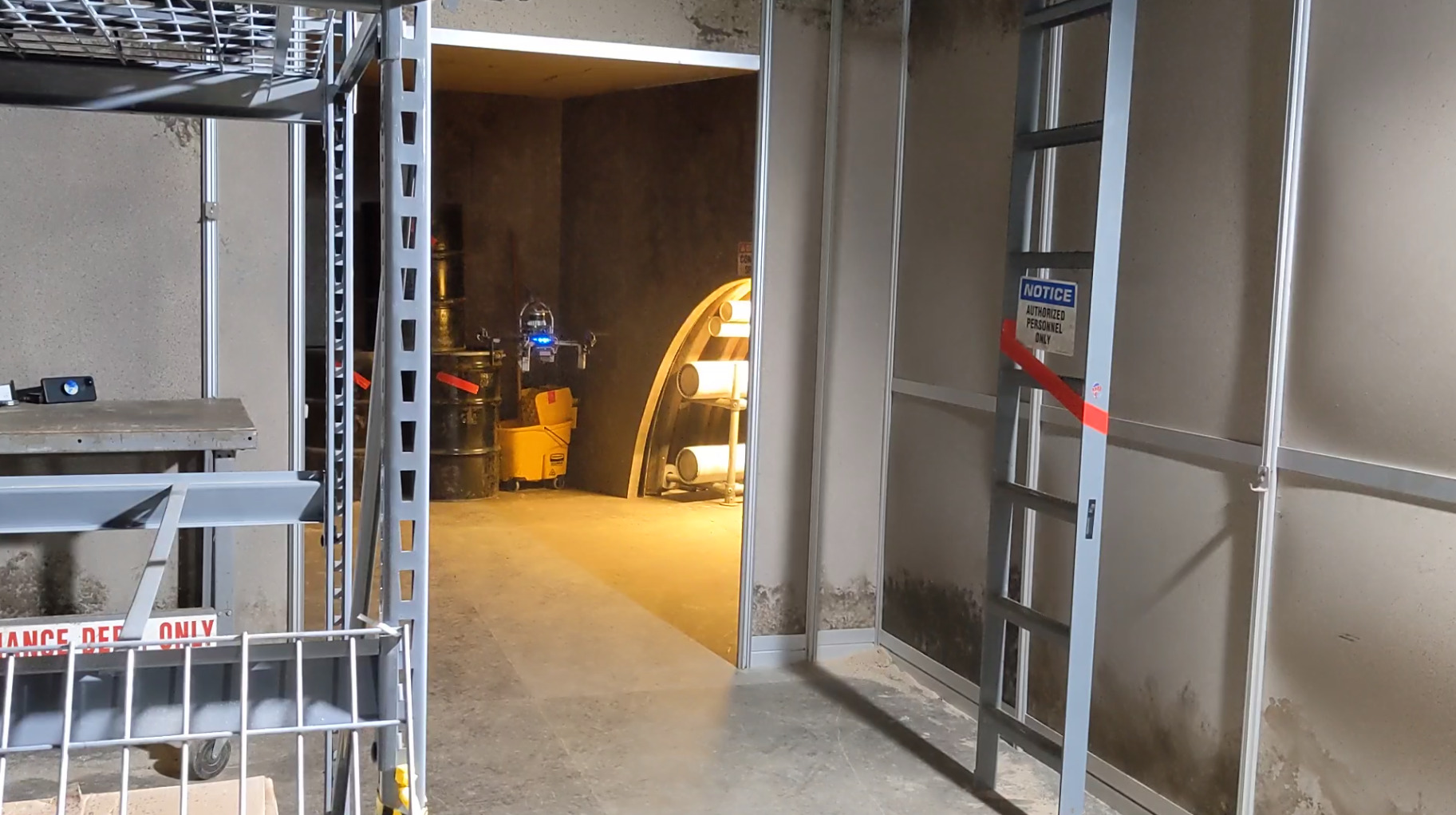}};%
    \begin{scope}[x={(b.south east)},y={(b.north west)}]
  \node[fill=black, fill opacity=\fillopa, text=white, text opacity=1.0] at (\xcap, \ycap) {\textbf{(d)}};
      \end{scope}
    \end{tikzpicture}
    % \begin{tikzpicture}
    %   \node[anchor=south west,inner sep=0] (b) at (0,0) {\adjincludegraphics[height=\imheight,trim={{0.00\width} {0.0\height} {0.00\width} {0.00\height}},clip]{./fig/x500/finals_mine_intersection.jpg}};%
    % \begin{scope}[x={(b.south east)},y={(b.north west)}]
  % \node[fill=black, fill opacity=\fillopa, text=white, text opacity=1.0] at (\xcap, \ycap) {\textbf{(e)}};
    %   \end{scope}
    % \end{tikzpicture}
    \begin{tikzpicture}
      \node[anchor=south west,inner sep=0] (b) at (0,0) {\adjincludegraphics[width=0.49\textwidth,trim={{0.00\width} {0.0\height} {0.00\width} {0.00\height}},clip]{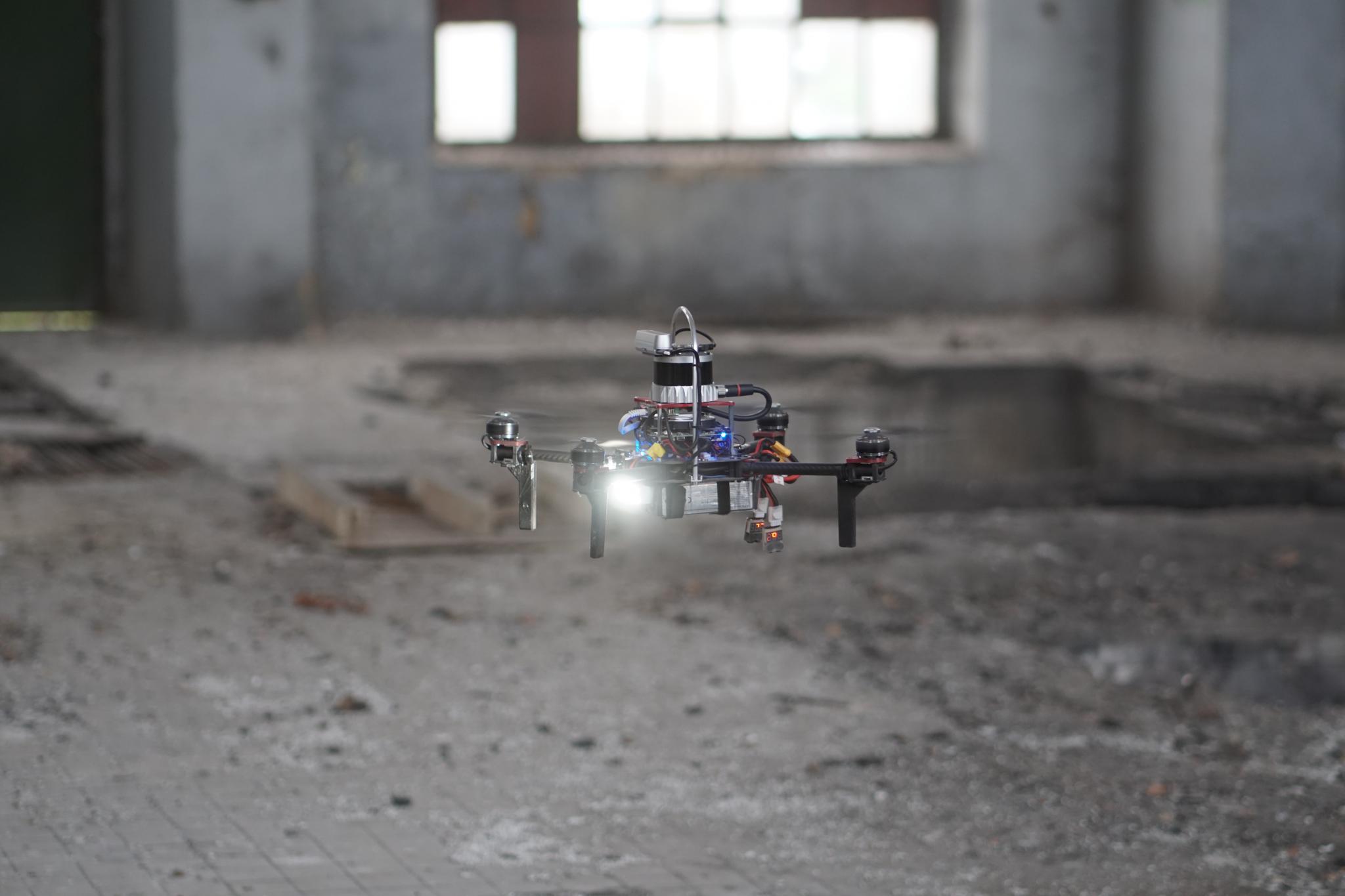}};%
    \begin{scope}[x={(b.south east)},y={(b.north west)}]
  \node[fill=black, fill opacity=\fillopa, text=white, text opacity=1.0] at (\xcap, \ycap) {\textbf{(e)}};
      \end{scope}
    \end{tikzpicture}
    \begin{tikzpicture}
      \node[anchor=south west,inner sep=0] (b) at (0,0) {\adjincludegraphics[width=0.49\textwidth,trim={{0.00\width} {0.0\height} {0.00\width} {0.00\height}},clip]{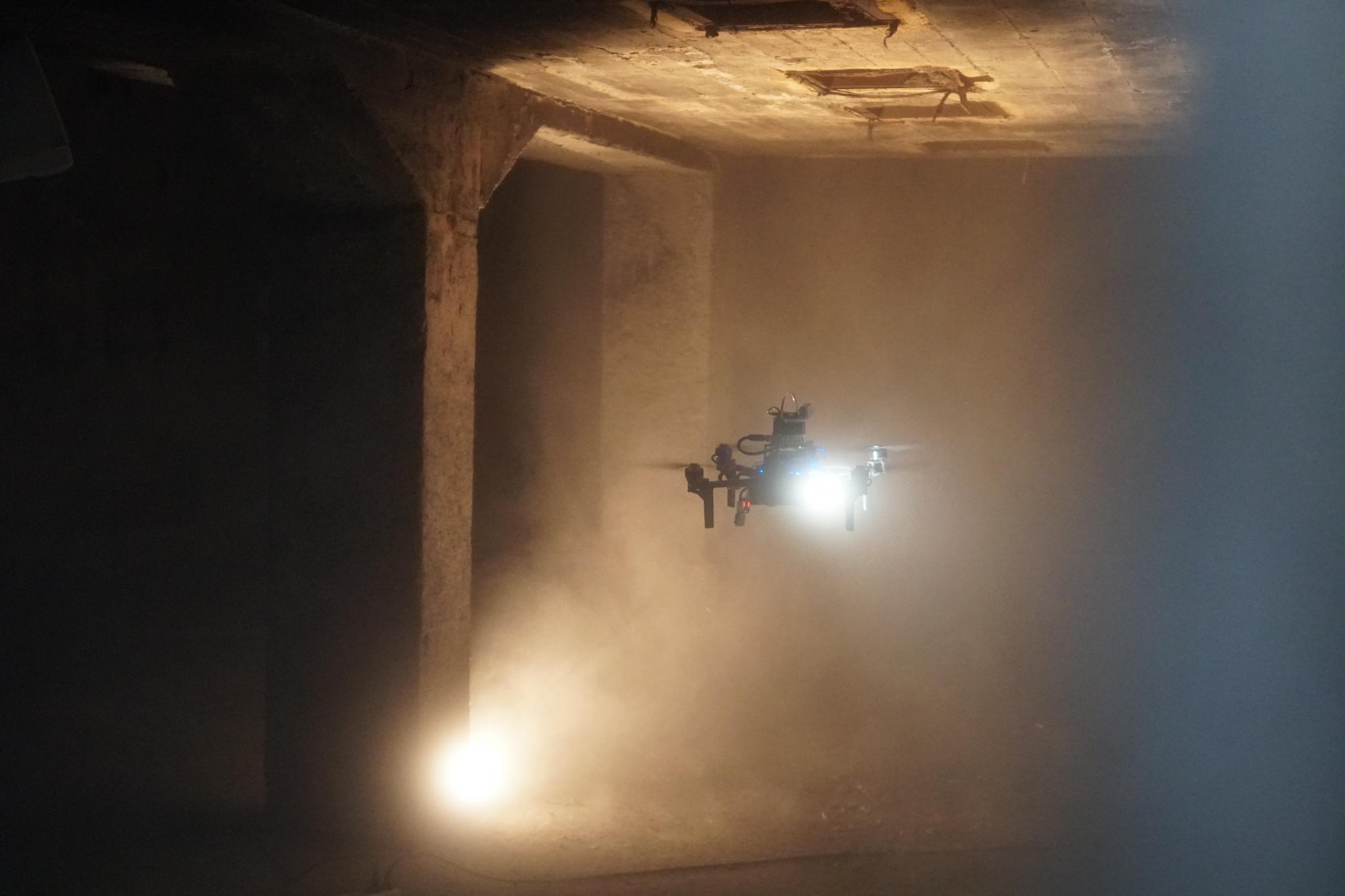}};%
    \begin{scope}[x={(b.south east)},y={(b.north west)}]
  \node[fill=black, fill opacity=\fillopa, text=white, text opacity=1.0] at (\xcap, \ycap) {\textbf{(f)}};
      \end{scope}
    \end{tikzpicture}
    \caption{Verification of the X500 platform in the following environments: dark and humid caves (a,b), man-made urban structures (c,d), and dusty ruins of industrial buildings (e,f).}
    \label{fig:field_testing}
  \end{figure}

Figure~\ref{fig:darpa_hw} shows the X500 platform with the most important components.
The platform was also replicated as a simulation model in the virtual track of the~\ac{darpa} \ac{subt}, where it was deployed by most of the teams and was also part of the winning solution.
%Our robotic team consisting of 5~\acp{UAV} and 2~\acp{UGV} achieved second place in the virtual track, and sixth place in the competition with physical robots, where 3~\acp{UAV} together with 6~\acp{UGV} found 7 artifacts in a complex confined environment.
In total, our team was composed of five~\acp{UAV} and two~\acp{UGV}.
We finished second in the virtual track, and we achieved sixth place in the competition with physical robots, where three~\acp{UAV} together with six~\acp{UGV} found seven artifacts in a complex confined environment.
The developed approach is extensively covered by~\cite{petrlikDarpaFinals, ebadi2022slam}.
The platform also proved its reliability in intensive tests in intricate environments, such as vast natural caves, ruins of industrial buildings, underground fortresses, mining tunnels, and large-scale outdoor environments with dense vegetation.
Some of the environments where the platform was experimentally verified are shown in Fig.~\ref{fig:field_testing}.

% %%{ FIGURE: Field testing
% %%}
% %%}

%%% END SECTION ============================================================

%%% START SECTION ==========================================================

%%[OWNER]: Pavel and Vit
\subsubsection{Dronument: documentation of historical monuments}
\label{sec:dronument}

The documentation of difficult-to-access areas of valuable historic buildings puts special emphasis on the robustness and reliability of the deployed systems. The realization of this task in complex environments requires minimizing dimensions, while still providing sufficient payload capacity for all necessary sensory equipment for operation in dark conditions. 
The~\ac{MRS} hardware stack includes platforms designed specifically for autonomous cooperative high-resolution photography in the interiors of buildings based on more than two years of experiments with the preliminary testing platforms.
The developed primary platform is an eight-rotor helicopter with a coaxial arrangement of rotors and a custom frame composed of aluminum profiles connected by carbon-fiber elements, including parts preventing collisions of the propellers with the environment.  

\begin{figure}[tb]
    \centering
    \input{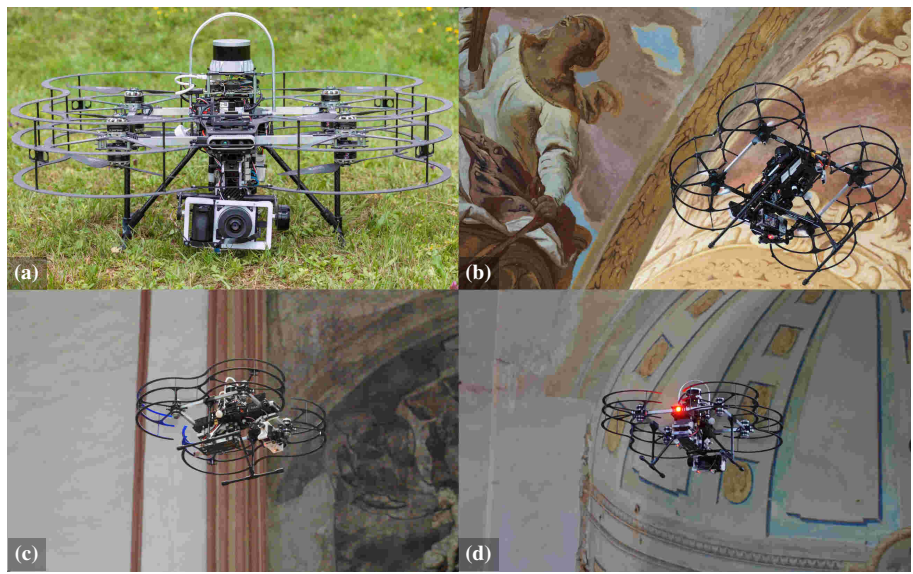}
    \vspace{-5mm}
    \caption{Platform (a) tailored for documentation of structures and valuables within interiors of historical monuments (b-d). Videos: \url{https://youtu.be/-_1Fjr58a28} and~\url{https://youtu.be/wyJdtzas1RI}.}
    \label{fig:dronument}
\end{figure}

The equipment of the platform comprises a powerful computational unit Intel NUCi7 onboard computer and a wide range of sensors to maximize the onboard perception and allow for sensory redundancy.
The platform further disposes of a dual-axis gimbal for stabilization of the main documentation sensor, with a weight of up to \SI{700}{\gram} --- usually, a camera operating in the visible spectrum.
However, the high modularity of the system allows for its replacement by, e.g.,~\ac{UV}, IR, or a multi-spectral camera.
To facilitate the process of autonomous image capturing, the camera and stabilization device are connected to the main distribution board, providing an interface to control the shutter trigger and tilt of the camera from an onboard computer.
In order to support a human supervisor with information about proximity to obstacles, the status of failure detection systems, and mission diagnostics, the~\ac{UAV} is equipped with a powerful onboard LED with adjustable colors that encode the health status of the~\ac{UAV} (see Fig.~\ref{fig:dronument}(d)).

\begin{figure}[tb]
    \centering
    \input{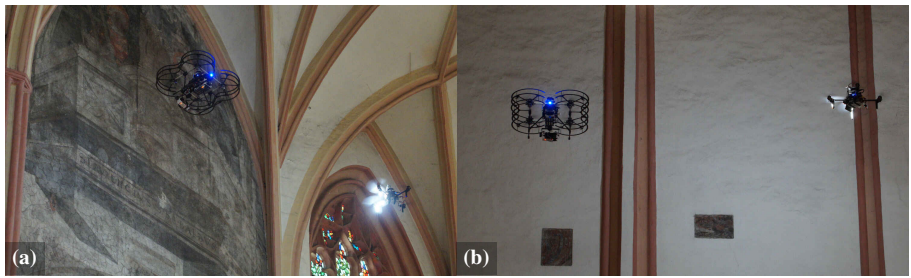}
    \vspace{-5mm}
    \caption{Deployment of cooperating platforms tailored for the realization of advanced documentation techniques in difficult-to-access areas of historical buildings. Video:~\url{https://youtu.be/-_1Fjr58a28}}
    \label{fig:dronument_cooperative}
\end{figure}

In several documentation techniques used by experts in the field of restoration and historical sciences, the primary platform must be coupled with a tightly cooperating secondary~\ac{UAV} to allow for positioning light at a predefined angle with respect to the scanned object.
Lower requirements on the payload capacity of the secondary platform allow for the use of smaller~\ac{UAV} based on the X500 frame, similar to the platform presented in Section~\ref{sec:darpaSubT}, equipped with a powerful directional light source (see Fig.~\ref{fig:dronument_cooperative}).
The developed platforms enable the implementation of advanced documentation techniques in hard-to-reach places in proximity to scanned objects. Thus, they collect valuable data for monitoring the state of historic objects and restoration works planning.

The main goal of the Dronument\footnote{\url{http://mrs.felk.cvut.cz/dronument}} project, 
whose results are summarized in~\cite{saska2017documentation, petracek2020dronument, 
kratky2020autonomous, kratky2021documentation, smrcka2020admittance, Bednar2022ICUAS, 
petracek2023dronument}, is the deployment of safety-critical platforms (see 
Fig.~\ref{fig:dronument}) as part of a~\ac{UAV} team for the documentation of historical 
buildings. The presented platforms have already been deployed in 15 historic objects, 
including two UNESCO World Heritage sites. During this long-term deployment, the system 
collected almost 11,000 images in more than 200 flights in historical buildings of various 
dimensions and characteristics. 

%TODO Vit, Pavel 
%- half of the column of text, maximum. Plus two small pictures of drones, link to videos (even all of them), and cite all related papers. The goal is to show how many experiments we did with the MRS HW system.
%%% END SECTION ============================================================

%%% START SECTION ==========================================================

%%[OWNER]: Matous
\subsubsection{Industrial inspection}
\label{sec:industrialInspection}

\begin{figure}[tb]
    \centering
    \input{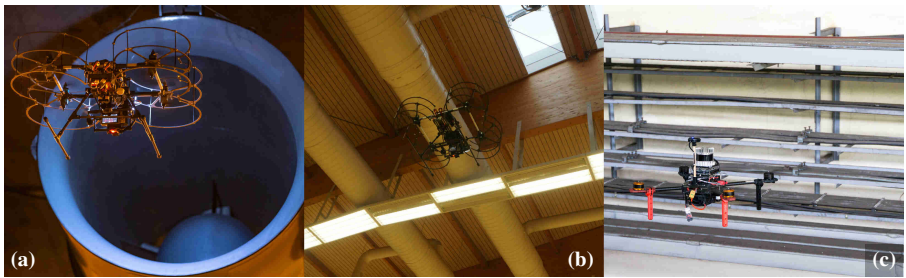}
    \vspace{-5mm}
    \caption{Snapshots of deployment of industrial platforms (a,b) and a general research platform (c) for inspection of ventilation systems, structural degradation, and electrical infrastructure in tunnels, halls, and storage houses. Video:~\url{https://youtu.be/60nKXamV2ds}.}
    \label{fig:industry_indoor_inspection}
    \vspace{-2em}
\end{figure}

Indoor inspection tasks have similar requirements on robot capabilities as the documentation of historical monuments.
Similarly equipped platforms are typically deployed with onboard~\ac{lidar} sensors for obstacle detection, as well as self-localization and high-resolution cameras for inspection.
However, in contrast to historical buildings, indoor industrial environments, such as maintenance shafts, manufacturing halls, or storage bays, often consist of repetitive and feature-sparse patterns (both visual and geometrical) with small and difficult-to-detect obstacles (e.g., ropes, cables, poles).
A higher resolution sensor onboard the~\ac{UAV} and specialized self-localization and obstacle detection algorithms must be employed to tackle these challenges.
Therefore, the~\ac{UAV} platforms require a trade-off between high payload capacity for carrying these sensors and maintaining a sufficiently compact form to pass through narrow corridors.
The risk of collisions with environmental obstacles can also be addressed with the mechanical protection of propellers and sensors.
Examples of the deployment of our custom~\ac{UAV} in these scenarios are showcased in Fig.~\ref{fig:industry_indoor_inspection}.

%%% END SECTION ============================================================

%%% START SECTION ==========================================================

%%[OWNER]: Tiago and Jiri Horyna
\subsection{Outdoor real robot experiments}
\label{sec:outdoorRealRobotExperiments}

For outdoor experimentation, the~\ac{MRS} platforms possess a versatile sensory system that allows the~\ac{UAV} to vary between full reliance on GNSS data, to deploying a multi-robot system in a GNSS-denied area (e.g., forests). In this subsection, we demonstrate the adaptability of the~\ac{MRS} platforms for use in a diverse range of outdoor scenarios and applications, including swarming in plain fields, desert, forest, and grass hill environments, power-line monitoring, human-robot interaction, and marine applications. In order for the robotic platforms to be used in such expansive environments, modifications on the frame must be performed. Such modifications can include different landing gears, water-proof encapsulation, sensor-fusions techniques, communication shielding, and so on.

%%% END SECTION ============================================================

%%% START SECTION ==========================================================

\begin{figure}[h]
    \centering
    \input{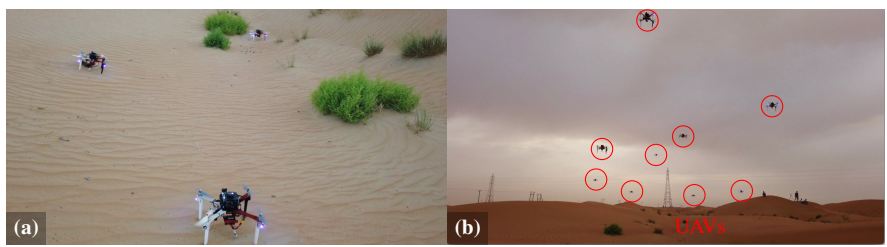}
    % \vspace{-5mm}
    \caption{Swarms of~\acp{UAV} in the desert (a) using~\ac{MRS} platform and 3D formation of 10~\acp{UAV} (b). Video:~\url{https://youtu.be/o8bphtbPCaA}.}
    \label{fig:swarms}
\end{figure}

%%[OWNER]: Tiago and Jiri Horyna
\subsubsection{Swarming in deserts, hills, and forests}
\label{sec:swarmingDesert}

The swarm is a multi-robot configuration in which robots are virtually merged to form a compact group (see Fig.~\ref{fig:swarms}) of dispensable and interchangeable robots~\cite{RODRIGUEZ2021, WU2022}.
Biological swarms take advantage of a high number of group members to travel long distances and protect individuals from predators. 
In aerial robotics, swarms of~\acp{UAV} increase their efficiency, scalability, and reliability in applications, such as environmental exploration and monitoring or search and rescue operations. 
Nevertheless, bringing these swarm systems into the real world is a long-standing problem, due to the significant operative differences between laboratory environments and diverse, real-world environments.
To deal with the demands of various real-world environments, the~\ac{MRS} platforms can be modified to handle the operating conditions required in plain fields, forests, dunes and deserts, hills, and much more~\cite{ICUAS_Thulio, afzal2021Icra, dmytruk2020safe, petracek2020bioinspired, novak2021fast, horyna2022sar, Akash2022ICUAS, PRITZL2023104315}.
One essential element of biologically-inspired swarming architectures is the perception-aware system, which allows for maintaining group cohesion and avoiding collisions between teammates. 
Our technological solution to this problem lies in using a communication-less mutual localization system called~\ac{UVDAR}. 

\begin{figure}[tb]
    \centering
    \input{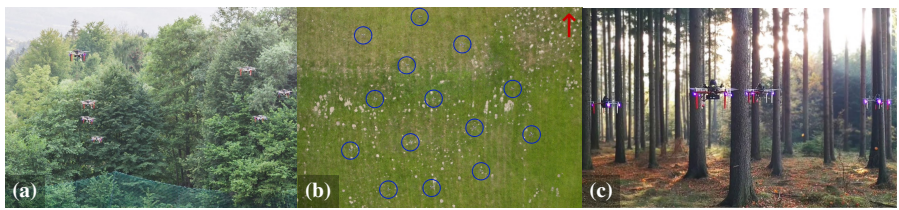}
    %\vspace{-5mm}
    \caption{Swarms of~\acp{UAV} in grass hill (a, b) and forest environments (c). Video:~\url{https://youtu.be/HH78AheC-DM}.}
    \label{fig:swarms2}
    %\vspace{-1em}
\end{figure}

\begin{figure}[tb]
    \centering
    \includegraphics[width=0.70\columnwidth]{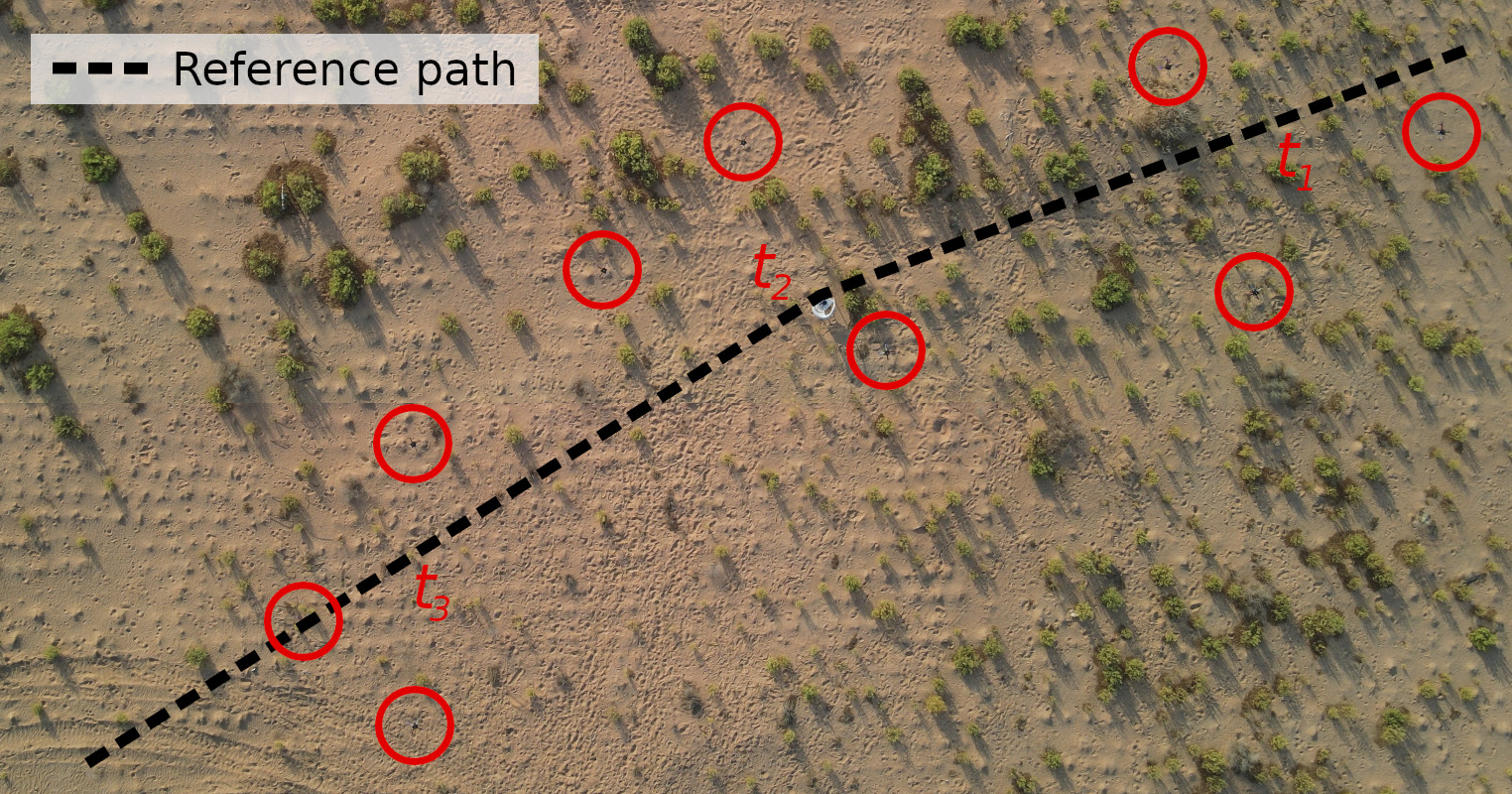}
    \caption{Swarm experiment in a desert environment using the optical flow method with a monocular camera for self-localization and supported by a novel multi-robot state estimator. Video:~\url{http://mrs.felk.cvut.cz/iros-2022-estimation}.}
    \label{fig:estimation}
\end{figure}

The swarm systems introduced in this section are fully decentralized, and thus the behavior of individual~\ac{UAV} is influenced only by the processing of onboard sensory information.
Using a 2D~\ac{lidar}, a swarm of~\acp{UAV} was able to navigate through a dense forest in~\cite{afzal2021Icra, Krizek2022ICUAS, dmytruk2020safe}.
The swarming solution in~\cite{afzal2021Icra} used a bio-inspired control approach, where a new approach for avoiding collisions with agents and obstacles to emphasize the safety of the~\ac{UAV} was introduced. 
In~\cite{Krizek2022ICUAS}, shared obstacle maps were used to perceive neighboring agents. This method showed better results than commonly used vision-based perception systems since the method from~\cite{Krizek2022ICUAS} is not influenced by the occlusion of obstacles.
A similar flocking model without any communication was introduced in~\cite{petracek2020bioinspired}. Moreover, we provided an analysis of the stability of the swarm concerning the accuracy of position estimation. 
A compact group of~\ac{MRS}~\acp{UAV} was deployed in a desert environment~\cite{horyna2022sar} for search and rescue applications with a CNN camera detector for human-victim detection in a control loop. These experiments also verified~\ac{MRS} platforms under demanding light and temperature conditions. A similar experimental verification was performed on a grassy hill, as shown in Fig.~\ref{fig:swarms2}, where the swarm system demonstrated the ability to follow over uneven terrain.
In~\cite{novak2021fast}, a bio-inspired evasion approach in a self-organized swarm of~\acp{UAV} was introduced. Using the~\ac{UVDAR} system,~\ac{MRS}~\acp{UAV} were able to avoid dynamic objects (predators) that were actively approaching the group. 
In the same line,~\cite{Ahmad2022Bioinspired} proposes an approach for achieving decentralized collective navigation of~\ac{UAV} swarms without communication and without global localization infrastructure. The work addresses several challenges related to the real-world deployment of a swarm of~\acp{UAV} and their collective motion in a cluttered environment. Compared to previous works~\cite{afzal2021Icra, dmytruk2020safe}, the approach uses relative localization information, recorded over a finite time horizon, to overcome the challenges of occlusion of~\acp{UAV} by obstacles.
In~\cite{horyna2022estimation}, a decentralized multi-robot state estimation approach was introduced to support onboard state estimation performance by swarm aggregation. Using this unique technique,~\ac{MRS}~\ac{UAV} platforms were able to fly above the sand's surface using a visual type of odometry (see Fig.~\ref{fig:estimation}), which proved to be unreliable over the uniform surface in standalone mode.

%\textbf{TODO: Jiri} - half of the column of text, maximum. Plus 3 small pictures of drones, link to videos (even all of them) and cite all related papers. The goal is to show how many experiments we did with the MRS HW system.} - J.H. Done

%%% END SECTION ============================================================

%%% START SECTION ==========================================================

%%[OWNER]: Tiago
\subsubsection{Marine environment applications}
\label{sec:marineEnvironmentApplications}

Another application tackled by the~\ac{MRS} system is the employment of~\acp{UAV} in a marine 
environment~\cite{GuptaRAL2023, Brandao2022RAL, Nekovar2023RAL}.
Our~\acp{UAV} were modified by integrating the principles and theoretical backgrounds of the behaviors of a~\ac{UAV} and/or a swarm of~\acp{UAV}. New control, localization, and perception approaches were employed in a single robot and in a multi-robot system to enable them to be adapted to an unfriendly environment, especially in cases like~\acp{UAV} landing on water platforms.
Thus, we were able to successfully apply the~\ac{MRS} system in the realistic marine scenarios of surveillance, reconnaissance, search and rescue, and so on.
This success was achieved through the development of new techniques that enable the~\acp{UAV} to interact with~\acp{USV} and objects on water and to interact with the water environment from an aerial perspective.

\begin{figure}[h]
    \centering
    \input{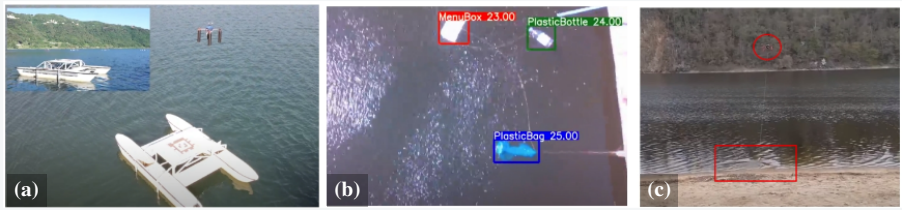}
    % \vspace{-5mm}
    \caption{Landing on a~\ac{USV} (a), detecting objects on water (b), and pulling debris out of water. Video:~\url{http://mrs.felk.cvut.cz/ral-landing-on-usv}.}
    \label{fig:marine}
\end{figure}

Among the applications investigated by our group are the physical interaction of a~\ac{UAV} with a~\ac{USV}, object detection on water, and the picking of debris on the water's surface by~\acp{UAV} (see Fig.~\ref{fig:marine}).
Furthermore, we developed a new~\ac{UAV} platform that contains a new set of sensors with the proper mechanical adaptation needed for such applications in the real world.
This new platform was based on the Tarot T650 frame (see Fig.~\ref{fig:marinedrone}) and contained floating gear to ensure safety when landing on water.
In addition, all sensory equipment was optimized to enable this~\ac{UAV} to autonomously fly over marine surfaces, such as rivers, lakes, and seas, within any weather conditions (e.g., rain, wind, waves).

%%% END SECTION ============================================================

%%% START SECTION ==========================================================

\begin{figure}[h]
\vspace{0.3cm}
    \centerline{\includegraphics[width = 0.67\textwidth]{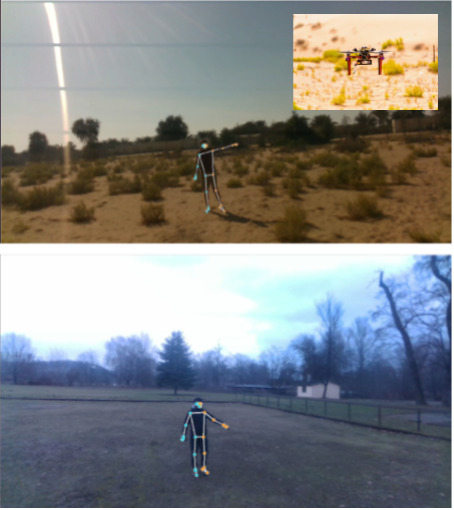}}
    \caption{Human operator performing actions in a sunny outdoor setting (top) and low-light outdoor setting (bottom).}
\label{fig:ihm}
\end{figure}

%%[OWNER]: Tiago
\subsubsection{Human-Robot Interaction}
\label{sec:IHM}

\ac{HRI} systems aim to ease the interaction between humans and robots. This is a growing area of research with open problems, especially when subjected to real-world conditions. The~\ac{HRI} using~\acp{UAV} makes the~\ac{HRI} problem even harder.
Control systems applied to~\ac{HRI} are also difficult to design due to the fact that they must be intuitive for the user.
Recently, \ac{UAV} swarms are being used in~\ac{HRI} problems. One recently published 
work~\cite{Jang2021} proposed a giant virtual architecture to enable the operator to control a 
swarm of robots in different ways, preserving the system's autonomy while enabling overall 
control of the mission objective. Another method found in the state-of-the-art of~\ac{HRI} 
techniques is to control a robot leader within a swarm as an avatar of the operator. The robot 
swarm would then follow the robot leader operated by a remote user~\cite{Ma2018}.

Recently, the~\ac{MRS} group has applied the~\ac{MRS} system to the~\ac{HRI} problem using a third method, in which the human operator controls a swarm of~\acp{UAV} using visual actions. Although gesture recognition is not new~\cite{Zhou2011, Ahmed2021}, the application of gesture recognition by the onboard cameras of multiple~\acp{UAV} with limited computational resources continues to be an open problem. The~\ac{MRS}~\ac{UAV} platforms can perform such action recognition and use it for the direct control of a~\ac{UAV} swarm (see Fig.~\ref{fig:ihm}).

%%% END SECTION ============================================================

%%% START SECTION ==========================================================

%%[OWNER]: Tomas and Robert
\subsubsection{\ac{MBZIRC} 2017 and 2020}
\label{sec:MBZIRC2017-2020}

% \textbf{TODO:  Tomas + Vojtech}}
%- 3/4 of the column of text, maximum. Plus 4 small pictures of drones (one for each challenge with a medal), link to videos (even all of them), and cite all related papers. The goal is to show how many experiments we did with the MRS HW system.

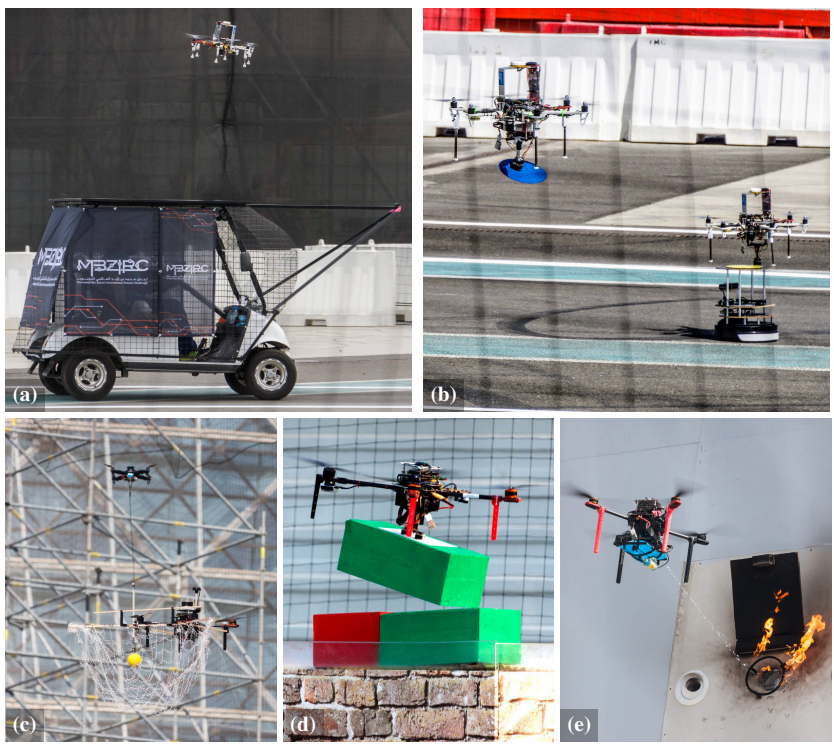
\begin{figure}[h]
    \centering
    \begin{tikzpicture}

    \pgfmathsetmacro{\vshift}{-4.37} % shift in vertical direction
    \pgfmathsetmacro{\hshiftone}{5.5} % shift in vertical direction
    \pgfmathsetmacro{\hshifttwo}{3.65} % shift in vertical direction
    \pgfmathsetmacro{\imgsizeone}{0.45}
    \pgfmathsetmacro{\imgsizetwo}{0.3}
    
    \node[anchor=south west,inner sep=0] (a) at (0.0,0) {
      \includegraphics[width=\imgsizeone\textwidth]{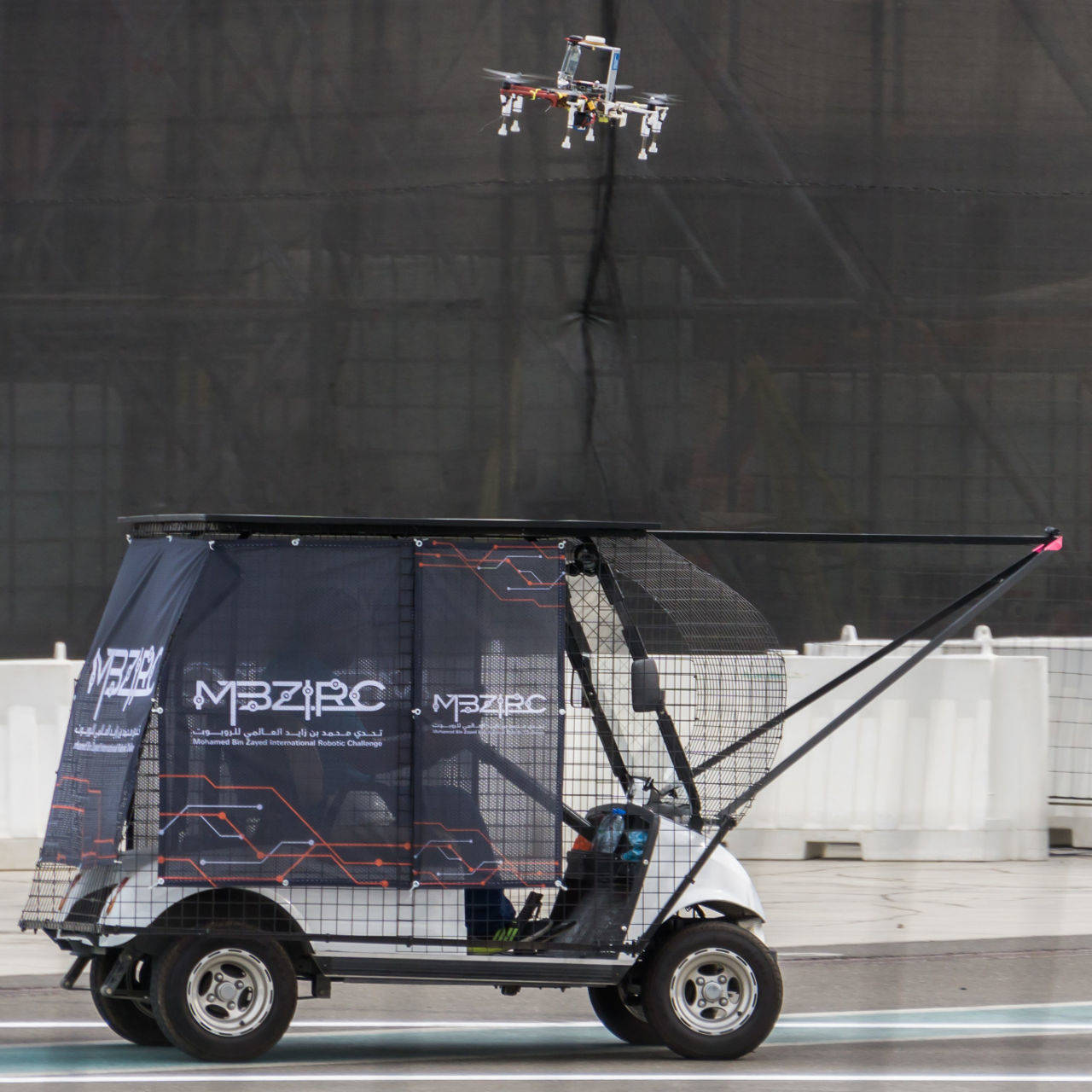}   
     };
    \node[anchor=south west,inner sep=0] (b) at (\hshiftone,0) {
      \includegraphics[width=\imgsizeone\textwidth]{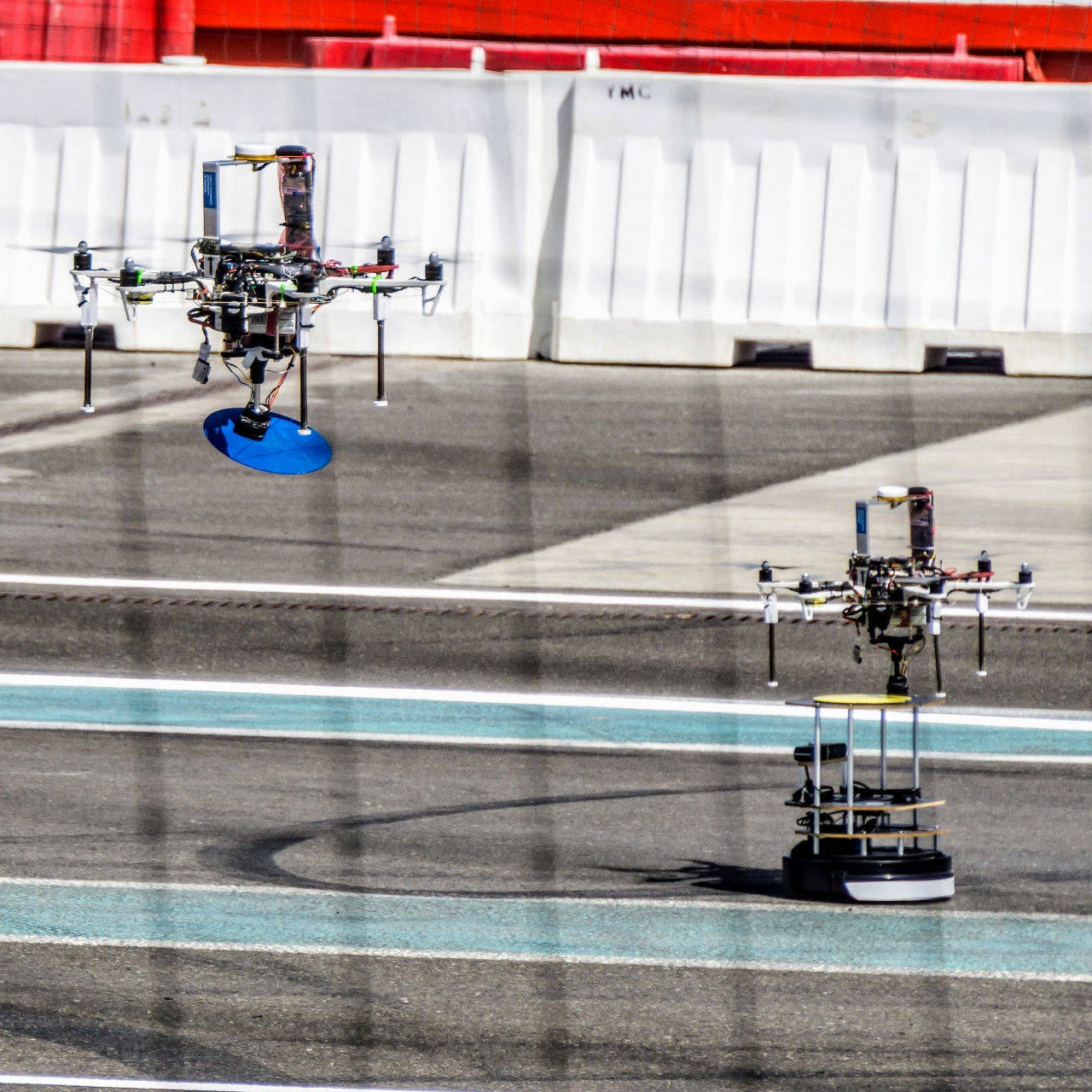}
    };
    \node[anchor=south west,inner sep=0] (c) at (0,\vshift) {
      \includegraphics[width=\imgsizetwo\textwidth]{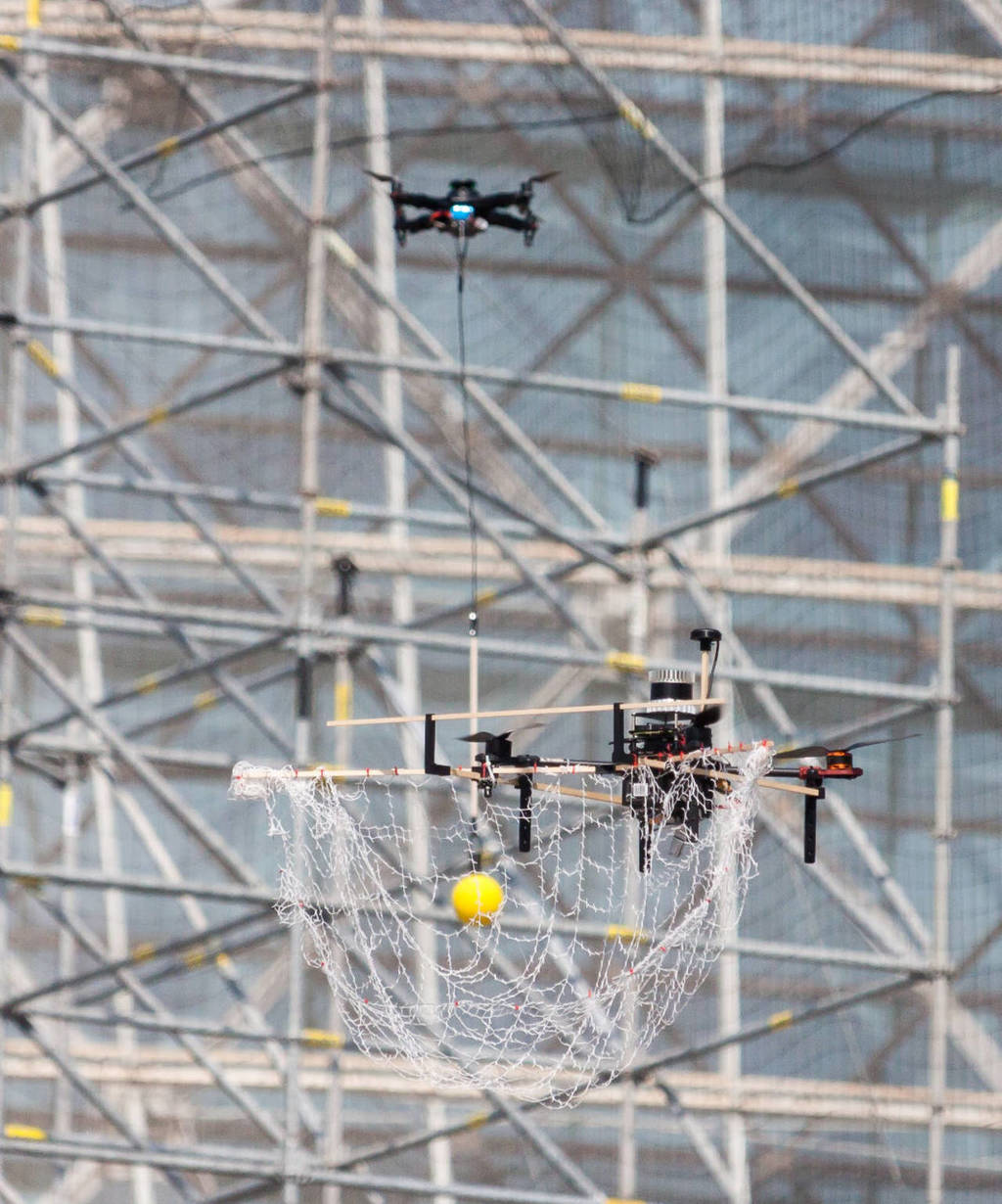}
    };
    \node[anchor=south west,inner sep=0] (d) at (\hshifttwo,\vshift) {
      \includegraphics[width=\imgsizetwo\textwidth]{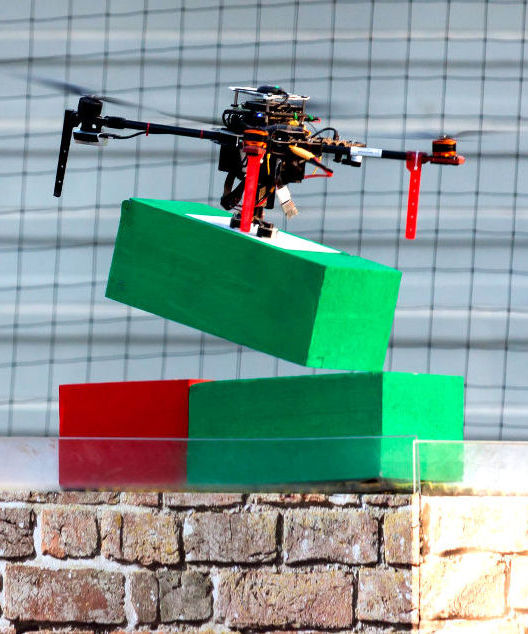}
    };
    \node[anchor=south west,inner sep=0] (e) at (2*\hshifttwo,\vshift) {
      \includegraphics[width=\imgsizetwo\textwidth]{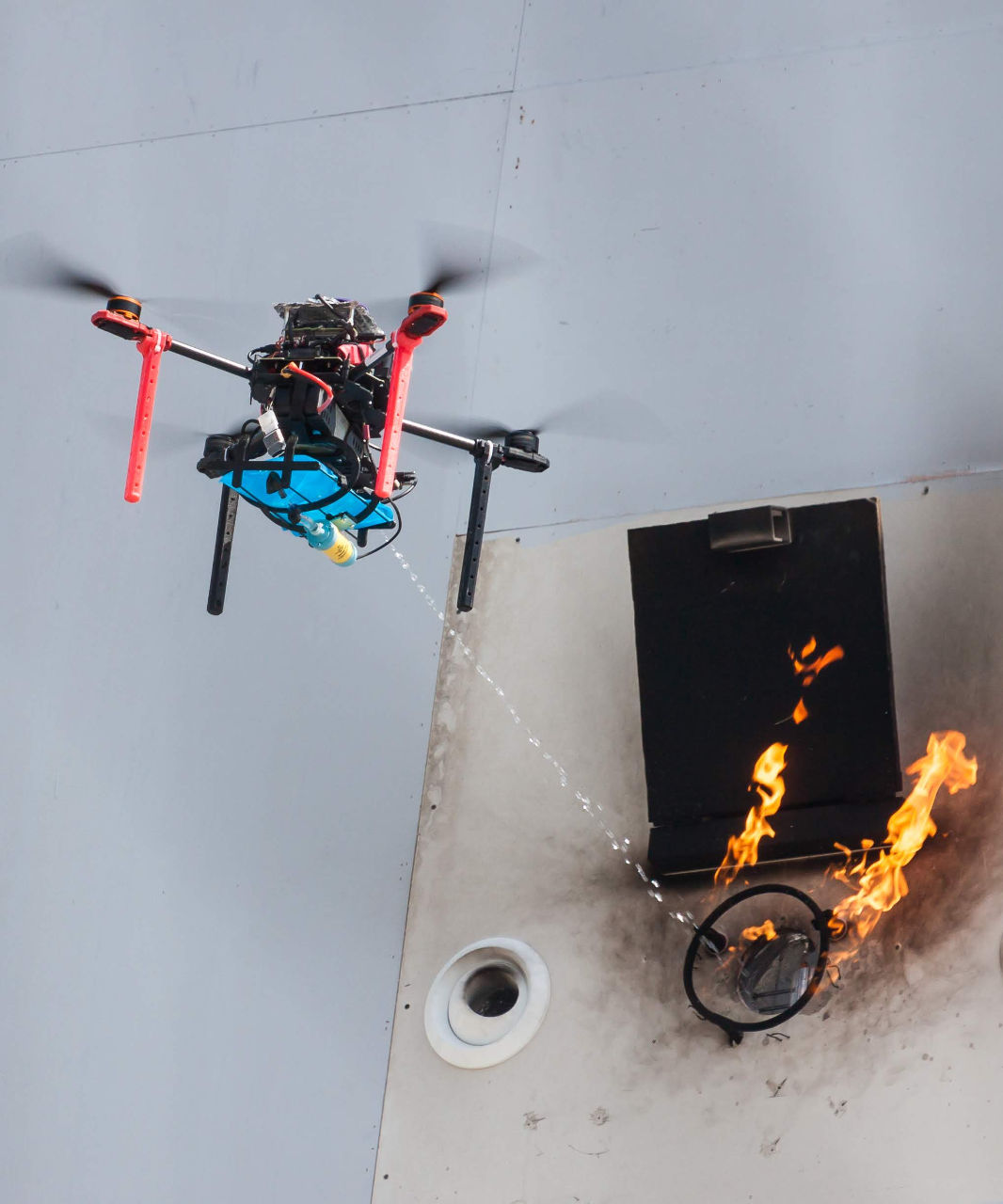}
    };
     
    \node[fill=black, fill opacity=0.3, text=white, text opacity=1.0, anchor=south west] at (a.south west) {\textbf{(a)}};
    \node[fill=black, fill opacity=0.3, text=white, text opacity=1.0, anchor=south west] at (b.south west) {\textbf{(b)}};
    \node[fill=black, fill opacity=0.3, text=white, text opacity=1.0, anchor=south west] at (c.south west) {\textbf{(c)}};
    \node[fill=black, fill opacity=0.3, text=white, text opacity=1.0, anchor=south west] at (d.south west) {\textbf{(d)}};
    \node[fill=black, fill opacity=0.3, text=white, text opacity=1.0, anchor=south west] at (e.south west) {\textbf{(e)}};
    
    \end{tikzpicture}
    % \vspace{-5mm}
    % \vspace{-0.5em}
    \caption{\ac{MBZIRC} 2017: autonomous landing on a moving vehicle (a) and cooperative collection of objects (b); \&~\ac{MBZIRC} 2020: autonomous capture of agile objects (c), cooperative wall building (d), and autonomous fire-fighting (e). Video:~\url{https://youtu.be/DEUZ77Vk2zE}.}
    \label{fig:mbzirc}
    \vspace{-1.5em}
\end{figure}

The~\ac{MBZIRC} 2017\footnote{\url{http://mrs.felk.cvut.cz/mbzirc}} and 2020\footnote{\url{http://mrs.felk.cvut.cz/mbzirc2020}} competitions were composed of various robotic challenges motivated by the intent to push technological and application boundaries beyond the current state-of-the-art (see Fig.~\ref{fig:mbzirc}). 
These technological challenges included fast autonomous navigation in semi-unstructured, complex, and dynamic environments with minimal prior knowledge, robust perception and tracking of dynamic objects in 3D, sensing and avoiding obstacles, GNSS-denied navigation in indoor-outdoor environments, physical interaction, complex mobile manipulations, and air-surface collaboration.

In~\ac{MBZIRC} 2017, the~\ac{MRS} platform participated in two challenges. 
Challenge~1 featured an autonomous landing on a moving vehicle (see Fig.~\ref{fig:mbzirc}(a)).
The task was to autonomously locate a vehicle, match its speed, and finally land on its roof, all of which used exclusively onboard sensors, such as a camera to detect its pattern and GPS for localization.
The~\ac{MRS} system achieved 2nd place in the Challenge~1\footref{footmbzirc}~\cite{baca2019autonomous,stepan2019vision}. 
The second challenge in~\ac{MBZIRC} 2017 that the MRS platform took part in was Challenge~3\footnote{\label{footmbzirc}\url{https://youtu.be/ogmQSjkqqp0}}.
This challenge featured a team of three~\acp{UAV} tasked with finding a set of static and moving colored ferromagnetic objects (see Fig.~\ref{fig:mbzirc}(b)) with consequent delivery of the objects to the target location.
Therefore, the~\acp{UAV} had to autonomously detect the objects with onboard cameras, be able to grasp the object with an electromagnetic gripper, and also coordinate with its multi-robot team to prevent collisions among the~\acp{UAV}.
The~\ac{MRS} platforms won Challenge~3~\cite{spurny2019cooperative, loianno2018localization}.

%Furthermore, the system achieved 2nd place in Challenge 1, where a single~\ac{UAV} had to autonomously localize a moving vehicle in the arena and land on  the vehicle\footref{footmbzirc}~\cite{baca2019autonomous,stepan2019vision}. 

In~\ac{MBZIRC} 2020, the~\ac{MRS} platform participated in all three challenges. 
The task of Challenge~1 was to autonomously track a flying object (enemy drone) and interact with it in order to catch a ball attached to the object (see Fig.~\ref{fig:mbzirc}(c)).
The MRS platforms took 2nd place in this challenge\footnote{\url{https://youtu.be/2-cLSjRCKDg}}~\cite{vrba_ras2022, stasinchuk2020multiuav, stasinchuk2020fr}.
In Challenge~2, the task was to autonomously build a wall using a team of three~\acp{UAV}\footnote{\url{https://youtu.be/1-aRtSarYz4}} and one ground vehicle~\cite{baca2020autonomous, stibinger2020mobile}.
The employed~\ac{MRS} platforms were able to place the largest number of bricks on the wall (see Fig.~\ref{fig:mbzirc}(d)), and therefore achieved 1st place in the challenge.
Next, Challenge~3 featured a fire-fighting task\footnote{\url{https://youtu.be/O8QBiAyP2c0}} as shown in Fig.~\ref{fig:mbzirc}(e)~\cite{jindal2020fr, walter2020fr, walter2020icuas, spurny2020autonomous, pritzl2021icra, Stibinger2020fr}.
Finally, the Grand Challenge combined all three challenges and was dominated by the~\ac{MRS}~\acp{UAV}, which collected the highest score among all participants to secure first place.

% In MBZIRC 2020, the system using MRS platforms achieved 1st place in Challenge~2 --- autonomous wall structure building by the team of three~\acp{UAV}\footnote{\url{https://youtu.be/1-aRtSarYz4}} and one UGV~\cite{baca2020autonomous,stibinger2020mobile}, and 2nd place in Challenge~1, where~\acp{UAV} should autonomously track and interact with a flying target\footnote{\url{https://youtu.be/2-cLSjRCKDg}}~\cite{vrba_ras2022, stasinchuk2020multiuav, stasinchuk2020fr}.
% The~\ac{MRS}~\ac{UAV} system had also been applied in the fire-fighting\footnote{\url{https://youtu.be/O8QBiAyP2c0}} Challenge~3 of MBZIRC 2020~\cite{jindal2020fr,walter2020fr,walter2020icuas,spurny2020autonomous,pritzl2021icra}. 

%%% END SECTION ============================================================

%%% START SECTION ==========================================================

\begin{figure}[tb]
    \centering
    \input{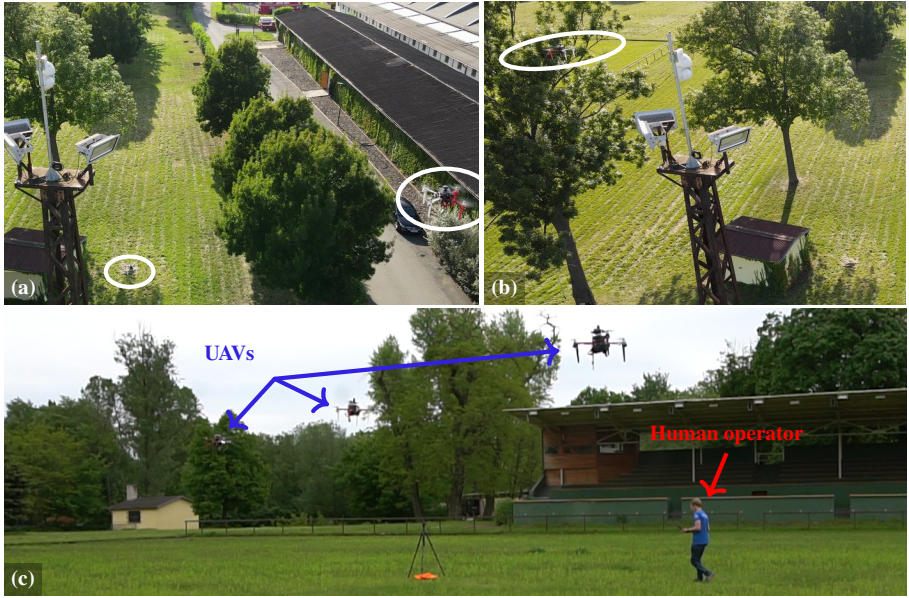}
    % \vspace{-5mm}
    % \vspace{-0.5em}
    \caption{Snapshots of the~\ac{MRS} drones performing \textit{inspection} and \textit{monitoring} operations in the AERIAL-CORE European project. A power line mock-up scenario is used to show the viability of the approach. Figures (a) and (b) show the \textit{inspection} scenario, while Fig.~(c) depicts the \textit{monitoring} operations. Solid circles are used to indicate the~\acp{UAV} approaching the tower. Video:~\url{https://youtu.be/ZqUCG7xqiE4}.}
    \label{fig:aerialCoreSnapshots}
    \vspace{-1em}
\end{figure}

%%[OWNER]: Giuseppe and Robert
\subsubsection{Aerial-Core}
\label{sec:aerialCore}

This section describes the results obtained within the H2020 European AERIAL-CORE\footnote{\url{https://aerial-core.eu}} project. 
The goal of the project is to develop cognitive aerial platforms for the application of autonomous power line inspection and maintenance.
This scenario is a typical example of~\ac{UAV} deployment in unknown and safety-critical environments where obstacles (e.g., branches, vegetation, balloons) may affect the autonomous navigation of the vehicles, and therefore the mission accomplishment. 
Two tasks of interest are considered: (i) \textit{inspection}, where a fleet of~\acp{UAV} perform a detailed investigation of power equipment by assisting a human operator in acquiring views of the power line (e.g., tower, cables, and insulators) that are not easily accessible, while also looking for damage to the mechanical structure and failures of electrical components, as depicted in Fig.~\ref{fig:aerialCoreSnapshots}(a) and Fig.~\ref{fig:aerialCoreSnapshots}(b); (ii) \textit{monitoring}, where a formation of~\acp{UAV} provides a view of the humans working on the power tower to their supervising team in order to monitor their status and ensure safety, as shown in Fig.~\ref{fig:aerialCoreSnapshots}(c). 

In both inspection and monitoring tasks, vision sensors are essential for visual examination and monitoring operations. 
In the~\ac{UAV} configuration, cameras are mounted in an \textit{eye-in-hand} configuration, i.e., rigidly attached to the body frame of the aircraft. 
Cameras are also used to mutually localize the~\acp{UAV} in the surrounding environment using the~\ac{UVDAR} system. 
The so-designed platform is suitable for both indoor and outdoor use, with an emphasis on onboard multi-sensor fusion. This allows vehicles to precisely detect their surroundings and flawlessly perform safe maneuvers in confined spaces.

The \acp{UAV} dedicated to the inspection task operate in a known environment represented by a previously acquired map, including the position of the power towers, cables, and vegetation.
Additionally, precise algorithms are used for~\ac{UAV} localization and navigation~\cite{baca2021mrs}.
The onboard controller encodes the mission execution as a task allocation and trajectory planning problem, where the vehicles need to move from an initial position through a sequence of target regions (places that need to be inspected) while avoiding obstacles and maintaining a safe distance between them~\cite{Silano2021RAL, CaballeroAR2022}.
The sequence of target regions is assigned to the~\acp{UAV} by solving a~\ac{VRP} so that all targets are visited by the~\acp{UAV} just once while minimizing the mission time.
At the same time, the~\acp{UAV} dealing with the monitoring operations can detect the human gestures~\cite{PapaioannidisEUSIPCO2021} used to provide high-level actions, such as requests for new tasks or new parameters for a previously requested task.
The onboard controller solves a leader-follower formation control problem where the vehicles need to keep the human worker within the camera field-of-view during the entire operation while providing complementary views from multiple~\acp{UAV}~\cite{Kratky2021RAL, Perez2022ICUAS}. 
Safety is ensured by computing safe corridors with integrated information from a local map and human tracking trajectories.
In both inspection and monitoring tasks, the~\acp{UAV} support maintenance operations of critical power grid infrastructure. 
Further details about the tasks, the designed algorithms, and the software architecture can be found in~\cite{Silano2021RAL, CaballeroAR2022, Kratky2021RAL, Silano2021ICUAS, Demkiv2021AIRPHARO, Nekovar2021RAL, Nekovar2022ETFA, Licea2021EUSIPCO, Calvo2022ICUAS, Silano2022ICUAS, CataffoSMC2022}. 
Illustrative videos of the experiments using the~\ac{MRS}~\ac{UAV} platforms are available\footnote{\url{https://mrs.felk.cvut.cz/projects/aerial-core}}.

%%% END SECTION ============================================================

%%% START SECTION ==========================================================

%%[OWNER]: Matous, Vojtech Spurny
\subsection{Industrial collaboration}
\label{sec:industrialCollaboration}

The \ac{MRS} hardware and software platforms have kick-started several industrial projects, serving as a basis for preliminary proof-of-concept prototypes and tests. These collaborations were followed by further development of specialized platforms.

%%% END SECTION ============================================================

%%% START SECTION ==========================================================

\begin{figure}[tb]
    \centering
    \input{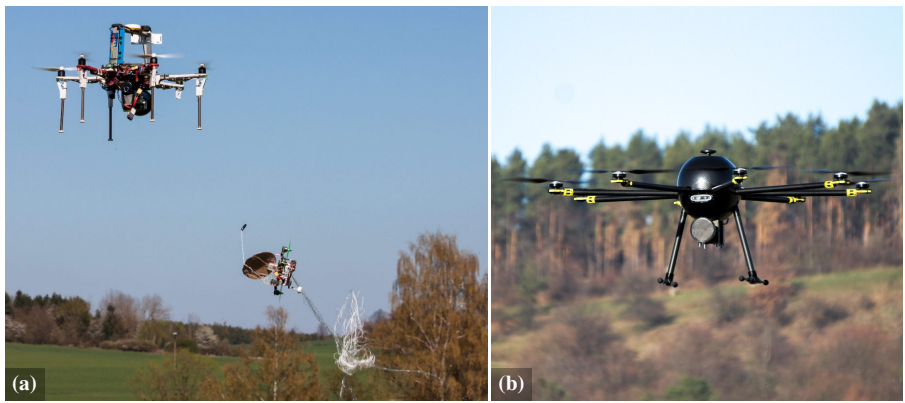}
    % \vspace{-0.5em}
    \caption{Development of the \textit{Eagle.One} autonomous aerial interception system based on the~\ac{MRS} hardware platform: (a) the first prototype using the~\ac{MRS}~\ac{UAV} platform during a successful interception; (b) the second prototype was designed by the~\ac{MRS} team specifically for the \textit{Eagle.One} project. Video:~\url{https://youtu.be/hEDGE7ofX1c}.}
    \label{fig:eagle}
    %\vspace{-1em}
\end{figure}

%%[OWNER]: Matous, Vojtech Spurny
\subsubsection{Airspace protection}
\label{sec:airspaceProtection}

The \textit{Eagle.One} autonomous aerial interception system is co-developed by our team as part of an industrial collaboration~\cite{vrba2019onboard, vrba2020markerless, vrba_ras2022, Vrba2023Journal}.
Several prototype platforms were designed as part of this project for evaluating the developed detection and interception algorithms (see Fig.~\ref{fig:eagle}).
The first prototype served only to test the trajectory planning for following and capturing a target using an onboard net launcher. It was based on our modular~\ac{UAV} platform utilizing the F550 frame and was equipped with standard onboard sensors, the PixHawk~\ac{FCU}, the NUC computer, an~\ac{RTK}-GPS for accurate positioning, a camera for detection of a visual marker placed on the target, and the net launcher.
The second iteration was designed with the capabilities of detecting a non-marked target and carrying the captured target in mind.
For this purpose, a specialized prototype platform based on the Tarot T18 with a payload capacity of \SI{12}{\kilo\gram} and a stereo camera for detection was developed.
Based on the results of these experimental tests, the "detection and catching" approach was reworked to rely on a 3D~\ac{lidar} sensor and a hanging net instead of the net launcher.
This enables the interceptor to retry failed catching attempts and provides a \SI{360}{\degree} field-of-view, which improves the re-detection of the target in such situations.
The final prototype is a custom-made \ac{UAV} octocopter platform with a 3D~\ac{lidar} sensor, a deployable net suspended below the interceptor, and a lightweight carbon frame to improve the carrying capacity and leave a sufficient thrust margin for dynamic maneuvering.
This prototyping approach enabled us to iteratively improve both the algorithms and hardware, and to effectively converge to a functioning system capable of solving the complex task of autonomous aerial interception.

%%% END SECTION ============================================================

%%% START SECTION ==========================================================

%%[OWNER]: Pavel Petracek
\subsubsection{Fire extinguishing}
\label{sec:fireExstinguish}

Multi-rotor UAVs are capable of quickly reaching great heights while carrying liquids, or extinguishants, that are able to extinguish fires located inside buildings.
When evaporated inside a fire, an extinguishant effectively pushes oxygen out of it, thereby removing its fuel and consequently extinguishing the fire. Furthermore, exterior-flying UAVs carrying capsules filled with such a fire extinguishant are able to operate near multi-floor buildings.
In order to transfer the extinguishant from the~\acp{UAV} and to the fire, the liquid-filled capsules are launched from onboard the~\acp{UAV}.

The~\ac{UAV} platforms and mechanisms for launching the capsules, together denoted as DOFEC\footnote{\href{http://mrs.felk.cvut.cz/projects/dofec}{\url{http://mrs.felk.cvut.cz/projects/dofec}}}, are custom made.
There are two DOFEC versions with distinct dimensions and payload capacities.
Both versions are built on octocopter~\ac{UAV} platforms carrying interoceptive and exteroceptive sensors (IMU, on-board computer, RGB-D, thermal camera, Ouster~\ac{lidar}, and downward rangefinder) required for~\ac{UAV} localization,~\ac{UAV} control, fire detection, fire localization, and positioning near high-rise buildings.
The~\ac{UAV} is localized on the basis of the inertial measurements, the GNSS, and the laser-based SLAM running onboard.
The fires are localized by fusing onboard thermal and RGB-D cameras.

\begin{figure}[h]
    \centering
    %\begin{tikzpicture}
    %   \node[anchor=south west,inner sep=0] (e) at (0,0) {
    %  \includegraphics[width=1.0\textwidth]{fig/dofec_both_versions.jpg}};
    %\node[fill=none, fill opacity=0.3, text=black, text opacity=1.0, anchor=south west] at (0.0, 7.1) {\textbf{(a)}};
    %\node[fill=none, fill opacity=0.3, text=black, text opacity=1.0, anchor=south west] at (7, 7.1) {\textbf{(b)}};
    %\end{tikzpicture}
    \includegraphics[width=1.0\textwidth]{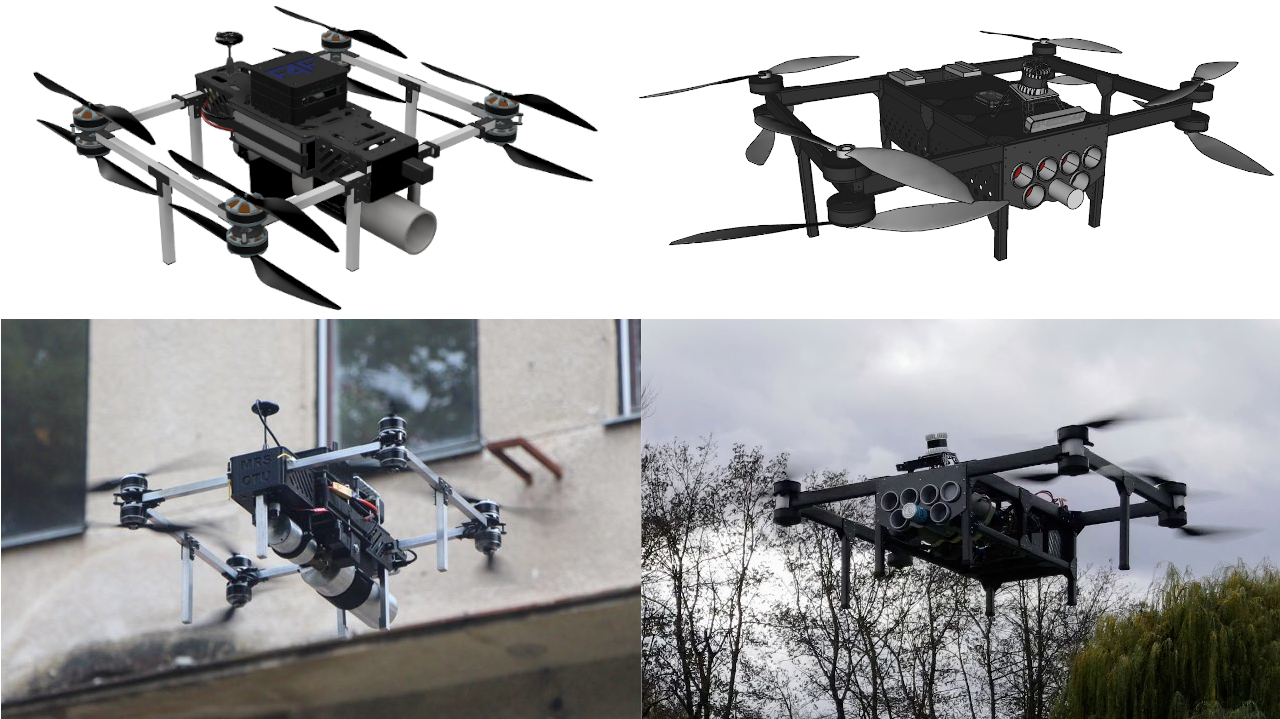}
    \caption{DOFEC platforms (small platform left column, large platform right column) capable of autonomously launching capsules filled with a fire extinguishant into thermal sources. The top row shows CAD models of the platforms, whereas the bottom row shows them during the mission. Video:~\url{https://youtu.be/QHpifXJzH5g}.}
    \label{fig:dofec_2}   
\end{figure}

The small and large platforms are shown in Fig.~\ref{fig:dofec_2}.
The smaller prototype carries a single \SI{500}{\milli\litre} capsule which gives the~\ac{UAV} only one chance of launching it into the fire.
The larger prototype carries a launcher mounted with six of these capsules, allowing for longer operation, wider room for error, and the capability of extinguishing larger fires.
The larger platform has rectangular dimensions of $119 \times \SI{83}{\centi\metre}$ between motors, 30-inch propellers, weighs \SI{41}{\kilo\gram} when fully loaded with six capsules, and reaches a maximal flight time of \SI{15}{\minute}.
Both the single and six-capsule launchers utilize pressurized $\text{CO}_2$ to fire the capsules.
With autonomous~\ac{UAV} control from the~\ac{MRS}~\ac{UAV} system, both platforms seamlessly handle the dynamic recoil from launching the capsules.

%TODO Pavel (only a few sentences, 2 pictures, links to videos)

%%% END SECTION ============================================================

%%% START SECTION ==========================================================

%%[OWNER]: Tomas and Petr Stib
\subsubsection{Localization of radiation sources}
\label{sec:localizationRadiationSources}

% \textbf{TODO:  Petr Stibinger + Tomas}}
An autonomous~\ac{UAV} for nuclear site surveying and homeland security applications was developed through our partnership with the RaDron\footnote{\url{http://mrs.felk.cvut.cz/projects/tacr-radron-project}} project.
The project aims to develop an aerial platform equipped with compact radiation detectors for autonomous operations in hazardous environments.
Over the course of the project, we have developed several platforms of varying sizes and payload capacities.
The final version of the platform is derived from the Holybro X500 in a configuration similar to the one used in the~\ac{darpa}~\ac{subt} Challenge (described in Section~\ref{sec:darpaSubT}).
The qualities proven in the challenge translate well in environments where radiological surveys are to be conducted, such as mine shafts, forests, and industrial objects.

The sensory payload is adapted for both indoor and outdoor operations.
In GNSS-enabled environments, the vehicle relies on the NEO-M8N receiver for positioning.
However, it also carries an Ouster OS0-128~\ac{lidar} for obstacle detection and GNSS-denied localization.

\begin{figure}[tb]
    \centering
    \input{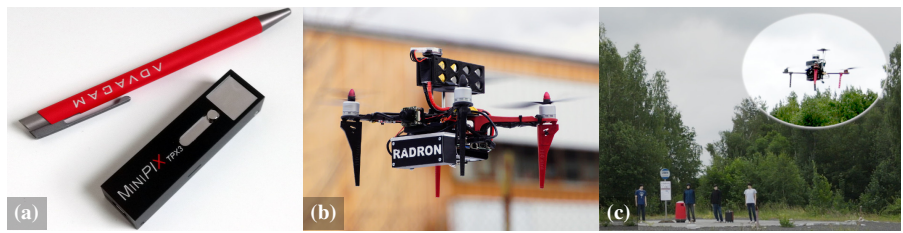}
    %\vspace{-1.0em}
    \caption{Miniature Compton Camera MiniPIX Timepix3 (a) used as an onboard radiation sensor for ionizing radiation localization by~\acp{UAV} (b,c). Video:~\url{https://youtu.be/oH4jMMHfGVA}.}
    \label{fig:radron}
    %\vspace{-1.0em}
\end{figure}

The platform is also equipped with a miniaturized Compton camera, which enables the measurement of radiation intensity, as well as the reconstruction of possible directions toward its source.
The Compton camera is based on the cutting-edge Timepix3 chip sensor developed by the Medipix3 collaboration\footnote{\url{https://medipix.web.cern.ch/medipix3}}.
The sensor is part of the pen-sized MiniPIX Timepix3 USB device (see Fig.~\ref{fig:radron}(a)).
Contrary to similar state-of-the-art devices, the MiniPIX Timepix3 Compton camera only uses a single detector~\cite{turecek2020single}.
This dramatically reduces the size of the device and removes the need for active cooling.

In our applications, we attach the detector to the~\ac{UAV} as a forward-facing camera~\cite{baca2019timepix, stibinger2020localization}.
This configuration enables operation much closer to the ground, and therefore closer to radiation sources than the downward-facing camera used in related projects~\cite{martin20163d, christie2017radiation, towler2012radiation}.
The camera is protected by a 3D-printed casing with a small passive aluminum and graphene heat sink at the top, as shown in Fig.~\ref{fig:radron}.

To facilitate the use of Timepix-based devices in other robotic systems, we have developed a ROS interface Rospix~\cite{baca2018rospix} and released it as open source\footnote{\url{https://github.com/rospix/rospix}}.
Rospix enables real-time data readout and is a convenient way to adjust measurement parameters on the fly.
Together with the device's USB interface, the MiniPIX Timepix3 Compton camera may be used as a plug-and-play device.
The aerial platform is capable of systematic radiation mapping, aerial surveillance, and fast proactive localization of compact radiation sources~\cite{baca2019timepix, stibinger2020localization, baca2021icuas}.
Videos of the platform in action are available at the RaDron project web site\footnote{\url{http://mrs.felk.cvut.cz/projects/tacr-radron-project}}.

%%% END SECTION ============================================================

%%% START SECTION ==========================================================

%%[OWNER]: Afzal Ahmad
\subsubsection{Asteroid inspection}
\label{sec:asteroid_inspection}

\begin{figure}[tb]
    \centering
    \includegraphics[width=1.0\columnwidth]{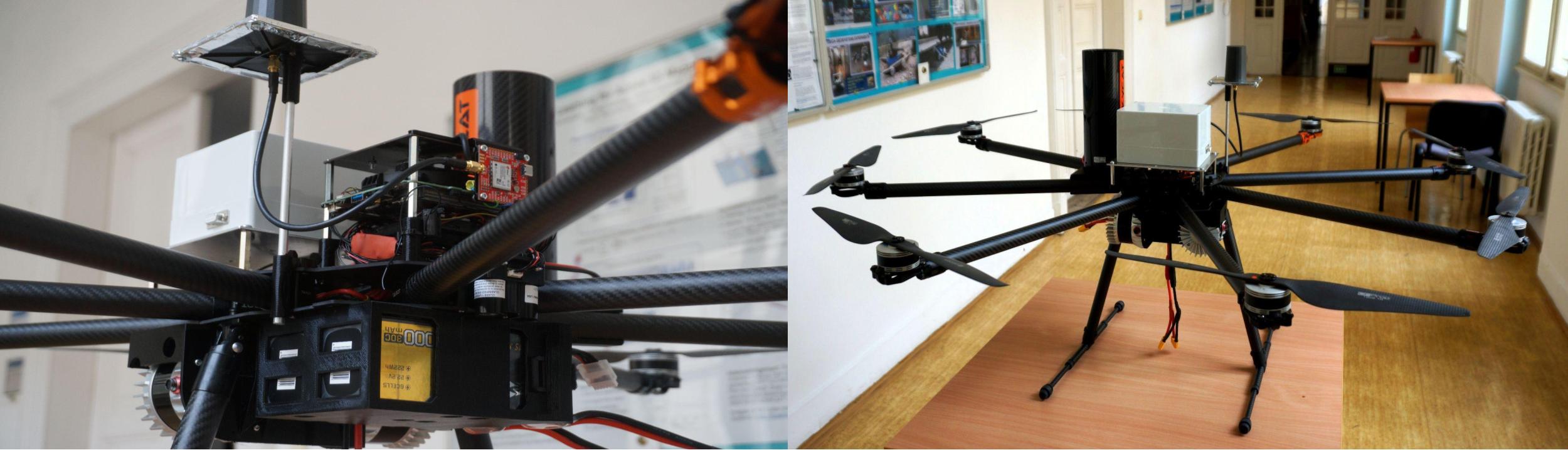}
    \vspace{-1.0em}
    \caption{Prototype platform used for exploration, mapping, and data collection for the asteroid inspection project.}
    \label{fig:asteroid}   
    \vspace{-1.0em}
\end{figure}

Asteroids, moons, and other planetary bodies are a rich source of information about our solar system and the origin of life itself.
However, such complicated terrain makes it challenging to use ground robots in these environments.
We have developed a platform equipped with Ouster~\acp{lidar} to perform autonomous mapping and navigation in hard-to-navigate environments (see Fig. \ref{fig:asteroid}), which can potentially carry several other sensors for collecting information about the planetary body. Our partners\footnote{\url{https://www.unibw.de/lrt9/lrt-9.1/forschung/projekte/kanaria_nakora}} have used the platform to evaluate their algorithms for targeted asteroid landing, gravitational field exploration, and topography and resource localization.

%\textbf{TODO:  Tiago, please see if the figures make sense here, we don't have any other pictures from this project}}

%%% END SECTION ============================================================

%%% START SECTION ==========================================================

%%[OWNER]: Afzal Ahmad
\subsubsection{Package delivery}
\label{sec:pkg_delivery}

\begin{figure}[tb]
    \centering
    \includegraphics[width=1.0\columnwidth]{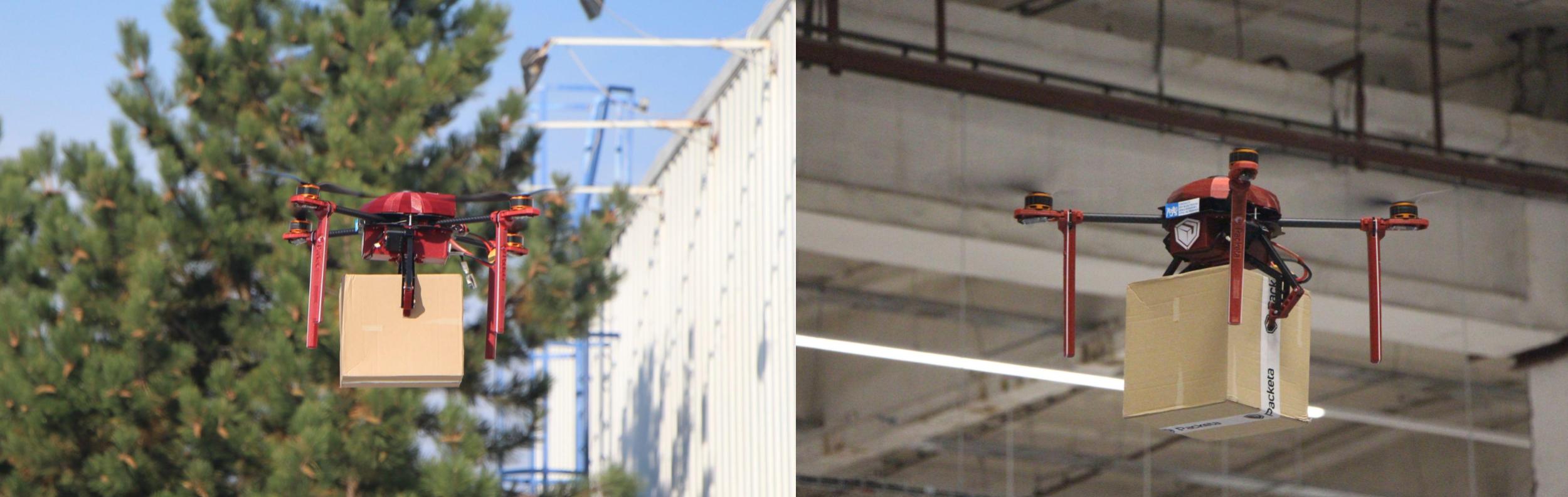}
    \vspace{-1.0em}
    \caption{Platform used for autonomous delivery of packages from a warehouse. Video:~\url{https://youtu.be/C0p514rz14c}.}
    \label{fig:package}   
    \vspace{-1.0em}
\end{figure}

Robots are widely used for logistical applications and have been utilized in several industries, including warehouse automation and assembly lines in manufacturing.
Although autonomous~\acp{UAV} are difficult to use in indoor environments, they have been extremely useful for package delivery applications.
Advancing in this direction, we show in Fig.~\ref{fig:package} a~\ac{UAV} platform developed by our group with a manipulator for quick package delivery\footnote{\url{https://youtu.be/C0p514rz14c}}.
Although the~\ac{UAV} was used to deliver commercial packages, it can be used for delivering medical and other critical supplies as well.
The quick response time of~\ac{UAV}-based delivery offers a significant advantage over conventional transportation methods.

%%% END SECTION ============================================================

%%% START SECTION ==========================================================

%%[OWNER]: Afzal Ahmad
\subsubsection{Indoor inspection}
\label{sec:indoor_inspection}

\begin{figure}[tb]
    \centering
    \input{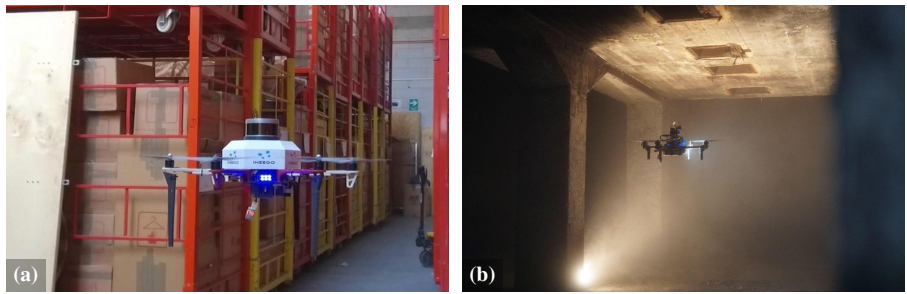}
    \caption{Autonomous inspection of a warehouse and an abandoned building using our~\ac{UAV} platform.}
    \label{fig:indoor_inspection}   
    \vspace{-1.0em}
\end{figure}

Similar to the inspection of outdoor environments like buildings, power lines, and construction sites,~\acp{UAV} can be very beneficial for inspecting large indoor spaces.
Equipped with Velodyne~\ac{lidar}’s Puck LITE-16 and several other optical sensors, we developed a~\ac{UAV} for autonomous warehouse inspection.
The~\ac{UAV} used a path planning method to autonomously map the environment and avoid obstacles, such as warehouse shelves and cargo.
We used an onboard 3D camera to capture high-quality pictures and depth information for later processing to look for specific regions in the warehouse.
We also developed another version of the platform with an Ouster~\ac{lidar} and different sensors to inspect old infrastructure for safety and regulatory purposes.
Figures~\ref{fig:indoor_inspection}(a) and~\ref{fig:indoor_inspection}(b) show the platforms inside a warehouse and an abandoned building, respectively.

%%% END SECTION ============================================================

%%% START SECTION ==========================================================

\section{Conclusions} %%[OWNER]: Martin and Robert
\label{sec:conclusions}

In this work, we have presented the MRS~\ac{UAV} hardware and software system  comprising our MRS Drone, which has been proposed to enable research groups to perform real robot experiments with~\acp{UAV} in real-world conditions.
The proposed platform is modular and the software system is open-source, allowing researchers to achieve proper experimental~\ac{UAV} deployment for both a single~\ac{UAV} or multi-\ac{UAV} system, in either indoor or outdoor scenarios.
Furthermore, we have detailed the technical specifications of the~\ac{UAV} platforms, enabling researchers to prototype and design~\acp{UAV} for research experimentation and validation. 
Finally, the~\ac{MRS} group has accumulated thousands of hours of experimentation and~\ac{UAV} deployments in a large variety of applications, which were also described in this paper. 
Therefore, we intend for this paper to facilitate~\ac{UAV} prototyping for autonomous aerial 
applications in a wider range of scenarios than those presented here.

%%% END SECTION ============================================================

%%% START SECTION ==========================================================

\section*{Declarations}

\noindent\textbf{Acknowledgments} The authors would like to thank the researchers from the Laboratory of Systems Engineering and Robotics Group at the Universidade Federal da Paraíba that, in partnership, helped improve and test our system. Furthermore, the paper is an extension of the manuscript with DOI: 10.1109/ICUAS54217.2022.9836083, from ICUAS 2022.\\

\noindent\textbf{Authors' contributions} In this work, M. Saska coordinated the entire project and group activities. T. Nascimento (the corresponding author) was responsible for coordinating the writing of this manuscript, and the writing of Sections 1, 5.3.2, and 5.3.2. D. Hert coordinated the hardware team and selected the electronic components, actuators, and sensors of the~\acp{UAV}, and wrote Section 3. D. Zaitlik designed the electronic~\ac{MRS} board. P. Stoudek was responsible for designing the various types of mechanical frames and writing Sections 2.1, 2.2, 2.3, and 2.8, while M. Sramek was responsible for printing all the 3D-printed parts and coordinating the mass production of~\ac{MRS}~\acp{UAV} and writing Sections 2.4, 2.5, and 2.7. On the~\ac{MRS} system, T. Baca coordinated the~\ac{MRS} software development and wrote Sections 4 and 5.3.4. V. Walter designed the~\ac{UVDAR}~\ac{MRS} system. D. Bonilla Licea helped improve communication through~\ac{UVDAR} and wrote Section 2.6. P. Stepan helped develop the vision system of~\ac{MRS}~\acp{UAV}. V. Spurny was responsible for developing the simulation environment and wrote Section 5.1. On upgrading the~\ac{MRS} Drone and~\ac{MRS}~\ac{SW} system to attend to various applications, M. Petrlik designed the~\ac{MRS} Drone localization system and wrote Section 5.2.1, while V. Pritzl helped improve the sensor fusion of the~\ac{MRS} Drone and helped write Sections 5.2.1 and 5.4.2. P. Petracek designed the~\ac{MRS} Drone (\ac{HW} and~\ac{SW}) for the Dronument project and wrote Sections 5.2.2 and 5.4.2, while V. Kratky designed the~\ac{MRS} Drone designed the~\ac{MRS} Drone (\ac{HW} and~\ac{SW}) for industrial inspections and wrote Section 5.2.3. J. Horyna focused on swarming in deserts and modified the~\ac{MRS} Drone for such applications, as well as wrote Section 5.3.1. M. Vrba focused on the vision system, localization, and AI-based applications of the~\ac{MRS} Drone, while also writing Sections 5.4, 5.4.1, and 5.4.6. P. Stibinger helped improve the~\ac{MRS} Drone~\ac{SW} system and wrote Section 5.4.3. A. Ahmad applied the~\ac{MRS} Drone to outer space applications and manipulation and wrote Sections 5.4.4 and 5.4.5. R. Penicka designed the trajectory and path planning of the~\ac{MRS} Drone, revised the entire manuscript, helped write Section 1, and wrote the abstract and Section 6. Finally, G. Silano wrote Section 5.3.5 and helped revise the entire manuscript.\\

\noindent\textbf{Code or data availability} The authors declare that the data supporting the findings of this study are available within the MRS software link (\url{https://github.com/ctu-mrs/mrs_uav_system}) and MRS Drone Builder link (\url{https://dronebuilder.fly4future.com/}).\\

\noindent\textbf{Consent for Publication} Informed consent was obtained from all the co-authors of this publication. \\

\noindent\textbf{Competing interests} The authors declare that they have no conflict of interest. \\

\noindent\textbf{Ethics approval and consent to participate} All applicable institutional and national guidelines were followed. \\

\noindent\textbf{Funding} This work was partially funded by the CTU grant no. SGS20/174/OHK3/3T/13, by the Czech Science Foundation (GAČR) under research project no. 20-10280S, no. 20-29531S, no. 22-24425S and no. 23-06162M, by TAČR project no. FW01010317, by the OP VVV funded project CZ.02.1.01/0.0/0.0/16 019/0000765 ``Research Center for Informatics", by the NAKI II project no. DG18P02OVV069, by the European Union's Horizon 2020 research and innovation program AERIAL-CORE under grant agreement no. 871479, by the~\acf{darpa}, and by the Technology Innovation Institute - Sole Proprietorship LLC, UAE. Furthermore, computational resources were supplied by the project ``e-Infrastruktura CZ" (e-INFRA LM2018140) provided within the program Projects of Large Research, Development and Innovations Infrastructures.

% ============================================================
\begin{acronym}
  \acro{CNN}[CNN]{Convolutional Neural Network}
  \acro{ESC}[ESC]{Elettronic Speed Controller}
  \acro{HW}[HW]{Hardware}
  \acro{IR}[IR]{infrared}
  \acro{GNSS}[GNSS]{Global Navigation Satellite System}
  \acro{HRI}[HRI]{Human-Robot interaction}
  \acro{MAV}[MAV]{Micro Aerial Vehicle}
  \acro{MOCAP}[mo-cap]{Motion capture}
  \acro{MPC}[MPC]{Model Predictive Control}
  \acro{MBZIRC}[MBIZRC]{Mohamed Bin Zayed International Robotics Challenge}
  \acro{MRS}[MRS]{Multi-robot Systems}
  \acro{ML}[ML]{Machine Learning}
  \acro{UAV}[UAV]{Unmanned Aerial Vehicle}
  \acro{UV}[UV]{UltraViolet}
  \acro{UVDAR}[UVDAR]{UltraViolet Direction And Ranging}
  \acro{USV}[USV]{Unmanned Surface Vehicle}
  \acro{UT}[UT]{Unscented Transform}
  \acro{REL}[REL]{Reinforced Epoxy Laminate}
  \acro{RTK}[RTK]{Real-Time Kinematic}
  \acro{ROS}[ROS]{Robot Operating System}
  \acro{SITL}[SITL]{Software-In-The-Loop}
  \acro{SW}[SW]{Software}
  \acro{wrt}[w.r.t.]{with respect to}
  \acro{UGV}[UGV]{Unmanned Ground Vehicle}
  \acro{darpa}[DARPA]{Defense Advanced Research Projects Agency}
  \acro{subt}[SubT]{Subterranean Challenge}
  \acro{sar}[S\&R]{Search and Rescue}
  \acro{lidar}[LiDAR]{Light Detection and Ranging}
  \acro{VRP}[VRP]{Vehicle Routing Problem}
  \acro{VIO}[VIO]{Visual-Inertial Odometry}
  \acro{FCU}[FCU]{flight control unit}
\end{acronym}

\bibliographystyle{spmpsci}
\bibliography{bib}% common bib file

\end{document}